%% file: main.tex
\pdfoutput=1
\documentclass{article}

\PassOptionsToPackage{numbers}{natbib}
\usepackage[preprint]{neurips_2026/neurips_2026}

\usepackage{amsmath}
\usepackage{amssymb}
\usepackage{graphicx}
\usepackage{placeins}
\usepackage{algorithm}
\usepackage{algpseudocode}
\usepackage{hyperref}
\hypersetup{
  pdftitle={SilverSpoon-KD: A General-Purpose Toolkit for Knowledge Distillation},
  pdfauthor={Xaver R. Davey and David Broman},
  colorlinks=true,
  allcolors=blue!50!black,
}
\usepackage{url}

\usepackage[acronym,nohypertypes={main,acronym}]{glossaries}
\PassOptionsToPackage{table}{xcolor}
\usepackage{tikz}
\usetikzlibrary{positioning, fit, calc, backgrounds, arrows.meta, shapes, matrix}
\usepackage{pgfplots}
\pgfplotsset{compat=1.18}
\usepgfplotslibrary{groupplots}
\usepackage{siunitx}
\usepackage{booktabs}
\usepackage{enumitem}
\usepackage{listings}

\lstdefinestyle{kdpython}{
  language=Python,
  basicstyle=\ttfamily\footnotesize,
  keywordstyle=\color{blue!60!black}\bfseries,
  stringstyle=\color{cStudent!50!black},
  commentstyle=\color{gray!60!black}\itshape,
  emph={create_alignments, TrainingArguments, kl_divergence_loss},
  emphstyle={\color{cStudent!80!black}\bfseries},
  showstringspaces=false,
  breaklines=true,
  breakatwhitespace=true,
  columns=fullflexible,
  keepspaces=true,
  xleftmargin=0pt,
  framesep=4pt,
  abovecaptionskip=4pt,
  belowcaptionskip=2pt,
}

\definecolor{cReskd}{HTML}{0072B2}
\definecolor{cHkd}{HTML}{E69F00}
\definecolor{cBkd}{HTML}{D55E00}
\definecolor{cRelkd}{HTML}{CC79A7}
\definecolor{cStudent}{HTML}{009E73}
\definecolor{cAlignment}{HTML}{F0E442}
\definecolor{cMulti}{HTML}{009E73}

\algnewcommand\algorithmicinput{\textbf{Input:}}
\algnewcommand\algorithmicoutput{\textbf{Output:}}
\algnewcommand\Input{\item[\algorithmicinput]}
\algnewcommand\Output{\item[\algorithmicoutput]}
\DeclareSIUnit{\times}{\text{\ensuremath{\times}}}

\newcommand{\pB}{{\setlength{\fboxsep}{1pt}\colorbox{cBkd!22}{\tiny\bfseries\sffamily B\strut}}}
\newcommand{\pH}{{\setlength{\fboxsep}{1pt}\colorbox{cHkd!28}{\tiny\bfseries\sffamily H\strut}}}
\newcommand{\pR}{{\setlength{\fboxsep}{1pt}\colorbox{cReskd!22}{\tiny\bfseries\sffamily Res\strut}}}
\newcommand{\pRel}{{\setlength{\fboxsep}{1pt}\colorbox{cRelkd!28}{\tiny\bfseries\sffamily Rel\strut}}}
\newcommand{\chk}{\textcolor{cStudent!80!black}{\ensuremath{\checkmark}}}

\newcommand{\pgrad}[1]{{\tiny\textcolor{red!70!black}{\ensuremath{#1}}}}

\colorlet{cOkBlue}{cReskd}
\colorlet{cOkOrange}{cHkd}
\colorlet{cOkVermil}{cBkd}
\colorlet{cOkPurple}{cRelkd}
\colorlet{cOkGreen}{cStudent}
\colorlet{cOkYellow}{cAlignment}
\colorlet{cOkSky}{cMulti}
\colorlet{cStock}{cOkSky!22}
\colorlet{cAdded}{cOkOrange!50}
\colorlet{cChanged}{cOkYellow!42}
\colorlet{cTSlot}{gray!28}
\colorlet{cSSlot}{cOkGreen!28}
\colorlet{cTSlotBorder}{gray!70!black}
\colorlet{cSSlotBorder}{cOkGreen!75!black}
\newcommand{\mtitle}[1]{{\sffamily\fontsize{6}{7}\selectfont\bfseries #1}}

\newglossaryentry{llm} {
        name={LLM},
        first={Large Language Model (LLM)},
        firstplural={Large Language Models (LLMs)},
        text={LLM},
        plural={LLMs},
        description={}
}

\newglossaryentry{kd} {
        name={KD},
        first={knowledge distillation (KD)},
        text={KD},
        description={}
}

\newglossaryentry{featkd} {
        name={FeatKD},
        first={Feature-Based Knowledge Distillation (FeatKD)},
        text={FeatKD},
        description={}
}

\newglossaryentry{reskd} {
        name={ResKD},
        first={Response-Based Knowledge Distillation (ResKD)},
        text={ResKD},
        description={}
}

\newglossaryentry{hkd} {
        name={HKD},
        first={Holistic Knowledge Distillation (HKD)},
        text={HKD},
        description={}
}

\newglossaryentry{bkd} {
        name={BKD},
        first={Blockwise Knowledge Distillation (BKD)},
        text={BKD},
        description={}
}

\newglossaryentry{relkd} {
        name={RelKD},
        first={Relation-Based Knowledge Distillation (RelKD)},
        text={RelKD},
        description={}
}

\newglossaryentry{linearization} {
        name=linearization,
        description={}
}

\newglossaryentry{open-weight} {
        name=open-weight,
        description={}
}

\title{SilverSpoon-KD: A General-Purpose Toolkit for Knowledge Distillation}

\author{%
  Xaver R.~Davey \\
  KTH Royal Institute of Technology \\
  Stockholm, Sweden \\
  \texttt{xrdavey@kth.se} \\
  \And
  David Broman \\
  KTH Royal Institute of Technology \\
  Stockholm, Sweden \\
  \texttt{dbro@kth.se} \\
}

\begin{document}
\maketitle

\begin{abstract}

Knowledge distillation transfers deep learning representations from a teacher model to a student model, typically reducing model size or inference cost. However, knowledge distillation tooling has not kept pace with the breadth of research methods: existing toolkits typically support only a limited subset of techniques and use cases. We present SilverSpoon-KD, an open-source PyTorch toolkit that unifies different distillation paradigms and architectural relationships across workloads and applications behind a single API. Moreover, SilverSpoon-KD offers the first general-purpose blockwise knowledge distillation implementation with unique scalability features. We validate its versatility across vision, masked language model, and causal language models on tasks spanning homogeneous, heterogeneous, and quantization distillation across paradigms. We also evaluate SilverSpoon-KD by comparing its performance---including throughput and GPU memory usage---with existing toolkits. Despite extensive support for convenience, metric calculation, and tracking, we have found SilverSpoon-KD's performance to be on par with or better than existing toolkits.

\end{abstract}

\section{Introduction}

\Gls{kd}~\cite{hinton2015distilling} is a family of techniques for transferring learned representations from a \emph{teacher} model (the trained source) to a \emph{student} model (the target being trained, typically smaller), primarily to reduce inference cost (Figure~\ref{fig:kd_taxonomy}a). \Gls{kd} workflows vary along three largely orthogonal axes (Figure~\ref{fig:kd_taxonomy}b--d): the \emph{workload} (general task or training objective), the \emph{architectural relationship} between teacher and student, and the \emph{distillation paradigm} (how signals flow from teacher to student). Architectural relationships span homogeneous compression (same architecture family, smaller student) \cite{sanh2019distilbert, jiao2020tinybert}, heterogeneous transfer (different architecture families with distinct internal operations) \cite{hao2023one, lin2025feature}, quantization-based distillation \cite{polino2018model}, and self-distillation \cite{furlanello2018born, zhang2019your}. These techniques are central to deploying large models at scale, with successes spanning compressed CNN image classifiers \cite{beyer2022knowledge}, encoder language models like DistilBERT \cite{sanh2019distilbert}, and the distillation of modern LLMs \cite{busbridge2025distillation}.

\input{static/kd_taxonomy}

The paradigm axis comprises four families. \emph{Response-based \gls{kd}} is the most fundamental approach, matching the teacher's output distribution \cite{hinton2015distilling}. By contrast, \emph{feature-based \gls{kd}} aligns intermediate representations either through \emph{holistic \gls{kd}}\footnote{We introduce ``holistic KD'' to name the previously implicit end-to-end default, distinguishing it from blockwise \gls{kd}.}\cite{romero2014fitnets} or \emph{blockwise \gls{kd}} \cite{wang2018progressive}, with isolated student blocks trained with teacher-forced inputs. Finally, \emph{relation-based \gls{kd}} \cite{park2019relational} matches inter-sample relationships and can compose orthogonally with any of the other paradigms. 

Despite the breadth and maturity of \gls{kd} research, tooling has arguably not kept pace: response-based \gls{kd} is well-served, but feature-based \gls{kd} demands capturing intermediate tensors that vary significantly across architectures, often forcing researchers to build custom pipelines per project. A few dedicated toolkits support only holistic \gls{kd}, and to our knowledge, no existing toolkit facilitates blockwise \gls{kd} out of the box. \Gls{kd} tools have also lagged behind the broader deep learning ecosystem: the community increasingly expects scalable training infrastructure with built-in evaluation, logging, and multi-device support---as exemplified by Hugging Face's Transformers \cite{wolf2020transformers}. However, current \gls{kd} tooling often lacks these engineering conveniences.

This paper presents SilverSpoon-KD\footnote{SilverSpoon-KD's source code: \url{https://github.com/silverspoon-dev/silverspoon-kd}}, a toolkit designed to be a comprehensive solution for \gls{kd}. Our key contributions are:

\begin{itemize}[topsep=-1pt, leftmargin=9pt, itemsep=-2pt]
    \item \textbf{Multi-workload, multi-architecture, multi-paradigm coverage in one API}: SilverSpoon-KD supports key workloads (vision, masked LM, causal LM), all architectural relationships, and all four KD paradigms behind a single API. A general-purpose abstraction is a non-trivial contribution (removing hard-coded assumptions while keeping the API ergonomic).

    \item \textbf{First general-purpose blockwise \gls{kd} implementation}: To our knowledge, SilverSpoon-KD is the first open-source toolkit to provide a general abstraction for blockwise \gls{kd}. Prior implementations have been narrowly scoped and tied to specific use-cases, leaving users to build custom pipelines.

    \item \textbf{Practical engineering breadth with measured performance gains}: SilverSpoon-KD interoperates with PyTorch~\cite{paszke2019pytorch}, the HuggingFace Trainer, and the standard parallelism stack (DDP, FSDP, DeepSpeed). It offers flexible teacher placement, automatic teacher--student dimension matching via projectors, asynchronous teacher--student CUDA streams, and built-in evaluation. The resulting throughput and memory gains over existing toolkits are modest but consistently present across all three workloads; they accrue primarily from the sum of many parts rather than from any single optimization, which is notable given SilverSpoon-KD's broader feature surface.
\end{itemize}
 
\section{Related Work}
\label{sec:related_work}

Several open-source toolkits provide varying degrees of support for \gls{kd}, though none currently offer the combination of workload, architecture coverage, paradigm breadth, and scalability features that SilverSpoon-KD targets (see Table~\ref{tab:feature_comparison}). The table records which features each toolkit supports, not how well each is implemented; performance benchmarks in \S\ref{exp:framework} address the latter.

\begin{table}[!tp]
\centering\footnotesize
\setlength{\tabcolsep}{6pt}
\renewcommand{\arraystretch}{0.95}
\caption{\textbf{Feature comparison of knowledge distillation toolkits.} SilverSpoon-KD provides the broadest feature surface across the five compared toolkits, including the only general-purpose blockwise KD implementation and FSDP-sharded teacher placement.}
\label{tab:feature_comparison}
\begin{tabular}{@{}lcccccc@{}}
\toprule
Feature & \rotatebox{60}{SilverSpoon-KD} & \rotatebox{60}{torchdistill} & \rotatebox{60}{TextBrewer} & \rotatebox{60}{torchtune} & \rotatebox{60}{DistillKit} & \rotatebox{60}{EasyDistill} \\
\midrule
\multicolumn{7}{@{}l}{\emph{Distillation paradigms}} \\
\quad Response-based \gls{kd}          & \chk & \chk & \chk & \chk & \chk & \chk \\
\quad Holistic feature \gls{kd}          & \chk & \chk & \chk &        & \chk &        \\
\quad Blockwise feature \gls{kd}                & \chk &        &        &        &        &        \\
\quad Relation-based \gls{kd}           & \chk & \chk   &        &        &        &        \\
\addlinespace
\multicolumn{7}{@{}l}{\emph{Workload support}} \\
\quad Encoder NLP (BERT, etc.)              & \chk & \chk & \chk &        &        &        \\
\quad Decoder LLM (GPT-2, Qwen, etc.)       & \chk &        & \chk$^\S$ & \chk & \chk & \chk \\
\quad Vision (CNN, ViT)                     & \chk & \chk &        &        &        &        \\
\addlinespace
\multicolumn{7}{@{}l}{\emph{Training features}} \\
\quad Auto dimension projectors             & \chk &        &        &        & \chk &        \\
\quad Sub-module alignment (e.g.\ attention) & \chk & \chk &        &        &        &        \\
\quad Cross-architecture distillation       & \chk &        &        &        & \chk &        \\
\quad Quantization-aware training           & \chk$^*$ &        &     & \chk   &        &        \\
\quad Multi-teacher distillation            &        &        & \chk &        &        & \chk \\
\quad Teacher output caching                &        &        & \chk &        & \chk & \chk \\
\quad Built-in evaluation                   & \chk &        &        & \chk &        & \chk \\
\addlinespace
\multicolumn{7}{@{}l}{\emph{Scalability}} \\
\quad DDP \emph{(student)}                  & \chk & \chk & \chk &        & \chk &        \\
\quad FSDP \emph{(student)}                 & \chk &        &        & \chk   &        &        \\
\quad DeepSpeed \emph{(student)}            & \chk &        &        &        & \chk & \chk \\
\quad FSDP sharding \emph{(teacher)}        & \chk &        &        & \chk   &        &        \\
\quad Pipeline / tensor parallel \emph{(teacher)} & \chk &        &        &        &        &        \\
\quad Split-GPU placement \emph{(teacher)}  & \chk &        &        &        &        &        \\
\quad Chunked KD losses                      & \chk &        &        & \chk &        &        \\
\quad Fused kernel KD losses                  & \chk &        &        &        &        &        \\
\bottomrule
\end{tabular}

\smallskip
{\footnotesize $^*$SilverSpoon-KD's quantization-aware training (QAT) support is via external model preparation, not via built-in QAT recipes; see Appendix~\ref{exp:qwen3_quant}. $^\S$TextBrewer is documented for encoder NLP only, but response-based \gls{kd}/holistic \gls{kd} on decoder LLMs work empirically through its generic distiller.\par}
\end{table}

\medskip

\paragraph{General-purpose KD toolkits.} The five toolkits in Table~\ref{tab:feature_comparison} are the most directly comparable general-purpose \gls{kd} libraries. \textit{torchdistill} \cite{matsubara2021torchdistill} is the broadest method library but does not target causal \glspl{llm} or blockwise training; \textit{TextBrewer} \cite{yang2020textbrewer} specializes in holistic \gls{kd} for encoder NLP, with feature alignment restricted to higher-level module outputs; \textit{torchtune} \cite{torchtune2024} and \textit{DistillKit} \cite{distillkit2024} target decoder \glspl{llm}, the latter adding hidden-state matching with automatic projectors and logit-compressed teacher caching; and \textit{EasyDistill} \cite{wang2025easydistill} emphasizes black-box pipelines (training on teacher-generated text) alongside multi-teacher aggregation and logit caching.

\paragraph{Adjacent toolkits.} Several other toolkits are in an adjacent \gls{kd} space rather than competing as general-purpose \gls{kd} toolkits, and so are noted here but excluded from Table~\ref{tab:feature_comparison}. \textit{RepDistiller} \cite{tian2019contrastive} is a research benchmark, not a toolkit: it bundles 12 vision \gls{kd} methods under shared infrastructure for reproducibility studies rather than as a substrate for building new pipelines, and its repository has been unmaintained for years. The other three are broader model-compression libraries that treat limited \gls{kd} methods as one technique among several. \textit{KD-Lib} \cite{shah2020kd} bundles vision \gls{kd} with pruning (removing redundant weights or neurons), quantization, and integrated hyperparameter tuning, with no decoder \gls{llm} or blockwise \gls{kd} support. \textit{NVIDIA ModelOpt} \cite{nvidia_modelopt} integrates \gls{llm} logit distillation into NeMo alongside quantization and pruning. Finally, \textit{Intel Neural Compressor} \cite{intel_neural_compressor} bundles \gls{kd} with broader compression across PyTorch and TensorFlow. All three of the latter prioritize deployment optimization over \gls{kd} methodology breadth, so a feature-by-feature comparison against SilverSpoon-KD's \gls{kd}-focused surface would mostly measure scope rather than capability.

\paragraph{Blockwise Knowledge Distillation.} Blockwise \gls{kd} trains student blocks in isolation against corresponding teacher blocks, decomposing a difficult end-to-end optimization problem into simpler per-block regression tasks. Wang et al.\ \cite{wang2018progressive} and Koratana et al.\ \cite{koratana2018lit} introduced the sequential and isolated per-block formulations; Blakeney et al.\ \cite{blakeney2020parallel} and Jang et al.\ \cite{jang2023pipe} extended these to parallel and pipelined variants; Shen et al.\ \cite{shen2021progressive} explored progressive network grafting; and Kao et al.\ \cite{kao2023stablekd} and Yu et al.\ \cite{yu2024decoupling} addressed inter-block optimization entanglement and feature-level decoupling. Despite this active research, existing blockwise \gls{kd} implementations remain tightly coupled to specific experimental setups; SilverSpoon-KD aims to provide a general-purpose blockwise \gls{kd} toolkit that can accommodate these diverse strategies.

\section{System Design and Usage}
\label{sec:system_design}

We first cover SilverSpoon-KD's internal architecture (\S\ref{ssec:design}) and then its programmatic API (\S\ref{ssec:usage}).

\subsection{Design}
\label{ssec:design}

SilverSpoon-KD is built on top of the HuggingFace Trainer, which lets the distiller inherit general-purpose deep learning boilerplate while adding \gls{kd}-specific logic. Figure~\ref{fig:architecture}a traces the distiller's training-step call flow, with nodes color-coded by provenance. The \gls{kd}-specific surface area is intended to be concentrated to leverage the general training logic provided by Hugging Face. Beyond this scaffolding, three components enable effective \gls{kd}, each described in turn below.

\textbf{(1)} Alignments (Figure~\ref{fig:architecture}b) map which student modules learn from which teacher modules. SilverSpoon-KD attaches forward hooks that record module inputs and outputs without modifying the underlying models, and a pairing API (module-name list or regex, if more convenient) targets any granularity---whole layers or embedded sub-modules---even across heterogeneous teacher--student naming. \textbf{(2)} Parallelism (Figure~\ref{fig:architecture}c) places teacher and student under selectable, independent regimes across devices and uses separate CUDA streams so their forward passes overlap; this works because the two forwards have no data dependency until the loss step. \textbf{(3)} Auto-projectors (Figure~\ref{fig:architecture}d) detect teacher--student hidden-dimension mismatches at alignment time and auto-build learnable projectors trained jointly with the student. Among compared toolkits, this combination of arbitrary-granularity alignment and overlapping multi-placement teacher--student forwards is, to our knowledge, unique; DistillKit is the only other compared toolkit with automatic projectors.

The load-bearing distinction between holistic and blockwise feature-based \gls{kd} is whether each student block receives the previous \emph{student} block's output (full end-to-end student forward pass; sequential gradients) or a detached teacher-forced input (per-block-independent gradients; lower peak memory; parallelizable across alignments). Appendix~\ref{app:paradigm_visualizations} visualizes the gradient-flow topologies.

\input{static/architecture}

\subsection{Usage}
\label{ssec:usage}

In contrast to most general-purpose \gls{kd} toolkits (torchdistill, torchtune, DistillKit, EasyDistill), which expose their main entry point as a YAML or JSON configuration schema, SilverSpoon-KD's API is intentionally programmatic Python. This keeps integration with bespoke training code direct. Users can embed a distiller inside an existing HuggingFace Trainer workflow, mix in custom callbacks or evaluation hooks, and avoid being locked into a fixed configuration format. The algorithmic difference between paradigms is hidden behind a single class swap (Figure~\ref{fig:api_snippet}): the same alignment objects, training arguments, dataset, and teacher/student models flow into either feature-based distiller, and response-based \gls{kd} drops the alignments entirely. Paradigm-specific options are surfaced transparently through each constructor.

\input{static/api_snippet}

More information is available in the online documentation.\footnote{SilverSpoon-KD's online documentation: \url{https://kd.silverspoon.dev}}

\section{Experiments}
\label{sec:experiments}

\input{static/experiment_coverage}

We evaluate SilverSpoon-KD across eight distillation-quality experiments and a ninth experiment benchmarking throughput and peak memory against four existing KD libraries. Figure~\ref{fig:experiment_coverage} summarizes the axes each experiment stresses and the SilverSpoon-KD capabilities it exercises; full per-experiment results are in Appendix~\ref{app:experiments}.

\subsection{Experimental Setup}

This experimental program addresses three research questions tied to the contributions in the introduction: (\textbf{RQ1}) Does the unified API support non-trivial distillation across workloads, architectures, and paradigms? (\textbf{RQ2}) Does SilverSpoon-KD's blockwise KD implementation deliver useful results across compression regimes, including in multi-stage runs? (\textbf{RQ3}) Does the toolkit's engineering breadth translate into measured throughput and memory gains over existing libraries? Figure~\ref{fig:experiment_coverage}'s rightmost column flags which experiments speak to each question.

The eight distillation-quality experiments are positioned as workflow demonstrations rather than method-novelty studies: each follows a published recipe or typical KD setup, exercising the toolkit end-to-end through the same unified API to verify that it faithfully covers the workload--architecture--paradigm matrix of Figure~\ref{fig:experiment_coverage}, with head-to-head comparisons against established toolkits or reference papers where possible.

Most experiments were conducted on NVIDIA A40 GPUs (48GB) from the Alvis cluster at Chalmers University of Technology in Gothenburg, Sweden; the exception is \S\ref{exp:framework}, where short, hard-to-compute-saturate runs used NVIDIA A100s (40GB). Taken together, these experiments required approximately 2\,000 GPU hours. Given the breadth of the workload, architecture, paradigm, and placement matrix, each reported configuration is a single-seed run, with per-experiment system metrics (peak memory, steps/s, wall time) reported alongside accuracy in Appendix~\ref{app:experiments}.

All the experiment code used for this paper can be found in a GitHub repository.\footnote{SilverSpoon-KD's experiments repository: \url{https://github.com/silverspoon-dev/silverspoon-kd-paper-experiments}}

\subsection{Distillation-Quality Results}
\label{exp:distillation_results}

Figure~\ref{fig:results_summary} summarizes the headline metric for each of the eight distillation-quality experiments. Full per-run results are reported in Appendix~\ref{app:experiments}.

\input{generated/results_summary}

\paragraph{Vision (Figure~\ref{fig:results_summary}, panels~1--2).} VGG Compression (Appendix~\ref{exp:vision_bkd}) replicates Wang et al.\ \cite{wang2018progressive} by distilling VGG-16 into a depthwise-separable VGG-16-DS on CIFAR-10~\cite{krizhevsky2009learning}; the BKD~$\rightarrow$~ResKD pipeline reaches 90.42\% (vs.\ teacher 87.26\%, baseline 82.69\%), with auto-projectors handling channel-dimension mismatches transparently. This demonstrates that the block abstraction generalizes to convolutional networks. VGG Relational KD (Appendix~\ref{exp:vision_relkd}) replicates Park et al.\ \cite{park2019relational} (ResNet-50~$\rightarrow$~VGG-11-BN on CIFAR-100), exercising SilverSpoon-KD's pluggable loss-function architecture: ResKD leads at 73.65\% and the relation-based variants (distance, angle, distance+angle) span 70.40--71.61\%, all reachable through configuration-only changes.

\paragraph{BERT (Figure~\ref{fig:results_summary}, panels~3--4).} BERT Pre-Training (Appendix~\ref{exp:bert_dist}) distills BERT-base into bert\_T6 (${\approx}59\%$ of teacher params) and bert\_T4\_tiny (${\approx}13\%$) on Dolma; HKD wins at both compression levels but the gap widens dramatically with aggressiveness---the three paradigms cluster within ${\approx}1.3$ percentage points (pp) at T6 (HKD +3.04pp MNLI~\cite{williams2018broad}) and fan out to a 5.2pp spread at T4-tiny, where HKD remains strongest (+3.52pp) while BKD turns actively harmful, exposing a compression--paradigm interaction that the unified class-swap API made cheap to surface. BERT Downstream (Appendix~\ref{exp:bert_downstream}) compares against TextBrewer on its own MNLI benchmark: HKD~$\rightarrow$~fine-tune (FT) matches BERT-base within 0.12pp on T6 and beats TextBrewer by 3.38pp.

\paragraph{Decoder LLM, compression and transfer (Figure~\ref{fig:results_summary}, panels~5--6).} GPT-2 Compression (Appendix~\ref{exp:gpt2_dist}) aggressively shrinks an already-small 124M GPT-2 (\texttt{n\_inner} 3{,}072~$\to$~1{,}536, a 50\% MLP-width reduction); standalone BKD diverges to perplexity (PPL) 245{,}281, but the three-stage BKD~$\rightarrow$~HKD~$\rightarrow$~LoRA pipeline rescues it (PPL~93.39, +28\% LAMBADA~\cite{paperno2016lambada} over baseline). Cross-Architecture (Appendix~\ref{exp:gpt2_cross_arch}) distills GPT-2~$\rightarrow$~DeepSeek-V3 across multi-head attention (MHA) vs.\ MLA, learned vs.\ RoPE, and GELU vs.\ SiLU; BKD was not feasible because per-block teacher-forced inputs land in a representation the student's blocks cannot consume. The teacher--student performance gap is notable, drawing a practical limit on how heterogeneous \gls{kd} can be.

\paragraph{Decoder LLM, capability showcases (Figure~\ref{fig:results_summary}, panels~7--8).} GPT-2 Attention Linearization (Appendix~\ref{exp:gpt2_linear}) exercises a unique SilverSpoon-KD capability---\emph{granular sub-module targeting}, where \gls{bkd} aligns only the attention mechanism within each layer rather than the full layer output; \gls{bkd}~$\rightarrow$~FT (PPL~52.01) is the only configuration that beats the LoLCATs~\cite{zhang2024lolcats} no-KD floor of 53.49, and the experiment shows that comparable toolkits would require manual hook plumbing for this configuration. Qwen3 INT4 QAT (Appendix~\ref{exp:qwen3_quant}) distills Qwen3-1.7B into a simulated quantized student: at INT4, HKD lifts MMLU~\cite{hendrycks2020measuring} by 6.76pp toward the bf16 ceiling and delivers the lowest perplexity, ResKD edges out perplexity at INT5, and BKD provides distillation at zero memory overhead vs.\ QAT alone.

\subsection{Toolkit Benchmarking}
\label{exp:framework}

We benchmark SilverSpoon-KD against four comparison toolkits on a featured use case of each: torchdistill (vision, response-based), TextBrewer (encoder NLP, response-based and holistic), torchtune (decoder LLM, response-based+LoRA), and DistillKit (decoder LLM, holistic), matching each toolkit's default setup as closely as possible; full per-toolkit configurations and additional placement sweeps are in Appendix~\ref{exp:framework_detail}. The other toolkits in Table~\ref{tab:feature_comparison} are excluded for methodological reasons: EasyDistill's pre-cached-teacher pipeline is incommensurable with our live-teacher comparison, and KD-Lib and RepDistiller target vision only.

\input{generated/framework}

\input{generated/framework_capability}

SilverSpoon-KD is within ${\pm}6\%$ wall time of every comparison toolkit while using less GPU memory in all four (13--24.4\% reductions). The capability frontier (Table~\ref{tab:framework_capability}) sweeps teacher size on 2$\times$A100: at modest scales (Qwen3-4B, 8B) all toolkits achieve comparable throughput, but at Qwen3-14B (28~GB bf16 weights), DistillKit fails to fit and TextBrewer collapses to bs=1, while SilverSpoon-KD's FSDP-sharded teacher halves the per-GPU teacher footprint, fits 3$\times$ TextBrewer's batch, and overtakes both on tokens/sec.

\subsection{Discussion}
\label{exp:summary}

Three cross-workload patterns emerge. First, multi-stage chaining wins consistently: essentially every workload's headline result comes from a chain (BKD~$\rightarrow$~ResKD on VGG, HKD~$\rightarrow$~FT on BERT, BKD~$\rightarrow$~HKD~$\rightarrow$~LoRA on GPT-2), making SilverSpoon-KD's easy setup a key composition mechanism rather than mere convenience. Second, paradigm--setting interactions emerge cheaply: BKD's role shifts from competitive at mild BERT compression to actively harmful at aggressive compression, and which paradigm wins partitions across three axes under Qwen3 INT4 QAT, both surfaced through class-swap rather than re-implementation. Finally, the toolkit recovers workflows where distillation alone fails: standalone-BKD divergence on GPT-2 is rescued by chaining, and attention-linearization PPL inflation by granular sub-module targeting plus a fine-tune. Together with \S\ref{exp:framework}, these address \textbf{RQ1}--\textbf{RQ3} of \S\ref{sec:experiments}.

\section{Limitations and Conclusion}
\label{sec:limitations_and_conclusion}

SilverSpoon-KD assumes \emph{white-box} access to the teacher---its hidden states and logits, computed live during student training---which rules out \emph{black-box} regimes (where the student trains on teacher-generated text without logits)~\cite{xu2024survey} and cached white-box pipelines that pre-compute teacher outputs offline; a run also currently trains a single student against a single teacher. QAT is supported via externally prepared quantized models, with native per-block precision reserved for future work. Methodological observations the evaluation surfaces (e.g., \gls{bkd}~$\rightarrow$~\gls{hkd} pipelines recovering from standalone-blockwise divergence on autoregressive \glspl{llm}) open natural directions for focused study. Other promising extensions include heterogeneous teacher--student hardware, multi-teacher distillation, black-box pipelines for closed-weight teachers \cite{wang2025easydistill}, teacher-output caching with logit compression, and a declarative (YAML/JSON) configuration layer.

SilverSpoon-KD unifies four \gls{kd} paradigms behind a single API and provides, to our knowledge, the first general-purpose blockwise \gls{kd}; with FSDP-sharded teacher placement, it fits 14B-parameter teachers on 2$\times$A100 where compared toolkits run out of memory. The unified API also made several methodological observations cheap to surface: multi-stage chaining produced essentially every headline result; paradigm--setting interactions partitioned cleanly across compression and quantization regimes; and standalone-blockwise divergence on autoregressive \glspl{llm} was rescued by chaining---each obtained through configuration rather than re-implementation. SilverSpoon-KD enables systematic paradigm comparison that would otherwise require bespoke pipelines.

\begin{ack}
We thank Oscar Eriksson, Shubhra Mishra, Martin Orrje, Viktor Palmkvist, and John Wikman for insightful discussions that informed this work. This project was partially supported by the Digital Futures Research Center and by the Swedish Research Council (Vetenskapsr\aa{}det, grant no.~\mbox{2024-05043}). The computations were enabled by resources provided by the National Academic Infrastructure for Supercomputing in Sweden (NAISS), partially funded by the Swedish Research Council through grant agreement no.~\mbox{2022-06725}.
\end{ack}

\bibliographystyle{unsrtnat}
\bibliography{references}

\clearpage
\appendix

\let\oldappendixsection\section
\renewcommand{\section}{\FloatBarrier\oldappendixsection}

\begin{center}
\Large\bfseries Appendix
\end{center}

\noindent\textbf{Contents}
\begin{itemize}\setlength\itemsep{0pt}
  \item \ref{app:paradigm_visualizations} \hyperref[app:paradigm_visualizations]{Training Paradigm Visualizations}\dotfill\pageref{app:paradigm_visualizations}
    \begin{itemize}\setlength\itemsep{0pt}
      \item Figure~\ref{fig:standard_training} \hyperref[fig:standard_training]{Standard Training}\dotfill\pageref{fig:standard_training}
      \item Figure~\ref{fig:reskd_visualization} \hyperref[fig:reskd_visualization]{Response-Based KD}\dotfill\pageref{fig:reskd_visualization}
      \item Figure~\ref{fig:hkd_visualization} \hyperref[fig:hkd_visualization]{Holistic KD}\dotfill\pageref{fig:hkd_visualization}
      \item Figure~\ref{fig:bkd_visualization} \hyperref[fig:bkd_visualization]{Blockwise KD}\dotfill\pageref{fig:bkd_visualization}
      \item Figure~\ref{fig:multistage_visualization} \hyperref[fig:multistage_visualization]{Multi-Stage Pipeline}\dotfill\pageref{fig:multistage_visualization}
    \end{itemize}
  \item \ref{app:algorithms} \hyperref[app:algorithms]{Full Algorithms}\dotfill\pageref{app:algorithms}
    \begin{itemize}\setlength\itemsep{0pt}
      \item Algorithm~\ref{alg:reskd} \hyperref[alg:reskd]{Response-Based KD}\dotfill\pageref{alg:reskd}
      \item Algorithm~\ref{alg:blockcapture} \hyperref[alg:blockcapture]{Block Capture}\dotfill\pageref{alg:blockcapture}
      \item Algorithm~\ref{alg:bkd} \hyperref[alg:bkd]{Blockwise KD}\dotfill\pageref{alg:bkd}
      \item Algorithm~\ref{alg:hkd} \hyperref[alg:hkd]{Holistic KD}\dotfill\pageref{alg:hkd}
    \end{itemize}
  \item \ref{app:experiments} \hyperref[app:experiments]{Detailed Experimental Results}\dotfill\pageref{app:experiments}
    \begin{itemize}\setlength\itemsep{0pt}
      \item \ref{exp:vision_bkd} \hyperref[exp:vision_bkd]{VGG Compression}\dotfill\pageref{exp:vision_bkd}
      \item \ref{exp:vision_relkd} \hyperref[exp:vision_relkd]{VGG Relational KD}\dotfill\pageref{exp:vision_relkd}
      \item \ref{exp:bert_dist} \hyperref[exp:bert_dist]{BERT Pre-Training}\dotfill\pageref{exp:bert_dist}
      \item \ref{exp:bert_downstream} \hyperref[exp:bert_downstream]{BERT Downstream}\dotfill\pageref{exp:bert_downstream}
      \item \ref{exp:gpt2_dist} \hyperref[exp:gpt2_dist]{GPT-2 Compression}\dotfill\pageref{exp:gpt2_dist}
      \item \ref{exp:gpt2_cross_arch} \hyperref[exp:gpt2_cross_arch]{Cross-Architecture}\dotfill\pageref{exp:gpt2_cross_arch}
      \item \ref{exp:gpt2_linear} \hyperref[exp:gpt2_linear]{GPT-2 Attention Linearization}\dotfill\pageref{exp:gpt2_linear}
      \item \ref{exp:qwen3_quant} \hyperref[exp:qwen3_quant]{Qwen3 Quantization}\dotfill\pageref{exp:qwen3_quant}
      \item \ref{exp:framework_detail} \hyperref[exp:framework_detail]{Toolkit Benchmarking}\dotfill\pageref{exp:framework_detail}
    \end{itemize}
\end{itemize}
\vspace{0.5em}

\section{Training Paradigm Visualizations}
\label{app:paradigm_visualizations}

The following figures illustrate the forward and backward pass data flow for standard training, response-based KD, holistic KD, and blockwise KD, plus a multi-stage pipeline that chains paradigms by initializing each stage from the previous stage's trained student. Black arrows denote the forward pass and are labeled with the block output $h_i^S$ (or $h_i^T$) that they carry, where $h_i^S$ is the output of student block $i$ (equivalent to $\mathbf{h}_{out}^S$ as used in the System Design algorithm boxes). Red arrows denote backpropagation and are labeled with the \emph{upstream activation gradient} $\partial \mathcal{L}/\partial h_i^S$ that they transmit during the chain-rule traversal. A small red annotation inside each student block shows the \emph{parameter gradient} $\partial \mathcal{L}/\partial \theta_i^S$ that is computed locally at that block from the arriving upstream gradient and subsequently used to update the block's weights. Teacher parameters are frozen (no red arrows or in-block annotations).

\input{static/standard_training}

\input{static/reskd_visualization}

\input{static/hkd_visualization}

\input{static/bkd_visualization}

\input{static/multistage_visualization}

\section{Full Algorithms}
\label{app:algorithms}

This appendix gives the full formal algorithms referenced from Section~\ref{sec:system_design}. These algorithms describe three distillation paradigms (each well-established in prior work) as they are exposed by SilverSpoon-KD's unified API; the contribution of the toolkit is not the algorithms themselves but the shared infrastructure---\textsc{CaptureFeatures} (Algorithm~\ref{alg:blockcapture}), learnable projectors, per-alignment optimizers, and the evaluation suite---that makes switching between them a single configuration change.

\begin{algorithm}[H]
\caption{Response-Based Knowledge Distillation (ResKD)}
\label{alg:reskd}
\begin{algorithmic}[1]
    \Input Teacher Model $\mathcal{T}$ with output logits $\mathbf{z}^T = \mathcal{T}(\mathbf{x})$.
    \Input Student Model $\mathcal{S}$ with output logits $\mathbf{z}^S = \mathcal{S}(\mathbf{x})$.
    \Input Training data $\mathcal{X}$, ground-truth labels $\mathbf{y}$.
    \Input Soft loss $\mathcal{L}_{soft}$ (e.g., temperature-scaled KL divergence), hard-label weight $\alpha \in [0, 1]$.
    \Input Optimizer $\text{Opt}$, max gradient norm $g_{max}$.

    \Procedure{TrainResponseBased}{$\mathcal{S}, \mathcal{T}, \mathcal{X}, \alpha$}
        \State Freeze all parameters of $\mathcal{T}$.

        \For{each batch $(\mathbf{x}, \mathbf{y}) \in \mathcal{X}$}
            \State $\mathbf{z}^T \gets \mathcal{T}(\mathbf{x})$ \Comment{Teacher forward pass (no gradient)}
            \State $\mathbf{z}^S, L_{hard} \gets \mathcal{S}(\mathbf{x}, \mathbf{y})$ \Comment{Student forward pass with hard loss}

            \State $L_{soft} \gets \mathcal{L}_{soft}(\mathbf{z}^S, \mathbf{z}^T)$ \Comment{Soft distillation loss}

            \State $L \gets (1 - \alpha) \cdot L_{soft} + \alpha \cdot L_{hard}$ \Comment{$\alpha{=}0$: pure distillation}

            \State \Call{Backpropagate}{$L, \theta^S$}
            \State \Call{ClipGradients}{$\theta^S, g_{max}$}
            \State $\text{Opt}$.\Call{Step}{}
        \EndFor

        \State \Return $\mathcal{S}$
    \EndProcedure
\end{algorithmic}
\end{algorithm}

\begin{algorithm}[H]
\caption{Block Capture Algorithm}
\label{alg:blockcapture}
\begin{algorithmic}[1]
    \Input Model sequence $\mathcal{B} = (b_1, \dots, b_N)$ where $b_i$ implements $f_i(\cdot; \theta_i)$.
    \Input Input tensor $\mathbf{x}$.
    \Input Set $\mathcal{S} \subset \mathcal{B}$ of blocks to capture, optional terminal block $b_{stop}$.
    \Input Boolean $\textit{capture\_inputs}$: whether to record block inputs (used by blockwise KD).
    \Output Dictionary $\mathcal{D}$ mapping blocks to captured tensors.

    \Procedure{CaptureFeatures}{$\mathcal{B}, \mathbf{x}, \mathcal{S}, b_{stop}, \textit{capture\_inputs}$}
        \State $\mathcal{D} \gets \emptyset$
        \State $\mathbf{h} \gets \mathbf{x}$ \Comment{Initialize hidden state with input}

        \For{$i \gets 1$ \textbf{to} $N$}
            \State $b_{curr} \gets b_i$

            \If{$b_{curr} \in \mathcal{S}$ \textbf{and} $\textit{capture\_inputs}$}
                \State $\mathcal{D}[b_{curr}]_{\text{in}} \gets \mathbf{h}$ \Comment{Capture input to function $f_i$}
            \EndIf

            \State $\mathbf{h} \gets f_i(\mathbf{h}; \theta_i)$ \Comment{Apply block function}

            \If{$b_{curr} \in \mathcal{S}$}
                \State $\mathcal{D}[b_{curr}]_{\text{out}} \gets \mathbf{h}$ \Comment{Capture output of function $f_i$}
            \EndIf

            \If{$b_{curr} = b_{stop}$}
                \State \textbf{break} \Comment{Early exit optimization}
            \EndIf
        \EndFor

        \State \Return $\mathcal{D}$
    \EndProcedure
\end{algorithmic}
\end{algorithm}

\begin{algorithm}[H]
\caption{blockwise KD Algorithm}
\label{alg:bkd}
\begin{algorithmic}[1]
    \Input Teacher Model $\mathcal{T}$ with blocks $\{f_j^T\}$.
    \Input Student Model $\mathcal{S}$ with blocks $\{f_i^S\}$.
    \Input Mapping $\mathcal{M} = \{(i, j)\}$ pairing Student block $i$ to Teacher block $j$.
    \Input Output selectors $\sigma_i^T, \sigma_i^S$ per alignment (extract relevant sub-tensors).
    \Input Optional Projector sets $\Phi_{in} = \{\phi_{in, i}\}$ and $\Phi_{out} = \{\phi_{out, i}\}$ for each Student block $i$.
    \Input Training data $\mathcal{X}$, Loss function $\mathcal{L}$, per-alignment Optimizers $\{\text{Opt}_i\}$, max gradient norms $\{g_{max,i}\}$.

    \Procedure{TrainBlockWise}{$\mathcal{S}, \mathcal{T}, \mathcal{M}, \Phi, \mathcal{X}$}
        \State Freeze all parameters of $\mathcal{T}$.
        \State $\mathcal{B}_{req} \gets \{ b_j^T \mid (i, j) \in \mathcal{M} \}$ \Comment{Required Teacher blocks}
        \State $b_{last} \gets \max_{order}(\mathcal{B}_{req})$ \Comment{Last required Teacher block}

        \For{each batch $\mathbf{x} \in \mathcal{X}$}
            \State $\mathcal{D}^T \gets \Call{CaptureFeatures}{\mathcal{T}, \mathbf{x}, \mathcal{B}_{req}, b_{last}, \textbf{true}}$ \Comment{Truncated teacher forward}

            \For{each $(i, j) \in \mathcal{M}$} \Comment{Parallelizable: blocks are independent}
                \State $\mathbf{h}_{in}^T \gets \mathcal{D}^T[b_j^T]_{\text{in}}$  \Comment{Teacher block input (detached)}
                \State $\mathbf{h}_{out}^T \gets \sigma_i^T(\mathcal{D}^T[b_j^T]_{\text{out}})$ \Comment{Select from teacher output}

                \If{$\phi_{in, i} \in \Phi_{in}$} \Comment{Input projection: Teacher dim $\to$ Student dim}
                    \State $\mathbf{h}_{in}^S \gets \phi_{in, i}(\mathbf{h}_{in}^T)$
                \Else
                    \State $\mathbf{h}_{in}^S \gets \mathbf{h}_{in}^T$
                \EndIf

                \State $\mathbf{h}_{out}^S \gets \sigma_i^S(f_i^S(\mathbf{h}_{in}^S; \theta_i^S))$ \Comment{Student forward + select (teacher-forced input)}

                \If{$\phi_{out, i} \in \Phi_{out}$} \Comment{Output projection: Student dim $\to$ Teacher dim}
                    \State $\mathbf{h}_{pred} \gets \phi_{out, i}(\mathbf{h}_{out}^S)$
                \Else
                    \State $\mathbf{h}_{pred} \gets \mathbf{h}_{out}^S$
                \EndIf

                \State $L_i \gets \mathcal{L}(\mathbf{h}_{pred}, \mathbf{h}_{out}^T)$
                \State \Call{Backpropagate}{$L_i, \{\theta_i^S, \phi_{in, i}, \phi_{out, i}\}$} \Comment{Per-block gradient, no cross-block flow}
                \State \Call{ClipGradients}{$\{\theta_i^S, \phi_{in, i}, \phi_{out, i}\}, g_{max,i}$}
                \State $\text{Opt}_i$.\Call{Step}{}
            \EndFor
        \EndFor

        \State \Return $\mathcal{S}, \Phi$
    \EndProcedure
\end{algorithmic}
\end{algorithm}

\begin{algorithm}[H]
\caption{holistic KD Algorithm}
\label{alg:hkd}
\begin{algorithmic}[1]
    \Input Teacher Model $\mathcal{T}$ with blocks $\{f_j^T\}$.
    \Input Student Model $\mathcal{S}$ with blocks $\{f_i^S\}$.
    \Input Mapping $\mathcal{M} = \{(i, j)\}$ pairing Student block $i$ to Teacher block $j$.
    \Input Output selectors $\sigma_i^T, \sigma_i^S$ per alignment.
    \Input Optional Projector set $\Phi_{out} = \{\phi_{out, i}\}$ for each Student block $i$.
    \Input Training data $\mathcal{X}$, Loss function $\mathcal{L}$, per-alignment loss weights $\{w_i\}$.
    \Input Optimizer $\text{Opt}$ over the full student (and projectors $\Phi_{out}$), max gradient norm $g_{max}$.
    \Input Boolean $\textit{magnitude\_aware}$: normalize losses by magnitude before weighting.

    \Procedure{TrainHolistic}{$\mathcal{S}, \mathcal{T}, \mathcal{M}, \Phi_{out}, \mathcal{X}$}
        \State Freeze all parameters of $\mathcal{T}$.
        \State $\mathcal{B}_{req}^T \gets \{ b_j^T \mid (i, j) \in \mathcal{M} \}$ \Comment{Required Teacher blocks}
        \State $\mathcal{B}_{req}^S \gets \{ b_i^S \mid (i, j) \in \mathcal{M} \}$ \Comment{Required Student blocks}

        \For{each batch $\mathbf{x} \in \mathcal{X}$}
            \State $\mathcal{D}^T \gets \Call{CaptureFeatures}{\mathcal{T}, \mathbf{x}, \mathcal{B}_{req}^T, \text{None}, \textbf{false}}$ \Comment{Teacher outputs only (detached)}
            \State $\mathcal{D}^S \gets \Call{CaptureFeatures}{\mathcal{S}, \mathbf{x}, \mathcal{B}_{req}^S, \text{None}, \textbf{false}}$ \Comment{Student outputs only (gradients retained)}

            \State $L_{total} \gets 0$
            \For{each $(i, j) \in \mathcal{M}$}
                \State $\mathbf{h}_{out}^T \gets \sigma_i^T(\mathcal{D}^T[b_j^T]_{\text{out}})$ \Comment{Select from teacher output (detached)}
                \State $\mathbf{h}_{out}^S \gets \sigma_i^S(\mathcal{D}^S[b_i^S]_{\text{out}})$ \Comment{Select from student output (with gradient)}

                \If{$\phi_{out, i} \in \Phi_{out}$} \Comment{Output projection: Student dim $\to$ Teacher dim}
                    \State $\mathbf{h}_{pred} \gets \phi_{out, i}(\mathbf{h}_{out}^S)$
                \Else
                    \State $\mathbf{h}_{pred} \gets \mathbf{h}_{out}^S$
                \EndIf

                \State $L_i \gets \mathcal{L}(\mathbf{h}_{pred}, \mathbf{h}_{out}^T)$
                \If{$\textit{magnitude\_aware}$} \Comment{Normalize so gradients $\propto w_i$}
                    \State $L_{total} \gets L_{total} + w_i \cdot L_i \,/\, |L_i|_{\text{detach}}$
                \Else
                    \State $L_{total} \gets L_{total} + w_i \cdot L_i$
                \EndIf
            \EndFor

            \State \Call{Backpropagate}{$L_{total}, \{\theta^S, \Phi_{out}\}$} \Comment{End-to-end gradient flow}
            \State \Call{ClipGradients}{$\{\theta^S, \Phi_{out}\}, g_{max}$} \Comment{Single global gradient clip}
            \State $\text{Opt}$.\Call{Step}{} \Comment{Single optimizer step over the full student}
        \EndFor

        \State \Return $\mathcal{S}, \Phi_{out}$
    \EndProcedure
\end{algorithmic}
\end{algorithm}

\section{Detailed Experimental Results}
\label{app:experiments}

This appendix presents the full per-experiment results for the eight distillation-quality experiments summarized in \S\ref{exp:summary}. Each subsection corresponds to one experiment and contains the experimental setup, the full results table, and a discussion of the findings. Subsection ordering matches the canonical ordering of the companion experiments repository and the experimental coverage matrix (Figure~\ref{fig:experiment_coverage}). Throughout, model and dataset identifiers in monospace (e.g., \texttt{Qwen/Qwen3-4B}, \texttt{timm/vgg16.tv\_in1k}, \texttt{allenai/dolma3\_pool}) are HuggingFace Hub paths.

\subsection{VGG Compression}
\label{exp:vision_bkd}

This experiment validates that SilverSpoon-KD supports blockwise KD and works beyond transformer workflows by distilling a VGG-16 teacher (\texttt{timm/vgg16.tv\_in1k}) into a VGG-16-DS student that replaces standard convolutions with depthwise-separable convolutions on CIFAR-10, replicating the setup of \cite{wang2018progressive}, one of the foundational papers for blockwise KD. Twelve cross-depth alignment points span the internal convolutional layers; SilverSpoon-KD automatically inserts input and output projectors to bridge channel-dimension mismatches, and BKD additionally uses \texttt{copy\_matching} initialization to seed the student with the teacher's classifier head and first convolution (frozen). We compare single-paradigm runs (blockwise KD; response-based KD with temperature $T{=}2$ and soft/hard mixing $\alpha{=}0.5$) and three two-stage pipelines (blockwise KD~$\rightarrow$~fine-tune, blockwise KD~$\rightarrow$~holistic KD, blockwise KD~$\rightarrow$~response-based KD) against a no-teacher baseline and the VGG-16 teacher ceiling. All runs use AdamW (LR $10^{-3}$, cosine schedule with 5\% warmup, weight decay 0, max grad norm 1.0), batch size 128, and float32 precision; single-stage and stage-2 distillation runs train 100 epochs while stage-2 fine-tuning runs 60 epochs.

\input{generated/vision_bkd}

Multi-stage chaining delivers gains beyond the teacher ceiling: the best result, blockwise KD~$\rightarrow$~response-based KD (90.42\%), exceeds the VGG-16 teacher (87.26\%) by 3.16pp and the no-teacher baseline (82.69\%) by 7.73pp. Standalone blockwise KD dips below baseline (72.49\%), but the same blockwise initialization followed by fine-tuning (88.92\%), holistic KD (87.97\%), or response-based KD (90.42\%) all yield strong final students---each chain recovers more than 15pp over standalone blockwise KD and outperforms standalone response-based KD (84.35\%). Blockwise alignment behaves here as a useful initialization rather than a standalone training procedure, validating SilverSpoon-KD's multi-stage pipeline as the central mechanism for harnessing blockwise KD in settings where it diverges in isolation. The result also demonstrates that the block abstraction generalizes beyond transformers (depthwise-separable CNNs are accommodated transparently) and that automatic projectors bridge the resulting channel-dimension mismatches without manual configuration.

\subsection{VGG Relational KD}
\label{exp:vision_relkd}

We replicate the setup of Park et al.\ \cite{park2019relational} to validate SilverSpoon-KD's relation-based KD support: a ResNet-50 teacher (77.81\% on our CIFAR-100 split) and a VGG-11-BN student. All runs train for 200 epochs at batch size 128 in float32 with SGD (momentum 0.9, weight decay $5\!\times\!10^{-4}$), starting from learning rate $0.1$ and decaying $\times 0.2$ at epochs 60, 120, and 160. We compare response-based KD (Hinton KD, temperature $T{=}16$ with hard-label weight $\alpha\!\approx\!0.06$ to match Hinton's $\lambda{=}16$ soft/hard mixing), holistic KD with MSE on global-pool features, and three relation-based KD configurations---distance loss, angle loss, and their combination---against Park et al.'s reported numbers and a no-teacher baseline. Feature alignment is placed on the global-pool layer (2048-d for ResNet-50, 512-d for VGG-11-BN); HKD with MSE bridges this dimension mismatch via SilverSpoon-KD's auto-projectors, while the relation-based variants disable auto-projection because their losses (pairwise distances, triplet angles) are computed within each side and compared structurally, requiring no pointwise dimension matching. The relation-based KD variants are followed by a CE fine-tune of the classifier head, since the holistic KD stage trains feature layers only and leaves the head randomly initialized. This experiment exercises SilverSpoon-KD's pluggable loss functions: switching between distance, angle, and combined relation-based KD losses requires only a configuration change.

\input{generated/relkd}

All distillation variants improve substantially over the no-teacher baseline (65.16\%), with response-based KD leading at 73.65\%---within 0.61pp of Park et al.'s reported 74.26\%, despite our weaker baseline (65.16\% vs.\ 71.26\%). Holistic KD with MSE on global-pool features reaches 72.78\%, between response-based KD and the relation-based variants. The three relation-based KD configurations (distance, angle, distance+angle) range from 70.40\% to 71.61\%, modestly trailing both response-based KD and Park et al.'s reported relation-based KD numbers (72.27--72.97\%) but confirming relation-based KD's capacity to transfer relational structure across the cross-architecture (ResNet-50~$\rightarrow$~VGG-11-BN) gap. Beyond the headline numbers, this experiment exercises SilverSpoon-KD's pluggable loss-function architecture: switching between distance, angle, and combined relation-based KD losses requires only a configuration change, with no per-loss training-loop modifications.

\subsection{BERT Pre-Training}
\label{exp:bert_dist}

We distill BERT-base-uncased (12 layers, 768 hidden, 110M parameters) during masked language model pre-training on Dolma~\cite{soldaini2024dolma} (\texttt{allenai/dolma3\_pool}) into two students with different compression regimes: \texttt{bert\_T6} (6 layers, hidden 768, intermediate 3072, ${\approx}65$M parameters, ${\approx}59\%$ of teacher) and \texttt{bert\_T4\_tiny} (4 layers, hidden 312, intermediate 1200, ${\approx}14$M parameters, ${\approx}13\%$). Both students are constructed by pruning BERT-base layers (T6 keeps layers $\{1,3,5,7,9,11\}$, T4-tiny keeps $\{2,5,8,11\}$ and additionally narrows the hidden dimension), and cross-depth alignment maps each retained student layer to its pre-pruning teacher counterpart. T4-tiny's narrower hidden dimension is bridged automatically by SilverSpoon-KD's auto-projectors at every alignment point; T6 shares the teacher's hidden size so no projector is needed. Stage-1 distillation runs for 20{,}000 steps in float32 at batch size 16 with sequence length 512, using AdamW (LR $2\!\times\!10^{-4}$, constant-with-warmup schedule, 10\% warmup, weight decay 0, max grad norm 1.0). Every distilled checkpoint is then fine-tuned on MNLI in Stage 2 for 30 epochs at batch size 128 (AdamW, LR $1\!\times\!10^{-4}$, linear schedule, 10\% warmup, weight decay 0.01, max grad norm 1.0) to measure downstream transfer. Beyond demonstrating SilverSpoon-KD's support for encoder pre-training with cross-depth alignment, this experiment surfaces a paradigm--compression interaction discussed below.

All three paradigms improve MNLI over the no-teacher baseline at T6 compression (${\approx}59\%$ of teacher parameters), and holistic KD leads (80.33\%, +3.04pp over baseline 77.29\%), followed by blockwise KD (79.13\%, +1.84pp) and response-based KD (79.01\%, +1.72pp). At the more aggressive T4-tiny compression (${\approx}13\%$), holistic KD's lead widens substantially: 70.44\% (+3.52pp over baseline 66.92\%) while standalone blockwise KD actually hurts ($-$0.82pp) and response-based KD provides only modest gains (+0.91pp). The same paradigms that cluster within ${\approx}1.3$pp of each other at T6 fan out to a 5.2pp spread at T4-tiny: as the teacher--student structural gap widens, end-to-end alignment (HKD) becomes increasingly indispensable while strict per-block matching (BKD) degrades into actively harmful supervision. The Stage-1 masked language modeling perplexity columns reflect the same gradient---both BKD and ResKD diverge to four-digit perplexities at T4-tiny while HKD's Stage-1 PPL stays substantially closer to the no-teacher baseline. This compression--paradigm interaction is exactly the kind of empirical finding SilverSpoon-KD's unified API is designed to surface, since switching paradigms across compression regimes requires only a class swap (Figure~\ref{fig:api_snippet}) rather than a re-implementation of the training loop.

\input{generated/bert_pretrain}

\subsection{BERT Downstream}
\label{exp:bert_downstream}

We directly compare SilverSpoon-KD against TextBrewer on MNLI, one of TextBrewer's primary benchmarks~\cite{yang2020textbrewer}. The teacher is BERT-base-cased fine-tuned on MNLI; the two students are again T6 (${\approx}65$M params) and T4-tiny (${\approx}14$M params), constructed by pruning teacher layers (T6: $\{1,3,5,7,9,11\}$; T4-tiny: $\{2,5,8,11\}$ with hidden-dim narrowing). All runs train for 30 epochs at batch size 128 with AdamW (LR $1\!\times\!10^{-4}$, linear schedule, 10\% warmup, weight decay 0.01, max grad norm 1.0); TextBrewer additionally uses early stopping with patience 5, while SilverSpoon-KD's runs train for the full epoch budget. TextBrewer's \texttt{GeneralDistiller} uses \texttt{hidden\_mse} for per-layer intermediate alignment and CE-soft KD on logits at temperature $T{=}8$, with linear projectors where T4-tiny's hidden dimension differs from the teacher's. SilverSpoon-KD's holistic KD is configured as a single-stage run combining hidden-state alignment with a response-based loss at $T{=}8$ (response weight $1/T^2$ to cancel SilverSpoon-KD's internal $T^2$ scaling so the effective KD weight matches TextBrewer's); SilverSpoon-KD's blockwise KD is a two-stage pipeline---per-block alignment, then a standard fine-tune at LR $2\!\times\!10^{-5}$. Because TextBrewer cannot express blockwise KD, this experiment tests how a paradigm absent from existing toolkits fares against holistic methods, particularly under aggressive compression, on TextBrewer's own turf---the only direct head-to-head comparison against an established encoder-distillation toolkit in this paper.

On the T6 student (${\approx}59\%$ of teacher parameters), SilverSpoon-KD's holistic KD reaches 83.47\% MNLI-m---outperforming TextBrewer T6 (80.09\%) by 3.38pp and matching the BERT-base teacher (83.59\%) within 0.12pp---on TextBrewer's own benchmark and against a toolkit specifically optimized for the encoder-NLP domain. At T4-tiny (${\approx}13\%$), TextBrewer narrowly leads SilverSpoon-KD's holistic KD (81.20\% vs.\ 80.87\%), reflecting its mature encoder-specific tuning at extreme compression, while SilverSpoon-KD's blockwise KD~$\rightarrow$~fine-tune pipeline reaches only 66.01\%, confirming that blockwise alignment alone struggles when the structural gap is too wide---consistent with the pre-training findings of Appendix~\ref{exp:bert_dist}. SilverSpoon-KD additionally enables blockwise KD on encoder tasks, a paradigm unavailable in TextBrewer; while blockwise KD does not outperform holistic KD here, its availability widens the experimental design space practitioners can explore.

\input{generated/bert_downstream}

\subsection{GPT-2 Compression}
\label{exp:gpt2_dist}

This experiment highlights multi-stage pipelines on a causal language model. The teacher is GPT-2 Small (124M, 12 layers, hidden 768, \texttt{n\_inner}~3072); the student is a GPT-2 variant with reduced MLP width (96M parameters, 12 layers, hidden 768, \texttt{n\_inner}~1536---a 50\% MLP-width reduction), keeping the teacher's hidden size so no projectors are required at any cross-depth alignment. All runs train on Dolma at context length 1024 with AdamW (LR $1\!\times\!10^{-3}$) in bf16, with step budgets sized to keep training tokens approximately constant across paradigms: standalone blockwise KD and the no-teacher baseline run 60{,}000 steps at batch size 32 (${\approx}1.97$B tokens), while standalone holistic KD, response-based KD, and the BKD~$\rightarrow$~HKD continuation run 117{,}000 steps at batch size 16 (${\approx}1.92$B tokens). The final LoRA stage of the three-stage pipeline runs 60{,}000 steps at batch size 32 with rank 64, $\alpha{=}128$, and dropout 0.05 over the Conv1D projection modules (\texttt{c\_attn}, \texttt{c\_proj}, \texttt{c\_fc}). Multi-stage pipelines chain paradigms by initializing each stage from the previous stage's trained student checkpoint. GPT-2 Small was chosen for its combination of historical importance and small size, which kept the many iterations needed to approach a Chinchilla-style token-to-parameter ratio~\cite{hoffmann2022training} computationally tractable.

\input{generated/gpt2_multistage}

Standalone blockwise KD catastrophically diverges (PPL~245{,}281, 0\% LAMBADA), confirming that blockwise alignment in isolation is unstable for autoregressive language models that lack an end-to-end generation signal. Multi-stage chaining---which initializes each stage from the previous stage's trained student---rescues this failure mode entirely: blockwise KD~$\rightarrow$~holistic KD reaches PPL~109.56 (within 12 PPL of standalone holistic KD at 97.69), and the three-stage blockwise KD~$\rightarrow$~holistic KD~$\rightarrow$~LoRA pipeline becomes the strongest configuration overall (PPL~93.39, LAMBADA~19.56\%---a 28\% relative improvement over baseline LAMBADA at 15.25\%). This BKD-divergence-then-recovery sequence is the central result for SilverSpoon-KD's multi-stage capability and, to our knowledge, the first open report of blockwise KD~$\rightarrow$~holistic KD chaining recovering from standalone-blockwise KD divergence on autoregressive LLMs. Standalone response-based KD reduces perplexity (133.99) but degrades LAMBADA from 15.25\% to 6.48\%, suggesting it transfers fluency without the task-specific knowledge that hidden-state alignment captures.

Ultimately, this experiment is one of the strongest demonstrations of SilverSpoon-KD's multi-stage pipeline mechanism. The blockwise KD-alone divergence and subsequent multi-stage recovery is a result that was obtained entirely within the SilverSpoon-KD library, demonstrating a capability and outcome unique to this KD toolkit.

\subsection{Cross-Architecture}
\label{exp:gpt2_cross_arch}

This experiment pushes cross-architecture distillation within the context of \gls{featkd} to its limit: teacher and student share \emph{no} architectural lineage. The teacher is GPT-2 Small (124M, 12 layers, standard multi-head attention, learned positional embeddings, GELU activations); the student is a 96M-parameter model built on the DeepSeek-V3 architecture with Multi-head Latent Attention (\texttt{kv\_lora\_rank}~96, \texttt{q\_lora\_rank}~256), RoPE positional embeddings, SiLU activations, and dense (non-mixture-of-experts) MLPs. The student has 6 layers mapped to teacher layers $\{0, 2, 4, 6, 8, 10\}$ via cross-depth alignment. Hidden size is intentionally shared (768) to isolate the architectural variable from projector effects, so all alignments are projector-free and the only difference between this and the same-family GPT-2 compression setting (Appendix~\ref{exp:gpt2_dist}) is the layer architecture itself. Standalone-paradigm runs (HKD, ResKD) and the no-teacher baseline train for 60{,}000 steps at batch size 32 on Dolma at context length 1024 with AdamW (LR $1\!\times\!10^{-3}$) in bf16; stage-2 fine-tunes from HKD and ResKD start from those checkpoints and continue for 60{,}000 steps at LR $5\!\times\!10^{-4}$ so the student can shake off any architecture-incompatible representations imposed by stage 1. Cross-architecture HKD additionally bumps the response-loss weight from the default (${\approx}2$ for 6 alignments) to 30 and uses temperature 4: the 6 feature-level MSE alignments otherwise dominate the single response alignment 3:1 and force the student's hidden states into a GPT-2-shaped representation the DeepSeek-V3 block cannot naturally produce, so the optimization is shifted toward output-distribution matching while feature alignment becomes a weaker regularizer. We attempted blockwise KD as well, but per-block alignment is not viable in this regime: teacher-forced GPT-2 hidden-state inputs land in a representation space (MHA $+$ learned positional encoding) that the DeepSeek-V3 block (MLA $+$ RoPE) cannot consume compatibly, and the run did not yield reportable results.

\input{generated/gpt2_deepseek}

Cross-architecture distillation between teacher and student that share no architectural lineage (multi-head vs.\ multi-head latent attention, learned vs.\ RoPE positional embeddings, GELU vs.\ SiLU) is feasible but more delicate than the same-family setting. Standalone holistic KD diverges (PPL~246.44 vs.\ baseline~117.78, with LAMBADA collapsing from 16.34\% to 0.87\%) even with the rebalanced response-loss weight, and standalone response-based KD also degrades LAMBADA (16.34\%~$\to$~6.02\%) despite a smaller perplexity hit. Multi-stage recovery via fine-tuning brings both paradigms back to near-baseline: holistic KD~$\rightarrow$~FT reaches PPL~118.49 with LAMBADA~15.06\%, and response-based KD~$\rightarrow$~FT reaches PPL~117.28 with LAMBADA~14.96\%. As described above, blockwise KD is not reported because its teacher-forced inputs land in a hidden-state space the DeepSeek-V3 block cannot consume compatibly---a hard limit of feature-based KD when per-block input geometry diverges, one that no projector or alignment weight resolves at the per-block level. While the absolute gains over baseline are smaller than in the same-family GPT-2 setting (Appendix~\ref{exp:gpt2_dist}), the experiment establishes that SilverSpoon-KD's cross-architecture layer-mapping and multi-stage chaining together make HKD- and ResKD-driven cross-family distillation tractable across fundamentally different attention mechanisms, positional encodings, and activation functions---a setting where the toolkit's flexible alignment and multi-stage pipeline are jointly load-bearing.

This experiment extends SilverSpoon-KD's applicability beyond model compression to cross-family knowledge transfer. It demonstrates that distillation has the potential to be viable even between fundamentally different architectures and outside of pure response-based KD, relevant for distilling from proprietary architectures into customized deployment models.

\subsection{GPT-2 Attention Linearization}
\label{exp:gpt2_linear}

This experiment distills standard softmax attention into a linearized-attention variant of the same model family (LoLCATs~\cite{zhang2024lolcats}), a cross-architecture scenario that builds on prior softmax-to-subquadratic distillation work~\cite{bick2024transformers, haller2025empirical}. The teacher is GPT-2 Small (124M, 12 layers, softmax attention); the student replaces each layer's softmax attention with LoLCATs' hybrid attention---sliding-window softmax over the most recent 64 tokens combined with low-rank linear attention over older tokens---while all other weights are copied from the teacher and frozen (\texttt{freeze\_copied=true}), so only the new attention modules are trained. All runs train on Dolma at context length 1024 for 76{,}000 steps with AdamW in bf16. Standalone HKD and ResKD use batch size 16 with two-step gradient accumulation (effective batch 32) due to memory pressure from caching hidden states across all 12 layers; BKD, baseline, and the stage-2 fine-tune use batch size 32 directly. Stage-1 distillation runs use LR $1\!\times\!10^{-2}$ (the higher rate reflects training only the freshly initialized attention modules); stage-2 fine-tune and HKD/ResKD continuations use LR $1\!\times\!10^{-3}$. The benefits of attention linearization tend to come at much larger context lengths and model sizes than tested here (given the efficiency of quadratic attention kernels at the time of writing), but this experiment serves as a proof of application of SilverSpoon-KD to the domain. Beyond full-layer blockwise KD, we exercise SilverSpoon-KD's \emph{granular sub-module targeting}: a blockwise KD variant whose alignment pattern (\texttt{\textasciicircum transformer.h.\textbackslash{}d+.attn\$}) targets only the attention sub-module within each layer rather than the full layer output, isolating the swap-in linearization to its own alignment loss.

\input{generated/gpt2_linear}

In this linearization setting---where the student replaces softmax attention with a linearized variant while retaining the teacher's other weights---distillation alone consistently degrades perplexity relative to the LoLCATs no-KD baseline (53.49). Standalone blockwise KD, holistic KD, and response-based KD all push PPL above 230, and even multi-stage chains without fine-tuning (blockwise KD~$\rightarrow$~holistic KD, blockwise KD~$\rightarrow$~response-based KD) remain in the 250--290 range. Only blockwise KD~$\rightarrow$~fine-tune produces a net gain (PPL~52.01), narrowly beating the no-KD baseline. The setting nonetheless exercises a unique SilverSpoon-KD capability: comparable toolkits would require manual hook plumbing for the attention-only alignment used here, while SilverSpoon-KD exposes it through the same alignment configuration as full-layer blockwise KD---making the distinction one of pattern syntax rather than custom code, and showing that the toolkit is at least usable for---if not yet decisive in---the practically motivated path from softmax to $O(n)$ attention.

\subsection{Qwen3 Quantization}
\label{exp:qwen3_quant}

This experiment combines quantization-aware training (QAT) with \gls{kd} for a practical compression scenario. The teacher is Qwen3-1.7B~\cite{yang2025qwen3} at bf16 full precision; the student is the same architecture loaded from the same pretrained weights but with simulated low-precision integer weights through the straight-through estimator (STE). All runs train on Dolma for 20{,}000 steps with the 8-bit AdamW optimizer at LR $2\!\times\!10^{-5}$ on a cosine schedule. The QAT baseline and blockwise KD use batch size 4 directly, while holistic KD uses batch size 2 with four-step gradient accumulation (effective batch 8) plus gradient checkpointing, and response-based KD uses batch size 2 with gradient checkpointing---both due to additional activation memory beyond the per-block-only gradients of BKD. Holistic KD additionally bumps the response-loss weight to 30 (mirroring the cross-architecture rebalance of Appendix~\ref{exp:gpt2_cross_arch}), and response-based KD uses pure soft distillation ($\alpha{=}0$, $T{=}1$) routed through a fused KD loss kernel from the Liger Kernel library~\cite{hsu2024liger} on \texttt{lm\_head}. We sweep INT4, INT5, and INT6: INT6 post-training quantization (PTQ) is already nearly lossless and serves as a practical upper bound for the sweep, while sub-INT6 precisions are becoming end-to-end viable as a new generation of kernels and hardware support matures---FlexQ~\cite{zhang2025flexq}, for instance, delivers real memory and throughput gains for INT6 LLM serving, alongside emerging FP4 and MXFP microscaling formats. Simulated quantization at these bit-widths is therefore no longer purely academic: weights trained at these precisions can actually be deployed at them. The key question is how much accuracy each distillation paradigm recovers at each precision relative to its PTQ floor and the bf16 ceiling.

\input{generated/qwen3_qat}

Distillation paradigms recover accuracy beyond QAT alone most clearly at the most aggressive precision: at INT4, holistic KD lifts MMLU from QAT-alone 41.50\% to 48.26\% (+6.76pp toward the bf16 ceiling of 55.55\%) and simultaneously delivers the lowest perplexity (29.98 vs.\ 36.15 for QAT alone), while response-based KD lands in between on accuracy (43.75\% MMLU, 31.11 PPL) and blockwise KD trails (42.67\% MMLU, 33.70 PPL). At INT5, the three paradigms converge---all reach 51.99--54.96\% MMLU within the noise of QAT-alone (53.27\%)---and response-based KD edges out the others on perplexity (21.96 vs.\ 22.09 HKD, 22.39 BKD). At INT6, where PTQ is already nearly lossless (54.66\% MMLU, 21.26 PPL), distillation contributes little, confirming INT6 as a practical upper bound for sweep-relevant quantization. Beyond accuracy recovery, the experiment exercises a unique SilverSpoon-KD differentiator: blockwise KD provides distillation at the same GPU memory cost as the QAT baseline (29\,144~MB at INT4) because per-block gradients eliminate the need to materialize hidden states across the full network simultaneously, whereas holistic KD adds ${\approx}6$\,GB and response-based KD adds ${\approx}3$\,GB. For QAT on large models where GPU memory is the binding constraint, blockwise KD's zero-overhead property is a unique practical advantage.

This experiment validates SilverSpoon-KD in a practical production scenario (INT4 quantization of a 1.7B-parameter model) and reveals clear paradigm-specific trade-offs---holistic KD for the largest accuracy gains at the most aggressive precision (INT4), response-based KD for slightly better perplexity at INT5, and blockwise KD wherever GPU memory is the binding constraint---that directly inform practitioner guidance for QAT-distillation pipelines. The setup also leaves a natural path open to non-uniform quantization research~\cite{kim2023squeezellm}, where SilverSpoon-KD's per-alignment configuration could compose with mixed-precision per-layer quantization.

\subsection{Toolkit Benchmarking}
\label{exp:framework_detail}

This appendix documents the per-toolkit configurations behind the head-to-head benchmark (Table~\ref{tab:framework}) and the capability-frontier sweep (Table~\ref{tab:framework_capability}) of \S\ref{exp:framework}. Each comparison is driven by a self-contained benchmarking script under \texttt{scripts/toolkit\_comparison/} in the experiments repository that runs both toolkits against an identical teacher checkpoint, dataset, batch size, step count, and learning-rate schedule, with toolkit-specific defaults preserved wherever they are publicly documented---each setting therefore reflects the configuration a typical user of the comparison toolkit would write. Wall time, throughput, and peak GPU memory are recorded under \texttt{torch.cuda.max\_memory\_allocated} on a freshly initialized CUDA context per run.

\paragraph{Vision DDP (torchdistill, ResNet-50$\rightarrow$ResNet-18 / ImageNet~\cite{deng2009imagenet}, ResKD, 2$\times$A40).} Teacher and student are loaded from \texttt{timm/resnet50.a1\_in1k} and \texttt{timm/resnet18.a1\_in1k} respectively; both toolkits use AdamW (LR $2\!\times\!10^{-4}$, batch size 32 per GPU, fp32, 2{,}000 steps, 10\% linear warmup). KD is response-based at temperature $T{=}2$ with soft/hard mixing $\alpha{=}0.5$. Both runs use PyTorch DDP across two A40s, matching torchdistill's recommended multi-GPU configuration.

\paragraph{Encoder NLP, ResKD (TextBrewer, BERT-base$\rightarrow$T6 / MNLI, 2$\times$A40).} The teacher is our fine-tuned \texttt{bert-base-cased} MNLI checkpoint and the student is a 6-layer T6 initialized by copying teacher layers $\{1,3,5,7,9,11\}$. Both toolkits use AdamW (LR $1\!\times\!10^{-4}$, batch 32 per GPU, max sequence length 128, 1{,}000 steps, 10\% linear warmup) and response-based KD at temperature $T{=}4$. TextBrewer uses its \texttt{GeneralDistiller} with soft-CE; SilverSpoon-KD uses its response-based distiller with KL-divergence loss.

\paragraph{Encoder NLP, HKD (TextBrewer, BERT-base$\rightarrow$T6 / MNLI, 1$\times$A40).} Same teacher, student, and dataset as the ResKD setting; this configuration runs single-GPU and adds hidden-state alignment via TextBrewer's \texttt{hidden\_mse} on every layer pairing alongside response-based KD on logits at temperature $T{=}4$. SilverSpoon-KD's HKD configuration matches: per-layer MSE on aligned hidden states, response loss with $T{=}4$, and the response weight set to $1/T^2$ to cancel SilverSpoon-KD's internal $T^2$ scaling so the effective KD weight matches TextBrewer's.

\paragraph{Decoder LLM, HKD (DistillKit, Qwen3-4B$\rightarrow$0.6B / Dolma, 1$\times$A100).} Both toolkits load \texttt{Qwen/Qwen3-4B} as teacher and \texttt{Qwen/Qwen3-0.6B} as student in bf16 with gradient checkpointing. Training: 500 steps, batch size 2, sequence length 1024, AdamW with cosine schedule (LR $1\!\times\!10^{-3}$). Both combine hidden-state alignment with response-based KD ($w_{\text{KL}}{=}0.5$, $w_{\text{CE}}{=}0.5$, $T{=}2$). DistillKit's HKD path uses linear projectors where dimensions mismatch (\texttt{force\_hidden\_state\_projection}); SilverSpoon-KD's auto-projectors handle the same alignment. SilverSpoon-KD additionally routes the response loss through Liger Kernel's fused KD loss, while DistillKit uses an unfused implementation.

\paragraph{Decoder LLM, ResKD+LoRA (torchtune, Qwen3-4B$\rightarrow$0.6B / Dolma, 1$\times$A100).} Same teacher and student as the DistillKit comparison, in bf16. Training: 1{,}000 steps, batch size 2, sequence length 1024, AdamW (LR $1\!\times\!10^{-3}$). The student is wrapped in LoRA adapters on the q/v/output projections and MLP modules (rank 64, $\alpha{=}128$, dropout 0.05); torchtune trains only the adapters via its \texttt{ForwardKLLoss} with \texttt{kd\_ratio}~$0.5$, while SilverSpoon-KD applies the same LoRA configuration through its standard adapter integration with response-based KD. torchtune's \texttt{ForwardKLLoss} does not expose a temperature parameter; SilverSpoon-KD's reported numbers correspond to its toolkit-default $T{=}2$, $\alpha{=}0.5$ to keep the comparison representative of typical use rather than artificially matched.

\paragraph{Capability frontier (Qwen3-\{4B, 8B, 14B\}$\rightarrow$0.6B / Dolma, 2$\times$A100).} For each (teacher, toolkit) pair the benchmarking script binary-searches the maximum batch size that fits on 2$\times$A100 (40~GB per-GPU) at sequence length 1024, then measures throughput at that batch size in bf16 with AdamW. SilverSpoon-KD is configured with FSDP-sharded teacher placement (\texttt{teacher\_placement="fsdp"}), splitting the 14B teacher's ${\sim}28$~GB of bf16 weights across the two GPUs at ${\sim}14$~GB per GPU; DistillKit and TextBrewer support only replicated teachers (the full ${\sim}28$~GB on each GPU), which leaves no headroom for student parameters, optimizer states, gradients, and activations on a 40~GB card and forces DistillKit to OOM and TextBrewer to bs=1. Reported peak memory and throughput are averaged over the post-warmup measurement window; incremental JSON saves between (toolkit, teacher) pairs let preempted jobs retain partial measurements.

\paragraph{Additional placement sweeps (not reported in \S\ref{exp:framework}).} The benchmarking driver supports three further sweeps beyond the head-to-head and teacher-scaling settings, exercised during development but kept out of the headline tables to preserve focus. (1)~A teacher-placement comparison on 4$\times$A40 contrasts replicated, FSDP-sharded, and split-GPU placement (teacher on GPUs 2--3, student DDP on GPUs 0--1) at fixed batch size; the FSDP-sharded and split-GPU variants are unavailable in DistillKit and TextBrewer. (2)~An OOM-boundary probe extends the batch-size search at sequence lengths $\{1024, 2048, 4096\}$ on 2$\times$A40, surfacing each toolkit's per-placement maximum batch. (3)~A FSDP-vs.-tensor-parallel probe on 4$\times$A40 substitutes tensor-parallel teacher inference for FSDP all-gather, trading a per-step weight gather for activation-communication cost; the trade-off is teacher-size-dependent. All three are reproducible from the experiments repository.

The configurations above establish that the headline numbers in Table~\ref{tab:framework} reflect each comparison toolkit's default or recommended setup rather than a stripped-down version, and that the capability-frontier discrepancy at Qwen3-14B (Table~\ref{tab:framework_capability}) is driven by SilverSpoon-KD's FSDP-sharded teacher placement rather than incidental engineering differences---the only configurable axis on which the compared toolkits genuinely diverge at scale.

\end{document}

%% file: static/kd_taxonomy.tex
\begin{figure}[!tp]
\centering
\resizebox{\linewidth}{!}{%
\begin{tikzpicture}[
  >=Stealth,
  every node/.style={font=\sffamily\footnotesize},
  panel/.style={draw, thick, rounded corners=4pt, fill=gray!2},
  panellbl/.style={font=\sffamily\bfseries\small, anchor=north west, inner sep=3pt},
  tblock/.style={draw, thick, fill=gray!28, rounded corners=0.5pt, inner sep=0pt},
  sblock/.style={draw, thick, fill=cStudent!30, rounded corners=0.5pt, inner sep=0pt},
  ssmblock/.style={draw, thick, fill=cStudent!30, rounded corners=0.9mm, inner sep=0pt},
  selfblock/.style={draw, thick, fill=gray!18, rounded corners=0.5pt, inner sep=0pt},
  tinybf/.style={font=\sffamily\scriptsize\bfseries},
  microf/.style={font=\sffamily\fontsize{5}{6}\selectfont},
  microbf/.style={font=\sffamily\fontsize{5}{6}\selectfont\bfseries},
  microit/.style={font=\sffamily\fontsize{6}{7}\selectfont\itshape},
  flowarr/.style={->, semithick, blue!55!black},
  alignH/.style={<->, thick, dashed, cHkd!85!black},
  alignB/.style={<->, thick, dashed, cBkd!85!black},
  detach/.style={->, thin, dotted, gray!55!black},
  bp/.style={->, thick, dashed, red!60!black},
]

\node[panel, minimum width=4.4cm, minimum height=3.6cm] (paneA) at (0,0) {};
\node[panellbl] at (paneA.north west) {(a) What KD does};

\foreach \i in {1,...,5} {
  \node[tblock, minimum width=10mm, minimum height=2.5mm]
    at (-1.20, -0.75 + \i*0.25) {};
}
\node[tinybf] at (-1.20, 0.95) {Teacher};
\node[microf, gray!70!black] at (-1.20, -0.90) {(inference, larger)};

\foreach \i in {1,...,3} {
  \node[sblock, minimum width=7mm, minimum height=2.5mm]
    at (1.20, -0.55 + \i*0.25) {};
}
\node[tinybf] at (1.20, 0.95) {Student};
\node[microf, gray!70!black] at (1.20, -0.60) {(trained, smaller)};

\draw[->, ultra thick, blue!55!black] (-0.5, 0.0) -- (0.5, 0.0);
\node[font=\sffamily\scriptsize\itshape, blue!55!black] at (-0.0, 0.25) {knowledge};

\node[font=\sffamily\scriptsize, text width=4.05cm, align=center, anchor=north]
  at (-0.0, -1.05)
  {transferred via logits, intermediate features, or sample relations};

\node[panel, minimum width=2.6cm, minimum height=3.6cm] (paneM) at (3.85, 0) {};
\node[panellbl] at (paneM.north west) {(b) Workloads};

\node[microbf] at (3.85, 1.25) {Vision};
\foreach \r in {0,1,2} {
  \foreach \c in {0,1,2} {
    \node[draw, fill=gray!22, minimum width=1.6mm, minimum height=1.6mm,
          inner sep=0pt, line width=0.3pt]
      at (3.65 + \c*0.20, 1.00 - \r*0.20) {};
  }
}

\node[microbf] at (3.85, 0.25) {Masked LM};
\foreach \i in {0,1,2} {
  \node[draw, fill=gray!32, rounded corners=0.5pt, minimum width=4mm, minimum height=2.6mm,
        inner sep=0pt] (encM\i) at (3.30 + \i*0.55, -0.10) {};
}
\draw[<->, thin, gray!60!black] (encM0.east) -- (encM1.west);
\draw[<->, thin, gray!60!black] (encM1.east) -- (encM2.west);

\node[microbf] at (3.85, -0.50) {Causal LM};
\foreach \i in {0,1,2} {
  \node[draw, fill=gray!42, rounded corners=0.5pt, minimum width=4mm, minimum height=2.6mm,
        inner sep=0pt] (decM\i) at (3.30 + \i*0.55, -0.85) {};
}
\draw[->, thin, gray!60!black] (decM0.east) -- (decM1.west);
\draw[->, thin, gray!60!black] (decM1.east) -- (decM2.west);

\node[microbf] at (3.85, -1.25) {Others};
\node[font=\sffamily\fontsize{11}{13}\selectfont\bfseries, gray!55!black]
  at (3.85, -1.55) {$\cdots$};

\node[panel, minimum width=7.5cm, minimum height=3.6cm] (paneB) at (9.20, 0) {};
\node[panellbl] at (paneB.north west) {(c) Architectural relationships};

\foreach \name/\xc/\yc/\title/\subtitle in {
    homo/7.40/0.45/{Homogeneous}/{same arch. family},
    hetero/11.00/0.45/{Heterogeneous}/{different arch. family},
    quant/7.40/-1.00/{Quantization}/{lower precision},
    selfd/11.00/-1.00/{Self-distillation}/{teacher \(\approx \) student}}
{
  \node[draw, rounded corners=2pt, minimum width=3.5cm, minimum height=1.30cm,
        fill=blue!4] (\name) at (\xc,\yc) {};
  \node[microbf, anchor=north west] at ([shift={(2pt,-1pt)}]\name.north west) {\title};
  \node[microf, gray!70!black, anchor=north east]
    at ([shift={(-2pt,-1pt)}]\name.north east) {\subtitle};
}

\foreach \i in {1,2,3} { \node[tblock, minimum width=4mm, minimum height=1.75mm] at (6.85, -0.05 + \i*0.175) {}; }
\foreach \i in {1,2}   { \node[sblock, minimum width=4mm, minimum height=1.25mm] at (7.95, 0.10 + \i*0.125) {}; }
\draw[->, gray!60!black, thin] (7.07, 0.30) -- (7.73, 0.30);

\foreach \i in {1,2,3} { \node[tblock,    minimum width=4mm, minimum height=1.75mm] at (10.45, -0.05 + \i*0.175) {}; }
\foreach \i in {1,2,3} { \node[ssmblock,  minimum width=4mm, minimum height=1.75mm] at (11.55, -0.05 + \i*0.175) {}; }
\draw[->, gray!60!black, thin] (10.67, 0.30) -- (11.33, 0.30);

\foreach \i in {1,2,3} { \node[tblock, minimum width=4mm, minimum height=1.75mm] at (6.85, -1.50 + \i*0.175) {}; }
\foreach \i in {1,2,3} { \node[sblock, minimum width=4mm, minimum height=1.75mm] at (7.95, -1.50 + \i*0.175) {}; }
\draw[->, gray!60!black, thin] (7.07, -1.15) -- (7.73, -1.15);
\node[microf, gray!60!black] at (6.85, -1.53) {FP32};
\node[microf, gray!60!black] at (7.95, -1.53) {INT8};

\foreach \i in {1,2,3} { \node[selfblock, minimum width=4mm, minimum height=1.75mm] at (11.00, -1.50 + \i*0.175) {}; }
\draw[->, gray!60!black, thin] (11.20, -1.15) .. controls (11.50, -1.15) and (11.50, -0.65) .. (11.00, -0.85);
\node[microf, gray!60!black, anchor=west] at (11.35, -1.00) {self};

\begin{scope}[xshift=0.525cm]
\node[panel, minimum width=15.15cm, minimum height=4.7cm] (paneC) at (4.85, -4.25) {};
\node[panellbl] at (paneC.north west) {(d) Distillation paradigms};

\node[draw, thick, rounded corners=4pt, fill=cReskd!18,
      minimum width=3.4cm, minimum height=3.6cm] (reskdR) at (-0.55, -4.6) {};
\node[font=\sffamily\bfseries, anchor=north] at ([yshift=-3pt]reskdR.north) {ResKD};
\node[font=\sffamily\scriptsize, anchor=north] at ([yshift=-15pt]reskdR.north) {Response-based};

\node[fill=gray!25, rounded corners=4pt,
      minimum width=7.2cm, minimum height=4.12cm, inner sep=0pt] (featBand) at (4.85, -4.42) {};
\node[font=\sffamily\scriptsize\itshape, green!40!black] at (4.85, -2.62) {FeatKD (feature-based)};

\node[draw, thick, rounded corners=4pt, fill=cHkd!18,
      minimum width=3.4cm, minimum height=3.6cm] (hkdR) at (3.05, -4.6) {};
\node[font=\sffamily\bfseries, anchor=north] at ([yshift=-3pt]hkdR.north) {HKD};
\node[font=\sffamily\scriptsize, anchor=north] at ([yshift=-15pt]hkdR.north) {Holistic (end-to-end)};

\node[draw, thick, rounded corners=4pt, fill=cBkd!18,
      minimum width=3.4cm, minimum height=3.6cm] (bkdR) at (6.65, -4.6) {};
\node[font=\sffamily\bfseries, anchor=north] at ([yshift=-3pt]bkdR.north) {BKD};
\node[font=\sffamily\scriptsize, anchor=north] at ([yshift=-15pt]bkdR.north) {Blockwise (independent)};

\node[draw, thick, dashed, rounded corners=4pt, fill=cRelkd!18,
      minimum width=3.4cm, minimum height=3.6cm] (relkdR) at (10.25, -4.6) {};
\node[font=\sffamily\bfseries, anchor=north] at ([yshift=-3pt]relkdR.north) {RelKD};
\node[font=\sffamily\scriptsize, anchor=north] at ([yshift=-15pt]relkdR.north) {Relation-based};

\foreach \i in {1,2,3} {
  \node[tblock, minimum width=7mm, minimum height=2.25mm]
    (tR\i) at (-1.45, -3.75 - \i*0.45) {};
}
\foreach \i in {1,2,3} {
  \node[sblock, minimum width=6mm, minimum height=2.25mm]
    (sR\i) at (0.25, -3.75 - \i*0.45) {};
}

\draw[->, thin, gray!55!black] ([yshift=2.25mm]tR1.north) -- (tR1.north);
\draw[->, thin, cStudent!80!black] ([yshift=2.25mm]sR1.north) -- (sR1.north);

\foreach \i/\j in {1/2, 2/3} {
  \draw[->, thin, gray!45!black] (tR\i.south) -- (tR\j.north);
}
\foreach \i/\j in {1/2, 2/3} {
  \draw[->, thin, cStudent!80!black] (sR\i.south) -- (sR\j.north);
}

\draw[<->, thick, dashed, cReskd!85!black] (tR3.south east) -- (sR3.south west);
\node[font=\sffamily\fontsize{5}{6}\selectfont\itshape, cReskd!85!black, anchor=south]
  at (-0.55, -5.30) {logits};

\node[font=\sffamily\fontsize{6}{7}\selectfont, text width=3.0cm, align=center, anchor=south]
  at (-0.55, -6.20) {output distribution only \\ (no intermediate access)};

\foreach \i in {1,2,3} {
  \node[tblock, minimum width=7mm, minimum height=2.25mm]
    (tH\i) at (2.15, -3.75 - \i*0.45) {};
}
\foreach \i in {1,2,3} {
  \node[sblock, minimum width=6mm, minimum height=2.25mm]
    (sH\i) at (3.85, -3.75 - \i*0.45) {};
}

\draw[->, thin, gray!55!black] ([yshift=2.25mm]tH1.north) -- (tH1.north);
\draw[->, thin, cStudent!80!black] ([yshift=2.25mm]sH1.north) -- (sH1.north);

\foreach \i/\j in {1/2, 2/3} {
  \draw[->, thin, gray!45!black] (tH\i.south) -- (tH\j.north);
}
\foreach \i/\j in {1/2, 2/3} {
  \draw[->, thin, cStudent!80!black] (sH\i.south) -- (sH\j.north);
}

\foreach \i in {1,2} {
  \draw[alignH] (tH\i.south east) -- (sH\i.south west);
}
\node[font=\sffamily\fontsize{5}{6}\selectfont\itshape, cHkd!85!black, anchor=south]
  at (3.00, -4.80) {features};

\draw[<->, thick, dashed, cReskd!85!black] (tH3.south east) -- (sH3.south west);
\node[font=\sffamily\fontsize{5}{6}\selectfont\itshape, cReskd!85!black, anchor=south]
  at (3.00, -5.30) {logits};

\draw[bp]
  ([xshift=3pt]sH3.east) -- ([xshift=3pt]sH1.east);
\node[microf, red!60!black, anchor=west] at ([xshift=3pt, yshift=0pt]sH2.east) {bp};

\node[font=\sffamily\fontsize{6}{7}\selectfont, text width=3.0cm, align=center, anchor=south]
  at (3.05, -6.20) {full student forward; one \\ backward through whole chain};

\foreach \i in {1,2,3} {
  \node[tblock, minimum width=7mm, minimum height=2.25mm]
    (tB\i) at (5.75, -3.75 - \i*0.45) {};
}
\foreach \i in {1,2,3} {
  \node[sblock, minimum width=6mm, minimum height=2.25mm]
    (sB\i) at (7.45, -3.75 - \i*0.45) {};
}

\foreach \i/\j in {1/2, 2/3} {
  \draw[->, thin, gray!45!black] (tB\i.south) -- (tB\j.north);
}

\draw[->, thin, gray!55!black] ([yshift=2.25mm]tB1.north) -- (tB1.north);
\draw[->, thin, gray!55!black] ([yshift=2.25mm]tB1.north) -- ([xshift=3pt]sB1.north west);

\foreach \i/\j in {1/2, 2/3} {
  \draw[->, thin, gray!55!black]
    ([xshift=2pt]tB\i.south) --
    ([xshift=3pt, yshift=0pt]sB\j.north west);
}

\foreach \i in {1,2} {
  \draw[alignB] (tB\i.south east) -- (sB\i.south west);
}
\node[font=\sffamily\fontsize{5}{6}\selectfont\itshape, cBkd!85!black]
  at (6.60, -4.65) {features};

\draw[<->, thick, dashed, cReskd!85!black] (tB3.south east) -- (sB3.south west);
\node[font=\sffamily\fontsize{5}{6}\selectfont\itshape, cReskd!85!black, anchor=south]
  at (6.60, -5.30) {logits};

\foreach \i in {1,2,3} {
  \draw[->, dashed, thin, red!55!black]
    ([xshift=3pt, yshift=-3pt]sB\i.east) -- ([xshift=3pt, yshift=3pt]sB\i.east);
}
\node[microf, red!60!black, anchor=west] at ([xshift=3pt, yshift=-13pt]sB1.east) {bp};

\node[font=\sffamily\fontsize{6}{7}\selectfont, text width=3.0cm, align=center, anchor=south]
  at (6.65, -6.20) {detached teacher-forced inputs; \\ independent per-block backward};

\node[draw, circle, fill=gray!28, inner sep=0pt, minimum size=3.5mm] (xa) at (9.85, -4.0) {};
\node[draw, circle, fill=gray!28, inner sep=0pt, minimum size=3.5mm] (xb) at (10.65, -4.0) {};
\node[microf] at (9.85, -4.0) {$x_1$};
\node[microf] at (10.65, -4.0) {$x_2$};

\node[draw, fill=gray!28, rounded corners=0.5pt, minimum width=6mm, minimum height=2.5mm,
      font=\sffamily\scriptsize\bfseries, inner sep=0pt] (TR) at (9.85, -4.55) {};
\node[draw, fill=cStudent!30, rounded corners=0.5pt, minimum width=6mm, minimum height=2.5mm,
      font=\sffamily\scriptsize\bfseries, inner sep=0pt] (SR) at (10.65, -4.55) {};

\draw[->, thin, gray!60!black] (xa.south) -- (TR.north);
\draw[->, thin, gray!60!black] (xa.south) -- (SR.north);
\draw[->, thin, gray!60!black] (xb.south) -- (TR.north);
\draw[->, thin, gray!60!black] (xb.south) -- (SR.north);

\node[draw, fill=purple!22, rounded corners=1pt, minimum width=6mm, minimum height=3mm,
      font=\sffamily\scriptsize, inner sep=0pt] (relT) at (9.55, -5.00) {d(T)};
\node[draw, fill=purple!22, rounded corners=1pt, minimum width=6mm, minimum height=3mm,
      font=\sffamily\scriptsize, inner sep=0pt] (relS) at (10.95, -5.00) {d(S)};

\draw[->, thin, gray!60!black] (TR.south) -- (relT.north);
\draw[->, thin, gray!60!black] (SR.south) -- (relS.north);

\draw[<->, thick, dashed, cRelkd!85!black] (relT.east) -- (relS.west);
\node[font=\sffamily\fontsize{5}{6}\selectfont\itshape, cRelkd!85!black, anchor=south]
  at (10.25, -5.00) {match};

\node[font=\sffamily\fontsize{6}{7}\selectfont, text width=3.0cm, align=center, anchor=south]
  at (10.25, -6.20) {match \emph{relationships between samples}; \\ \emph{can compose} with previous};
\end{scope}

\end{tikzpicture}%
}
\caption{\textbf{Knowledge distillation at a glance.} KD spans diverse workloads, architectural relationships, and paradigms; existing tooling covers narrow slices of this combinatorial space, motivating a general-purpose toolkit. Panel (d) shows the four paradigms used throughout the paper: response-based (ResKD), feature-based (FeatKD), holistic (HKD), blockwise (BKD), and relation-based (RelKD).}
\label{fig:kd_taxonomy}
\end{figure}
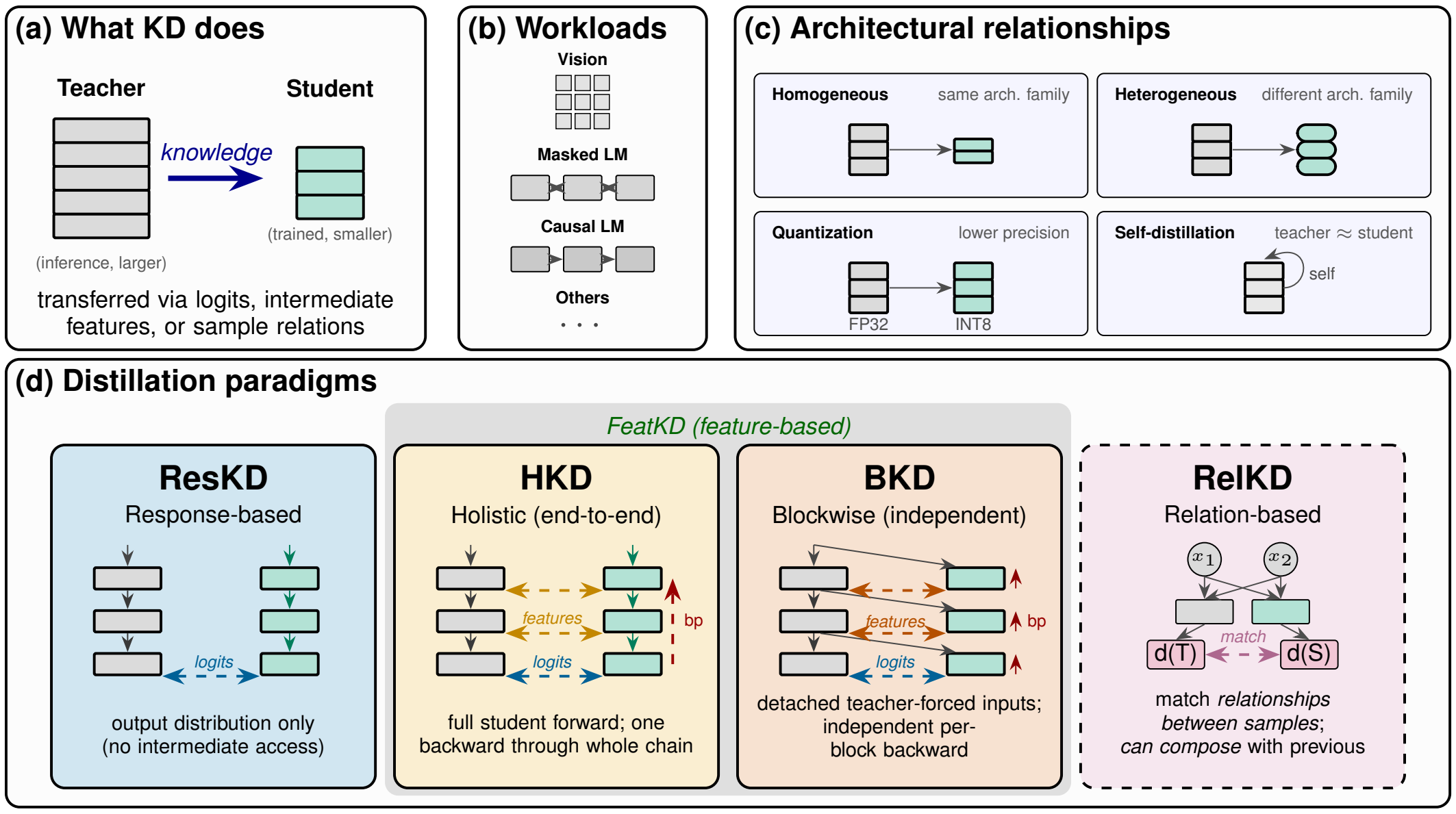

%% file: static/architecture.tex
\begin{figure}[htp]
\centering
\resizebox{\textwidth}{!}{%
\begin{tikzpicture}[
    >={Stealth[scale=0.7]},
    font=\sffamily,
    method/.style={
        draw, rounded corners=2pt, thick,
        minimum width=2.0cm, minimum height=0.42cm,
        align=center, anchor=center,
        inner xsep=2pt, inner ysep=2pt,
    },
    stock/.style    ={method, fill=cStock,   draw=cOkSky!60!black},
    added/.style    ={method, fill=cAdded,   draw=cOkOrange!70!black},
    changed/.style  ={method, fill=cChanged, draw=cOkYellow!40!black},
    flow/.style       ={->, thick, gray!60!black},
    looplabel/.style  ={font=\sffamily\fontsize{6}{7}\selectfont\itshape, gray!50!black},
    looprule/.style   ={draw=gray!30, thin, rounded corners=4pt},
    sectionlabel/.style={font=\sffamily\bfseries\footnotesize, anchor=west},
    panellabel/.style ={font=\sffamily\bfseries\normalsize},
    anchormark/.style ={draw, fill=white, circle, inner sep=0pt,
        minimum size=4mm, font=\sffamily\bfseries\fontsize{6}{7}\selectfont},
    note/.style       ={font=\sffamily\fontsize{6}{7}\selectfont\itshape, gray!45!black,
        align=center},
    dimlabel/.style   ={font=\sffamily\fontsize{6}{7}\selectfont, gray!45!black},
    tlayer/.style={draw, thick, fill=gray!22, rounded corners=1pt,
        minimum width=1.2cm, minimum height=0.42cm,
        font=\sffamily\fontsize{6}{7}\selectfont, inner sep=1pt},
    slayer/.style={draw, thick, fill=cOkGreen!22, rounded corners=1pt,
        minimum width=1.2cm, minimum height=0.42cm,
        font=\sffamily\fontsize{6}{7}\selectfont, inner sep=1pt},
    tdim/.style={draw=gray!55!black, dashed, thick,
        fill=gray!10, rounded corners=1pt,
        minimum width=1.2cm, minimum height=0.42cm,
        font=\sffamily\fontsize{6}{7}\selectfont\itshape\color{gray!55!black},
        inner sep=1pt},
    sdim/.style={draw=cOkGreen!50!black, dashed, thick,
        fill=cOkGreen!10, rounded corners=1pt,
        minimum width=1.2cm, minimum height=0.42cm,
        font=\sffamily\fontsize{6}{7}\selectfont\itshape\color{gray!55!black},
        inner sep=1pt},
    align/.style={<->, thick, dashed, cOkYellow!55!black},
    hook/.style={draw, fill=cOkYellow!70, rounded corners=0.5pt,
        minimum width=1.5mm, minimum height=1.5mm, inner sep=0pt},
    code/.style={draw=gray!50, rounded corners=2pt, fill=gray!8,
        font=\ttfamily\fontsize{6}{7}\selectfont, inner sep=2pt,
        minimum height=0.42cm},
    tact/.style={draw=gray!60!black, thick, fill=gray!22, rounded corners=1pt,
        minimum height=0.4cm, font=\sffamily\fontsize{6}{7}\selectfont,
        inner sep=2pt},
    sact/.style={draw, thick, fill=cOkGreen!22, rounded corners=1pt,
        minimum height=0.4cm, font=\sffamily\fontsize{6}{7}\selectfont,
        inner sep=2pt},
    proj/.style={draw, thick, fill=cOkYellow!40, rounded corners=2pt,
        minimum height=0.5cm, font=\sffamily\fontsize{6}{7}\selectfont\bfseries,
        inner sep=2pt},
    gpu/.style={draw=gray!60!black, rounded corners=2pt, thick,
        fill=gray!12, minimum width=1.5cm, minimum height=1.4cm,
        anchor=north, inner sep=0pt},
    tslot/.style={draw=cTSlotBorder, thick, fill=cTSlot,
        rounded corners=1pt, minimum width=1.25cm, minimum height=0.45cm,
        anchor=north, font=\sffamily\fontsize{6}{7}\selectfont, inner sep=1pt,
        text=black},
    sslot/.style={draw=cSSlotBorder, thick, fill=cSSlot,
        rounded corners=1pt, minimum width=1.25cm, minimum height=0.45cm,
        anchor=north, font=\sffamily\fontsize{6}{7}\selectfont, inner sep=1pt,
        text=black},
    panelframe/.style={draw=gray!45, thin, rounded corners=3pt,
        anchor=center},
    streamT/.style={draw=gray!60!black, thick, fill=gray!30,
        rounded corners=1pt, minimum height=0.28cm, anchor=west,
        font=\sffamily\fontsize{5}{6}\selectfont, inner sep=1pt},
    streamS/.style={draw=cOkGreen!60!black, thick, fill=cOkGreen!30,
        rounded corners=1pt, minimum height=0.28cm, anchor=west,
        font=\sffamily\fontsize{5}{6}\selectfont, inner sep=1pt},
    config/.style={font=\sffamily\fontsize{6}{7}\selectfont, anchor=west,
        align=left},
]

\begin{scope}[on background layer]
\node[panelframe, minimum width=14cm, minimum height=3.4cm]
    (frameA) at (7, -1.7) {};
\node[panelframe, minimum width=4.67cm, minimum height=5.0cm]
    (frameB) at (2.33, -5.9) {};
\node[panelframe, minimum width=4.67cm, minimum height=5.0cm]
    (frameC) at (7.0, -5.9) {};
\node[panelframe, minimum width=4.67cm, minimum height=5.0cm]
    (frameD) at (11.67, -5.9) {};
\end{scope}

\node[panellabel, anchor=north west]
    (labelA) at ([shift={(0.15,-0.12)}]frameA.north west) {(a)};
\node[sectionlabel, anchor=west]
    at ([xshift=0.1cm]labelA.east)
    {Training Flow};

\node[panellabel, anchor=north west]
    (labelB) at ([shift={(0.15,-0.12)}]frameB.north west) {(b)};
\node[sectionlabel, anchor=west]
    at ([xshift=0.1cm]labelB.east) {Alignment mechanism};

\node[panellabel, anchor=north west]
    (labelC) at ([shift={(0.15,-0.12)}]frameC.north west) {(c)};
\node[sectionlabel, anchor=west]
    at ([xshift=0.1cm]labelC.east) {Parallelism};

\node[panellabel, anchor=north west]
    (labelD) at ([shift={(0.15,-0.12)}]frameD.north west) {(d)};
\node[sectionlabel, anchor=west]
    at ([xshift=0.1cm]labelD.east) {Auto-projectors};

\begin{scope}[shift={(0.95,-0.745)}]

\node[stock]   (loadModels)  at (1.2, -0.45)
    {\mtitle{\texttt{from\_pretrained()}}};
\node[added]   (createAlign) at (3.6, -0.45)
    {\mtitle{\texttt{create\_alignments()}}};
\node[added]   (initHooks)   at (6.0, -0.45)
    {\mtitle{\texttt{\_\_init\_\_()}}};
\node[changed] (initOpt)     at (8.4, -0.45)
    {\mtitle{\texttt{create\_optimizer()}}};

\foreach \a/\b in {loadModels/createAlign, createAlign/initHooks, initHooks/initOpt}
    \draw[flow] (\a.east) -- (\b.west);

\node[changed] (trainStep)   at (1.2, -1.30)
    {\mtitle{\texttt{training\_step()}}};
\node[added]   (computeLoss) at (3.6, -1.30)
    {\mtitle{\texttt{compute\_dist\_loss()}}};
\node[changed] (backward)    at (6.0, -1.30)
    {\mtitle{\texttt{backward()}}};
\node[changed] (optStep)     at (8.4, -1.30)
    {\mtitle{\texttt{optimizer.step()}}};
\node[changed] (logStep)     at (10.8, -1.30)
    {\mtitle{\texttt{log()}}};

\foreach \a/\b in {trainStep/computeLoss, computeLoss/backward,
                   backward/optStep, optStep/logStep}
    \draw[flow] (\a.east) -- (\b.west);

\draw[flow]
    (initOpt.east) -- ++(0.3,0)
    -- ++(0,-0.4)
    -| (trainStep.north);

\node[changed] (evalStep)    at (3.6, -2.05)
    {\mtitle{\texttt{evaluate()}}};
\node[changed] (saveStep)    at (6.0, -2.05)
    {\mtitle{\texttt{\_save\_checkpoint()}}};

\draw[flow] (evalStep.east) -- (saveStep.west);

\draw[flow]
    (trainStep.south) |- (evalStep.west);

\draw[flow, gray!55]
    (saveStep.east) -- ++(0.6,0);
\node[font=\sffamily\fontsize{7}{8}\selectfont\itshape, gray!55!black,
      anchor=west]
    at ([xshift=0.6cm]saveStep.east) {\dots};

\draw[flow]
    (logStep.south) -- ++(0,-0.2)
    -| ([xshift=-10pt] trainStep.south east);

\end{scope}

\begin{scope}[shift={(0.93,-4.40)}]

\node[code, anchor=north]
    (regex) at (1.4, 0)
    {modules=r"layers\textbackslash.\textbackslash d+\$"};
\node[font=\sffamily\fontsize{5}{6}\selectfont, gray!55!black, anchor=south]
    at (regex.north)
    {user-supplied pattern};

\draw[->, thick, gray!55!black]
    (regex.south) -- ++(0,-0.4)
    node[midway, right=1pt, font=\sffamily\fontsize{5}{6}\selectfont\itshape,
         text=gray!55!black] {};

\node[font=\sffamily\bfseries\fontsize{6}{7}\selectfont, anchor=south]
    at (0.5, -1.05) {Teacher (frozen)};
\node[tdim, anchor=north] (te) at (0.5, -1.10)  {embed};
\node[tlayer, anchor=north] (tL0) at (0.5,-1.6) {layers.0};
\node[tlayer, anchor=north] (tL1) at (0.5,-2.1) {layers.1};
\node[tlayer, anchor=north] (tL2) at (0.5,-2.6) {layers.2};
\node[tdim, anchor=north]   (th)  at (0.5,-3.1) {head};

\node[font=\sffamily\bfseries\fontsize{6}{7}\selectfont, anchor=south]
    at (2.3, -1.05) {Student};
\node[sdim, anchor=north]   (se)  at (2.3,-1.10) {embed};
\node[slayer, anchor=north] (sL0) at (2.3,-1.6) {layers.0};
\node[slayer, anchor=north] (sL1) at (2.3,-2.1) {layers.1};
\node[slayer, anchor=north] (sL2) at (2.3,-2.6) {layers.2};
\node[sdim, anchor=north]   (sh)  at (2.3,-3.1) {head};

\foreach \n in {tL0,tL1,tL2}
    \node[hook] at ([xshift=-0.7cm]\n.center) {};
\foreach \n in {sL0,sL1,sL2}
    \node[hook] at ([xshift= 0.7cm]\n.center) {};

\foreach \tn/\sn in {tL0/sL0, tL1/sL1, tL2/sL2} {
    \draw[align] (\tn.east) -- (\sn.west);
}

\end{scope}

\begin{scope}[shift={(10.27,-4.30)}]

\def\barW{2.2cm}
\def\centerX{1.3}

\node[font=\sffamily\bfseries\fontsize{6}{7}\selectfont, anchor=south]
    at (\centerX,-0.1) {captured pair};

\node[tact, minimum width=\barW, anchor=north]
    (tact) at (\centerX,-0.25)
    {teacher activation};
\node[dimlabel, anchor=west]
    at ([xshift=2pt,yshift=6pt]tact.east) {$d_T$};

\node[sact, minimum width=\barW, anchor=north]
    (sact) at (\centerX,-0.85)
    {student activation};
\node[dimlabel, anchor=west]
    at ([xshift=2pt]sact.east) {$d_S$};

\node[font=\sffamily\fontsize{6}{7}\selectfont\itshape, text=cOkVermil!75!black,
      anchor=north]
    (mismatch) at (\centerX,-1.42)
    {$d_T \!\neq\! d_S$ detected};

\node[proj, anchor=north, minimum width=\barW]
    (proj) at (\centerX,-1.78)
    {Linear ($d_S \!\to\! d_T$)};

\draw[flow] (sact.west) -- ++(-0.4,0) |- (proj.west);

\node[sact, minimum width=\barW, anchor=north,
      fill=cOkGreen!12, draw=cOkGreen!50!black, dashed]
    (sproj) at (\centerX,-2.5)
    {projected student};
\node[dimlabel, anchor=west]
    at ([xshift=2pt]sproj.east) {$d_T$};

\draw[flow] (proj.south) -- (sproj.north);

\node[anchor=north, draw=cOkVermil!70!black, thick,
      fill=cOkVermil!12, rounded corners=2pt,
      font=\sffamily\fontsize{6}{7}\selectfont\bfseries,
      minimum width=\barW, minimum height=0.42cm]
    (loss) at (\centerX,-3.05)
    {\texttt{loss\_fn}};

\draw[flow, dashed, draw=cOkVermil!70!black]
    (sproj.south) -- (loss.north);
\draw[flow, dashed, draw=cOkVermil!70!black]
    (tact.east) -- ++(0.85,0) |- (loss.east);

\end{scope}

\begin{scope}[shift={(5.35,-4.25)}]

\node[draw=black, thick, rounded corners=2pt, fill=white,
      inner sep=2pt, anchor=north,
      font=\sffamily\fontsize{5}{6}\selectfont, text=black]
    (configA) at (1.5,0)
    {\texttt{teacher\_placement="sharded"}};
\node[draw=black, thick, rounded corners=2pt, fill=white,
      inner sep=2pt, anchor=north,
      font=\sffamily\fontsize{5}{6}\selectfont, text=black]
    (config) at ([yshift=-0.5pt]configA.south)
    {\texttt{fsdp="full\_shard"}};

\node[gpu, anchor=north]
    (gpu0) at (0.5,-0.95) {};
\node[font=\sffamily\bfseries\fontsize{6}{7}\selectfont, anchor=north]
    at ([yshift=0pt]gpu0.north) {GPU 0};

\node[gpu, anchor=north]
    (gpu1) at (2.3,-0.95) {};
\node[font=\sffamily\bfseries\fontsize{6}{7}\selectfont, anchor=north]
    at ([yshift=0pt]gpu1.north) {GPU 1};

\node[tslot, anchor=north]
    (t0) at ([yshift=-0.35cm]gpu0.north)
    {T shard 0};
\node[tslot, anchor=north]
    (t1) at ([yshift=-0.35cm]gpu1.north)
    {T shard 1};

\node[sslot, anchor=north]
    (s0) at ([yshift=-0.05cm]t0.south)
    {S shard 0};
\node[sslot, anchor=north]
    (s1) at ([yshift=-0.05cm]t1.south)
    {S shard 1};

\node[font=\sffamily\bfseries\fontsize{6}{7}\selectfont, anchor=west]
    at (-0.05,-2.7)
    {Forward (independent streams)};

\draw[gray!50, thick] (0,-3.25) -- (3.0,-3.25);

\node[streamT, minimum width=1.7cm]
    at (0,-3.08) {teacher fwd};
\node[streamS, minimum width=2.0cm]
    at (0.35,-3.48) {student fwd};

\draw[<->, thick, gray!60!black]
    (0.35,-3.80) -- (1.7,-3.80);
\node[font=\sffamily\fontsize{5}{6}\selectfont\itshape, gray!60!black,
      anchor=north]
    at (1.025,-3.82) {overlap};

\end{scope}

\end{tikzpicture}%
}
\caption{\textbf{SilverSpoon-KD design overview.} (a)~Training-flow spine, methods color-coded by provenance (\colorbox{cStock}{inherited}, \colorbox{cAdded}{added}, \colorbox{cChanged}{extended}). (b)~Regex/list alignment hooks bind teacher--student modules. (c)~Independent placement overlaps teacher/student forwards on disjoint GPUs. (d)~Auto-projectors bridge hidden-dim mismatches.}
\label{fig:architecture}
\end{figure}
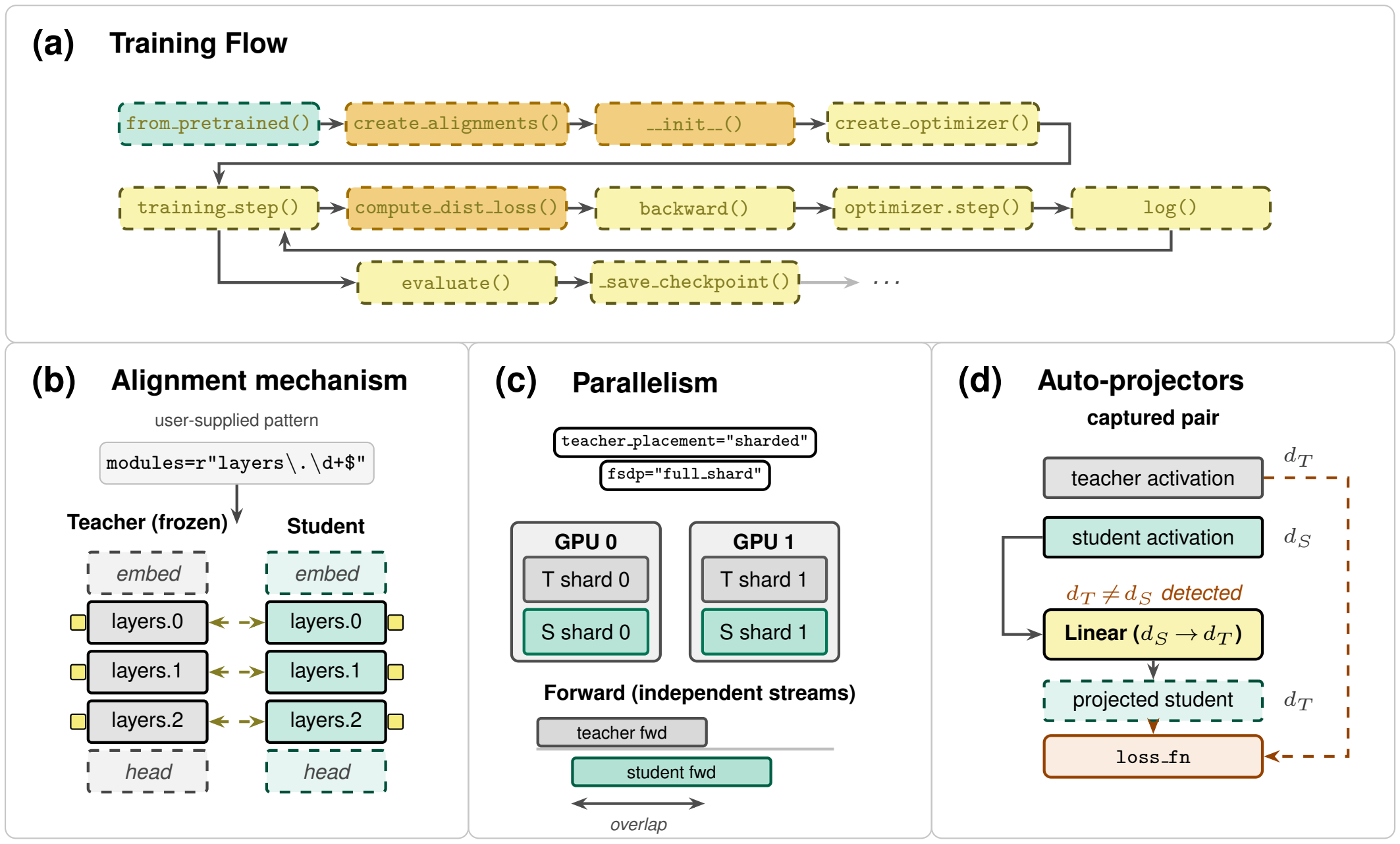

%% file: static/api_snippet.tex
\begin{figure}[htp]
\centering
\begin{minipage}[t]{0.32\linewidth}
\centering\textbf{Response-based KD}\par
\begin{lstlisting}[style=kdpython, basicstyle=\ttfamily\tiny,
  backgroundcolor=\color{cReskd!6},
  emph={[2]ResponseBasedDistiller},
  emphstyle={[2]\color{cReskd!85!black}\bfseries}]
# Logit matching only
distiller = ResponseBasedDistiller(
    teacher_model=teacher,
    student_model=student,
    soft_loss_fn=kl_divergence_loss(
        temperature=4.0,
    ),
    train_dataset=ds,
    args=TrainingArguments(alpha=0.5),
)
distiller.train()
\end{lstlisting}
\end{minipage}\hfill
\begin{minipage}[t]{0.32\linewidth}
\centering\textbf{Holistic KD}\par
\begin{lstlisting}[style=kdpython, basicstyle=\ttfamily\tiny,
  backgroundcolor=\color{cHkd!7},
  emph={[2]HolisticDistiller},
  emphstyle={[2]\color{cHkd!85!black}\bfseries}]
# Hook layers + lm_head (final logits)
alignments = create_alignments(
    teacher, student,
    modules=r"^(layers\.\d+|lm_head)$",
)
for i, a in enumerate(alignments):
    a.loss_weight = i + 1
distiller = HolisticDistiller(
    teacher_model=teacher,
    alignments=alignments,
    train_dataset=ds,
    args=TrainingArguments(),
)
distiller.train()
\end{lstlisting}
\end{minipage}\hfill
\begin{minipage}[t]{0.32\linewidth}
\centering\textbf{Blockwise KD}\par
\begin{lstlisting}[style=kdpython, basicstyle=\ttfamily\tiny,
  backgroundcolor=\color{cBkd!6},
  emph={[2]BlockwiseDistiller},
  emphstyle={[2]\color{cBkd!85!black}\bfseries}]
# Hook all transformer layers
alignments = create_alignments(
    teacher, student,
    modules=r"^layers\.\d+$",
)
distiller = BlockwiseDistiller(
    teacher_model=teacher,
    alignments=alignments,
    train_dataset=ds,
    args=TrainingArguments(),
)
distiller.train()
\end{lstlisting}
\end{minipage}
\caption{\textbf{KD paradigms via class swap.} Switching between KD paradigms merely requires swapping the distiller class; alignments, training arguments, and dataset flow into all three. \texttt{teacher}, \texttt{student}, and \texttt{train\_dataset=ds} assumed loaded; configurations are illustrative.}
\label{fig:api_snippet}
\end{figure}

%% file: static/experiment_coverage.tex
\begin{figure}[htp]
    \centering
    \begin{tikzpicture}[font=\sffamily\scriptsize]

    \matrix (M) [
        matrix of nodes,
        nodes in empty cells,
        nodes={
            anchor=center,
            inner sep=1pt,
            minimum height=7mm,
            text depth=0pt,
            font=\sffamily\scriptsize,
        },
        row sep=-\pgflinewidth,
        column sep=-\pgflinewidth,
        column 1/.style={nodes={text width=29mm, align=right, anchor=east, inner sep=2pt, draw=none, fill=none}},
        column 2/.style={nodes={text width=14mm, align=center, draw=gray!40, fill=green!6}},
        column 3/.style={nodes={text width=16mm, align=center, draw=gray!40, fill=blue!6}},
        column 4/.style={nodes={text width=16mm, align=center, draw=gray!40, fill=blue!6}},
        column 5/.style={nodes={text width=18mm, align=center, draw=gray!40, fill=blue!6}},
        column 6/.style={nodes={text width=14mm, align=center, draw=gray!40, fill=cAlignment!18}},
        column 7/.style={nodes={text width=10mm, align=center, draw=gray!40, fill=gray!12}},
        column 8/.style={nodes={text width=12mm, align=center, draw=gray!40, fill=cStudent!12}},
        row 1/.style={nodes={font=\sffamily\scriptsize\bfseries, minimum height=9mm}},
    ] {
        & Workload & Teacher & Student(s) & Architecture Relation & Paradigms & Stages & RQ \\
        \S\ref{exp:vision_bkd}~VGG Compression              & CNN  & VGG-16      & VGG-16-DS          & Homogeneous         & \pR\,\pH\,\pB     & \chk      & 1, 2 \\
        \S\ref{exp:vision_relkd}~VGG Relational KD          & CNN  & ResNet-50   & VGG-11-BN          & Heterogeneous       & \pR\,\pH\,\pRel   & ---       & 1    \\
        \S\ref{exp:bert_dist}~BERT Pre-Training              & Lang / Enc.   & BERT-base   & T6, T4-tiny        & Homogeneous         & \pR\,\pH\,\pB     & ---      & 1, 2 \\
        \S\ref{exp:bert_downstream}~BERT Downstream          & Lang / Enc.   & BERT-base   & T6, T4-tiny        & Homogeneous         & \pH\,\pB          & ---      & 1, 2 \\
        \S\ref{exp:gpt2_dist}~GPT-2 Compression             & Lang / Dec.   & GPT-2 small    & GPT-2 96M          & Homogeneous         & \pR\,\pH\,\pB     & \chk      & 1, 2 \\
        \S\ref{exp:gpt2_cross_arch}~Cross-Architecture       & Lang / Dec.   & GPT-2 small    & DSV3 96M           & Heterogeneous       & \pR\,\pH\,\pB     & \chk     & 1    \\
        \S\ref{exp:gpt2_linear}~GPT-2 Attn.\ Linear.        & Lang / Dec.   & GPT-2 small    & LoLCATs            & Heterogeneous       & \pR\,\pH\,\pB     & \chk      & 1, 2 \\
        \S\ref{exp:qwen3_quant}~Qwen3 Quantization           & Lang / LLM    & Qwen3-1.7B  & INT4/5/6           & Quantization  & \pR\,\pH\,\pB     & ---      & 1, 2 \\
        \S\ref{exp:framework}~Toolkit Bench.            & \emph{mixed}  & \emph{mixed} & \emph{mixed}      & \emph{mixed}        & \pR\,\pH\,\pB     & ---      & 3    \\
    };

    \draw[thick, gray!70]
        ($(M-1-2.south west)$) -- ($(M-1-8.south east)$);

    \draw[thick, dashed, gray!55]
        ($(M-10-2.north west)+(-1.5mm,0)$) -- ($(M-10-8.north east)+(1.5mm,0)$);

    \end{tikzpicture}

    \caption{\textbf{Experimental coverage matrix.} The eight distillation-quality experiments above the dashed line span vision, encoder LM, and decoder LM workloads under homogeneous, heterogeneous, and quantization-based settings, exercising all four paradigms; the toolkit-performance benchmark below the line tests SilverSpoon-KD against four peers under matched workloads. Pills: \pR\,response-based, \pH\,holistic, \pB\,blockwise, \pRel\,relation-based KD.}
    \label{fig:experiment_coverage}
\end{figure}

%% file: generated/results_summary.tex
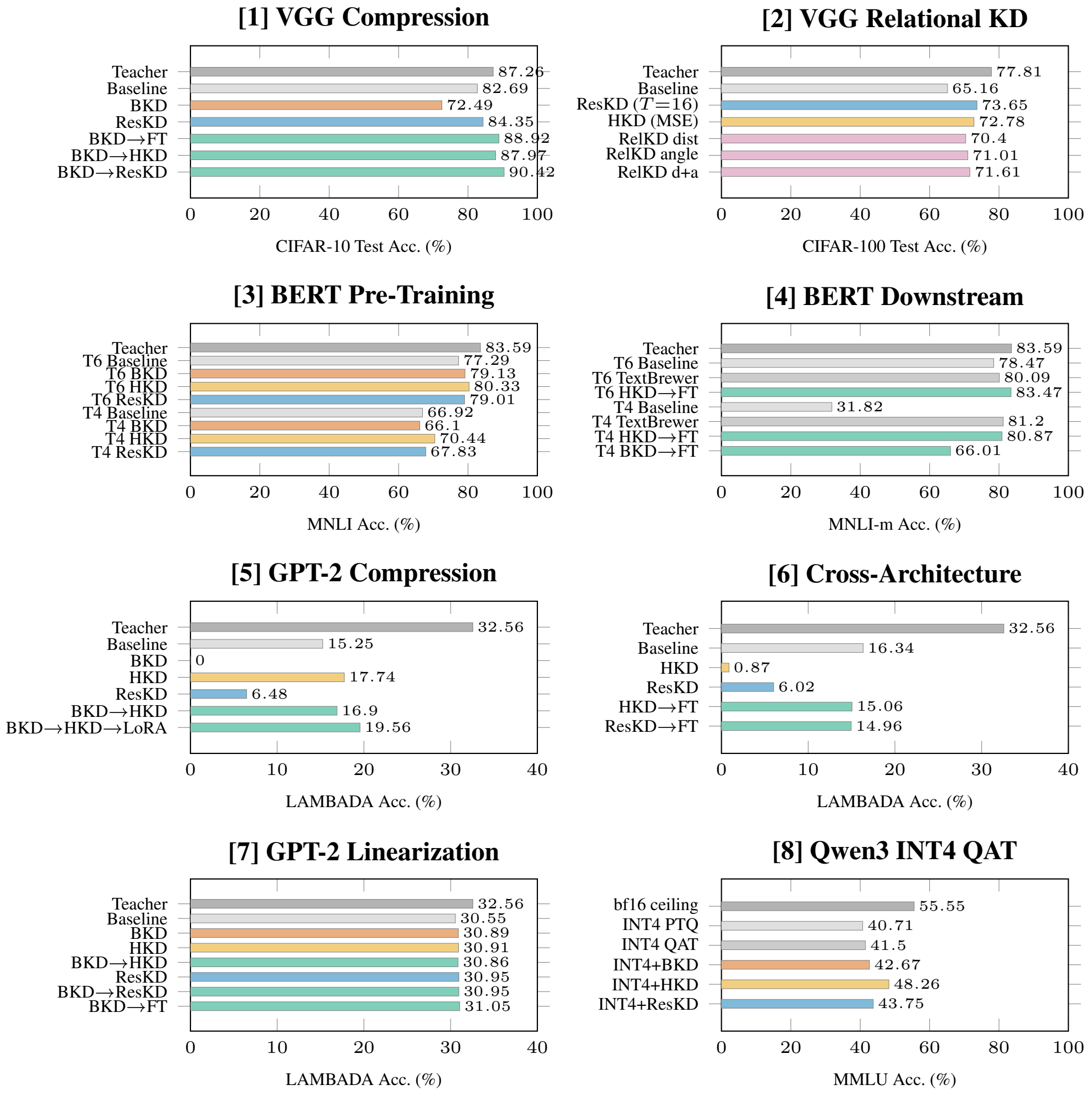
\begin{figure}[htp]
\centering
\pgfplotsset{
  bar/.style={xbar, bar shift=0pt, bar width=3pt, draw=black!50, line width=0.15pt},
}
\begin{tikzpicture}
\begin{groupplot}[
  group style={group size=2 by 4, vertical sep=1.5cm, horizontal sep=2.2cm},
  width=0.41\textwidth, height=3.4cm,
  xbar,
  enlarge y limits=0.15,
  y dir=reverse,
  yticklabel style={font=\tiny, align=right},
  xticklabel style={font=\tiny},
  title style={font=\small\bfseries, yshift=-1ex},
  xlabel style={font=\tiny},
  nodes near coords, nodes near coords style={font=\fontsize{4.5}{5.5}\selectfont,
    /pgf/number format/precision=2, /pgf/number format/fixed,
    inner sep=0pt, text height=3pt, text depth=0pt},
  every node near coord/.append style={anchor=west, xshift=1.5pt},
]

\nextgroupplot[title={[1] VGG Compression},
  xlabel={CIFAR-10 Test Acc.\ (\%)},
  ytick={1,...,7}, ymin=0.5, ymax=7.5,
  yticklabels={Teacher,Baseline,BKD,ResKD,BKD\textrightarrow FT,BKD\textrightarrow HKD,BKD\textrightarrow ResKD},
  xmin=0, xmax=100]
\addplot[bar,fill=gray!60] coordinates {(87.26,1)};
\addplot[bar,fill=gray!25] coordinates {(82.69,2)};
\addplot[bar,fill=cBkd!50] coordinates {(72.49,3)};
\addplot[bar,fill=cReskd!50] coordinates {(84.35,4)};
\addplot[bar,fill=cMulti!50] coordinates {(88.92,5)};
\addplot[bar,fill=cMulti!50] coordinates {(87.97,6)};
\addplot[bar,fill=cMulti!50] coordinates {(90.42,7)};

\nextgroupplot[title={[2] VGG Relational KD},
  xlabel={CIFAR-100 Test Acc.\ (\%)},
  ytick={1,...,7}, ymin=0.5, ymax=7.5,
  yticklabels={Teacher,Baseline,ResKD ($T{=}16$),HKD (MSE),RelKD dist,RelKD angle,RelKD d+a},
  xmin=0, xmax=100]
\addplot[bar,fill=gray!60] coordinates {(77.81,1)};
\addplot[bar,fill=gray!25] coordinates {(65.16,2)};
\addplot[bar,fill=cReskd!50] coordinates {(73.65,3)};
\addplot[bar,fill=cHkd!50] coordinates {(72.78,4)};
\addplot[bar,fill=cRelkd!50] coordinates {(70.40,5)};
\addplot[bar,fill=cRelkd!50] coordinates {(71.01,6)};
\addplot[bar,fill=cRelkd!50] coordinates {(71.61,7)};

\nextgroupplot[title={[3] BERT Pre-Training},
  xlabel={MNLI Acc.\ (\%)},
  ytick={1,...,9}, ymin=0.5, ymax=9.5,
  yticklabels={Teacher,T6 Baseline,T6 BKD,T6 HKD,T6 ResKD,T4 Baseline,T4 BKD,T4 HKD,T4 ResKD},
  xmin=0, xmax=100]
\addplot[bar,fill=gray!60] coordinates {(83.59,1)};
\addplot[bar,fill=gray!25] coordinates {(77.29,2)};
\addplot[bar,fill=cBkd!50] coordinates {(79.13,3)};
\addplot[bar,fill=cHkd!50] coordinates {(80.33,4)};
\addplot[bar,fill=cReskd!50] coordinates {(79.01,5)};
\addplot[bar,fill=gray!25] coordinates {(66.92,6)};
\addplot[bar,fill=cBkd!50] coordinates {(66.10,7)};
\addplot[bar,fill=cHkd!50] coordinates {(70.44,8)};
\addplot[bar,fill=cReskd!50] coordinates {(67.83,9)};

\nextgroupplot[title={[4] BERT Downstream},
  xlabel={MNLI-m Acc.\ (\%)},
  ytick={1,...,8}, ymin=0.5, ymax=8.5,
  yticklabels={Teacher,T6 Baseline,T6 TextBrewer,T6 HKD\textrightarrow FT,T4 Baseline,T4 TextBrewer,T4 HKD\textrightarrow FT,T4 BKD\textrightarrow FT},
  xmin=0, xmax=100]
\addplot[bar,fill=gray!60] coordinates {(83.59,1)};
\addplot[bar,fill=gray!25] coordinates {(78.47,2)};
\addplot[bar,fill=gray!40] coordinates {(80.09,3)};
\addplot[bar,fill=cMulti!50] coordinates {(83.47,4)};
\addplot[bar,fill=gray!25] coordinates {(31.82,5)};
\addplot[bar,fill=gray!40] coordinates {(81.20,6)};
\addplot[bar,fill=cMulti!50] coordinates {(80.87,7)};
\addplot[bar,fill=cMulti!50] coordinates {(66.01,8)};

\nextgroupplot[title={[5] GPT-2 Compression},
  xlabel={LAMBADA Acc.\ (\%)},
  ytick={1,...,7}, ymin=0.5, ymax=7.5,
  yticklabels={Teacher,Baseline,BKD,HKD,ResKD,BKD\textrightarrow HKD,BKD\textrightarrow HKD\textrightarrow LoRA},
  xmin=0, xmax=40]
\addplot[bar,fill=gray!60] coordinates {(32.56,1)};
\addplot[bar,fill=gray!25] coordinates {(15.25,2)};
\addplot[bar,fill=cBkd!50] coordinates {(0.00,3)};
\addplot[bar,fill=cHkd!50] coordinates {(17.74,4)};
\addplot[bar,fill=cReskd!50] coordinates {(6.48,5)};
\addplot[bar,fill=cMulti!50] coordinates {(16.90,6)};
\addplot[bar,fill=cMulti!50] coordinates {(19.56,7)};

\nextgroupplot[title={[6] Cross-Architecture},
  xlabel={LAMBADA Acc.\ (\%)},
  ytick={1,...,6}, ymin=0.5, ymax=6.5,
  yticklabels={Teacher,Baseline,HKD,ResKD,HKD\textrightarrow FT,ResKD\textrightarrow FT},
  xmin=0, xmax=40]
\addplot[bar,fill=gray!60] coordinates {(32.56,1)};
\addplot[bar,fill=gray!25] coordinates {(16.34,2)};
\addplot[bar,fill=cHkd!50] coordinates {(0.87,3)};
\addplot[bar,fill=cReskd!50] coordinates {(6.02,4)};
\addplot[bar,fill=cMulti!50] coordinates {(15.06,5)};
\addplot[bar,fill=cMulti!50] coordinates {(14.96,6)};

\nextgroupplot[title={[7] GPT-2 Linearization},
  xlabel={LAMBADA Acc.\ (\%)},
  ytick={1,...,8}, ymin=0.5, ymax=8.5,
  yticklabels={Teacher,Baseline,BKD,HKD,BKD\textrightarrow HKD,ResKD,BKD\textrightarrow ResKD,BKD\textrightarrow FT},
  xmin=0, xmax=40]
\addplot[bar,fill=gray!60] coordinates {(32.56,1)};
\addplot[bar,fill=gray!25] coordinates {(30.55,2)};
\addplot[bar,fill=cBkd!50] coordinates {(30.89,3)};
\addplot[bar,fill=cHkd!50] coordinates {(30.91,4)};
\addplot[bar,fill=cMulti!50] coordinates {(30.86,5)};
\addplot[bar,fill=cReskd!50] coordinates {(30.95,6)};
\addplot[bar,fill=cMulti!50] coordinates {(30.95,7)};
\addplot[bar,fill=cMulti!50] coordinates {(31.05,8)};

\nextgroupplot[title={[8] Qwen3 INT4 QAT},
  xlabel={MMLU Acc.\ (\%)},
  ytick={1,...,6}, ymin=0.5, ymax=6.5,
  yticklabels={bf16 ceiling,INT4 PTQ,INT4 QAT,INT4{+}BKD,INT4{+}HKD,INT4{+}ResKD},
  xmin=0, xmax=100]
\addplot[bar,fill=gray!60] coordinates {(55.55,1)};
\addplot[bar,fill=gray!25] coordinates {(40.71,2)};
\addplot[bar,fill=gray!40] coordinates {(41.50,3)};
\addplot[bar,fill=cBkd!50] coordinates {(42.67,4)};
\addplot[bar,fill=cHkd!50] coordinates {(48.26,5)};
\addplot[bar,fill=cReskd!50] coordinates {(43.75,6)};

\end{groupplot}
\end{tikzpicture}
\caption{\textbf{Per-experiment results summary.} Colored bars are SilverSpoon-KD runs: single-stage bars use paradigm colors (\textcolor{cBkd}{\textbf{BKD}}, \textcolor{cHkd}{\textbf{HKD}}, \textcolor{cReskd}{\textbf{ResKD}}, \textcolor{cRelkd}{\textbf{RelKD}}), and multi-stage chains use \textcolor{cMulti}{green}, with the chain (e.g., BKD\textrightarrow HKD\textrightarrow LoRA) shown in the row label. Gray bars are reference points (not SilverSpoon-KD): teacher (dark), no-distillation student baseline (light), and external comparison from a competing toolkit or reference paper (medium). FT = full fine-tuning; LoRA = low-rank adapters, base frozen.}
\label{fig:results_summary}
\end{figure}

%% file: generated/framework.tex
\begin{table}[htp]
\centering\footnotesize
\caption{\textbf{Head-to-head toolkit benchmark} (fixed batch size, single GPU unless noted). SilverSpoon-KD wins peak memory in all 5 settings and wall time in 3/5. $\Delta$~columns are SilverSpoon-KD relative to comparison; negative is better. \textbf{Bold} = winner per metric.}
\label{tab:framework}
\begin{tabular}{@{}llccccccc@{}}
\toprule
Setting & Toolkit & Wall (s) & Tput (samp/s) & Peak Mem (MB) & $\Delta$ Wall & $\Delta$ Mem \\
\midrule
\multicolumn{7}{@{}l}{\emph{ResNet-50 $\rightarrow$ ResNet-18 / ImageNet (vision, \gls{reskd}, DDP $\times$ 2)}} \\
 & torchdistill   & 232.0 & 551.7 & 1\,341 & --- & --- \\
 & SilverSpoon-KD & \textbf{219.6} & \textbf{582.8} & \textbf{1\,014} & $-5.3\%$ & $-24.4\%$ \\
\addlinespace
\multicolumn{7}{@{}l}{\emph{BERT-base $\rightarrow$ T6 / MNLI (encoder NLP, \gls{reskd}, DDP $\times$ 2)}} \\
 & TextBrewer     & 130.1 & 491.8 & 2\,758 & --- & --- \\
 & SilverSpoon-KD & \textbf{124.5} & \textbf{514.1} & \textbf{2\,747} & $-4.3\%$ & $-0.4\%$ \\
\addlinespace
\multicolumn{7}{@{}l}{\emph{BERT-base $\rightarrow$ T6 / MNLI (encoder NLP, \gls{hkd})}} \\
 & TextBrewer (default)        & \textbf{425.5} & \textbf{300.8} & 6\,360 & --- & --- \\
 & SilverSpoon-KD & 449.6 & 284.7 & \textbf{6\,356} & $+5.7\%$ & $-0.1\%$ \\
\addlinespace
\multicolumn{7}{@{}l}{\emph{Qwen3-4B $\rightarrow$ 0.6B / Dolma (decoder LLM, \gls{hkd}, bf16)}} \\
 & DistillKit (gradient ckpt)   & 294.4 & 3.4 & 23\,992 & --- & --- \\
 & SilverSpoon-KD (Liger fused) & \textbf{288.6} & \textbf{3.5} & \textbf{20\,750} & $-2.0\%$ & $-13.5\%$ \\
\addlinespace
\multicolumn{7}{@{}l}{\emph{Qwen3-4B $\rightarrow$ 0.6B / Dolma (decoder LLM, \gls{reskd}+LoRA, bf16)}} \\
 & torchtune      & \textbf{639.1} & \textbf{3.1} & 23\,090 & --- & --- \\
 & SilverSpoon-KD & 675.3 & 3.0 & \textbf{19\,609} & $+5.7\%$ & $-15.1\%$ \\
\bottomrule
\end{tabular}
\end{table}

%% file: generated/framework_capability.tex
\begin{table}[htp]
\centering\footnotesize
\caption{\textbf{Capability frontier as teacher size grows} on 2$\times$A100 (${\sim}40$~GB per-GPU), Qwen3-0.6B student, seq=1024. SilverSpoon-KD is the only setup that scales to Qwen3-14B with batch~$>$~1; DistillKit \textit{OOM}s at 14B. \textbf{Bold} = best per metric per teacher.}
\label{tab:framework_capability}
\begin{tabular}{@{}llccc@{}}
\toprule
Teacher & Toolkit & Batch & Tput (tok/s) & Peak Mem (MB) \\
\midrule
Qwen3-4B (8 GB)   & DistillKit     & 4 & 15\,184 & 37\,967 \\
                  & TextBrewer     & 5 & \textbf{15\,786} & \textbf{35\,510} \\
                  & SilverSpoon-KD & \textbf{6} & 14\,700 & 37\,333 \\
\addlinespace
Qwen3-8B (16 GB)  & DistillKit     & 2 & 11\,161 & \textbf{33\,027} \\
                  & TextBrewer     & 3 & \textbf{11\,514} & 34\,119 \\
                  & SilverSpoon-KD & \textbf{4} & 9\,975 & 34\,842 \\
\addlinespace
Qwen3-14B (28 GB) & DistillKit     & \multicolumn{3}{c}{\textit{OOM}} \\
                  & TextBrewer     & 1 & 6\,217 & \textbf{37\,379} \\
                  & SilverSpoon-KD & \textbf{3} & \textbf{6\,812} & 37\,397 \\
\bottomrule
\end{tabular}
\end{table}

%% file: static/standard_training.tex
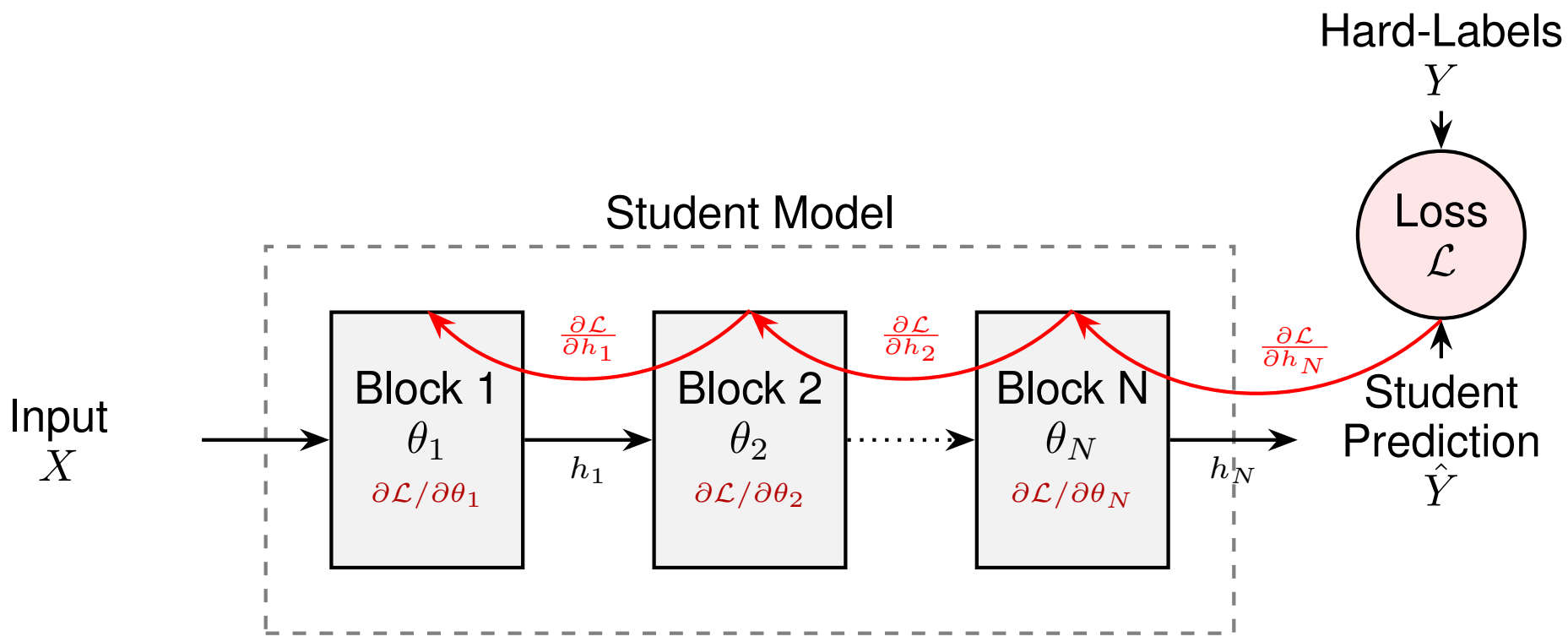
\begin{figure}[htbp]
    \centering
    \begin{tikzpicture}[
        node distance=1cm,
        >=Stealth,
        font=\sffamily
      ]

    \tikzset{
      block/.style={
        rectangle, draw, thick, fill=gray!10,
        minimum height=2cm, minimum width=1.5cm, align=center
      },
      data/.style={align=center, text width=2cm},
      loss/.style={
        circle, draw, thick, fill=red!10,
        minimum size=1cm, align=center
      },
      line/.style={draw, thick, -Stealth},
      backprop/.style={draw, thick, red, -Stealth},
      surround/.style={
        draw, thick, dashed, gray, rectangle,
        inner sep=0.5cm, label=above:Student Model
      }
    }

    \node [data] (in_x) {Input \\ $X$};
    \node [block, right=of in_x] (block1) {Block 1 \\ $\theta_1$ \\ \pgrad{\partial \mathcal{L}/\partial \theta_1}};
    \node [block, right=of block1] (block2) {Block 2 \\ $\theta_2$ \\ \pgrad{\partial \mathcal{L}/\partial \theta_2}};
    \node [block, right=of block2] (blockn) {Block N \\ $\theta_N$ \\ \pgrad{\partial \mathcal{L}/\partial \theta_N}};
    \node [data, right=of blockn] (output) {Student \\ Prediction \\ $\hat{Y}$};
    \node [loss, above=0.3cm of output] (loss) {Loss \\ $\mathcal{L}$};
    \node [data, above=0.3cm of loss] (true_y) {Hard-Labels \\ $Y$};

    \begin{scope}[on background layer]
      \node [surround, fit=(block1) (block2) (blockn)] (model_box) {};
    \end{scope}

    \draw [line] (in_x) -- (block1);
    \draw [line] (block1) -- node[below,font=\scriptsize]{$h_1$} (block2);
    \draw [line, dotted] (block2) -- (blockn);
    \draw [line] (blockn) -- node[below,font=\scriptsize]{$h_N$} (output);
    \draw [line] (output) -- (loss);
    \draw [line] (true_y) -- (loss);

    \draw [backprop] (loss.south) to[bend left=45]
        node[above,font=\tiny,pos=0.4]{$\frac{\partial \mathcal{L}}{\partial h_N}$} (blockn.north);
    \draw [backprop] (blockn.north) to[bend left=45]
        node[above,font=\tiny,pos=0.5]{$\frac{\partial \mathcal{L}}{\partial h_2}$} (block2.north);
    \draw [backprop] (block2.north) to[bend left=45]
        node[above,font=\tiny,pos=0.5]{$\frac{\partial \mathcal{L}}{\partial h_1}$} (block1.north);

    \end{tikzpicture}

    \caption{Standard training from hard labels (no teacher). The student's learnable parameters $\theta$ are updated via backpropagation to minimize the loss between its prediction and the ground-truth labels.}
    \label{fig:standard_training}
\end{figure}

%% file: static/reskd_visualization.tex
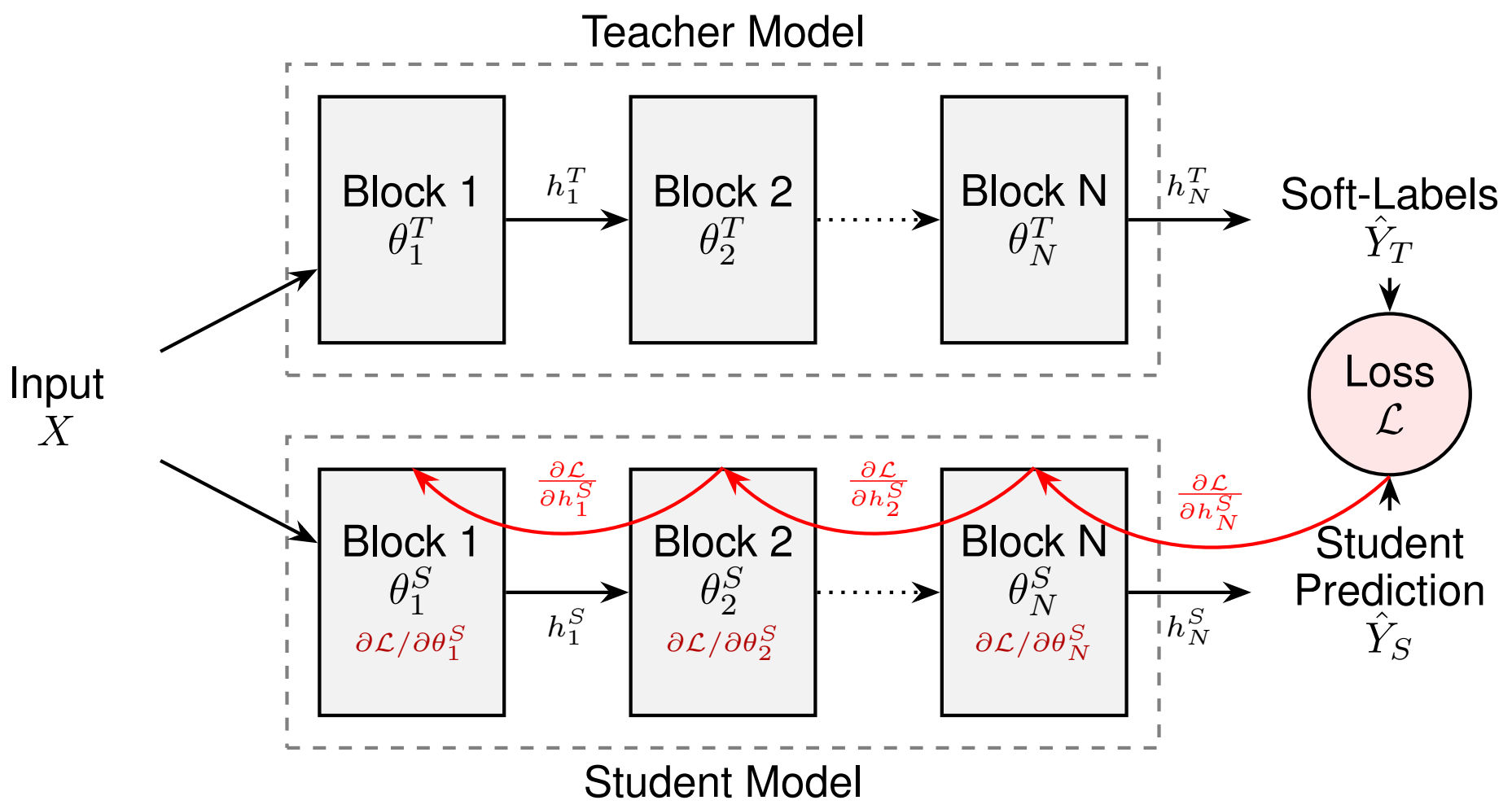
\begin{figure}[htbp]
    \centering

    \begin{tikzpicture}[
        node distance=1cm and 1cm,
        >=Stealth,
        font=\sffamily
      ]

    \tikzset{
      block/.style={
        rectangle, draw, thick, fill=gray!10,
        minimum height=2cm, minimum width=1.5cm, align=center
      },
      data/.style={align=center, text width=2cm},
      loss/.style={
        circle, draw, thick, fill=red!10,
        minimum size=1cm, align=center
      },
      line/.style={draw, thick, -Stealth},
      backprop/.style={draw, thick, red, -Stealth},
      surround/.style={
        draw, thick, dashed, gray, rectangle,
        inner sep=0.25cm
      }
    }

    \node [block] (s_block1) {Block 1 \\ $\theta_1^S$ \\ \pgrad{\partial \mathcal{L}/\partial \theta_1^S}};
    \node [block, right=of s_block1] (s_block2) {Block 2 \\ $\theta_2^S$ \\ \pgrad{\partial \mathcal{L}/\partial \theta_2^S}};
    \node [block, right=of s_block2] (s_blockn) {Block N \\ $\theta_N^S$ \\ \pgrad{\partial \mathcal{L}/\partial \theta_N^S}};
    \node [data, right=of s_blockn] (s_output) {Student \\ Prediction \\ $\hat{Y}_S$};

    \node [block, above=of s_block1] (t_block1) {Block 1 \\ $\theta_1^T$};
    \node [block, above=of s_block2] (t_block2) {Block 2 \\ $\theta_2^T$};
    \node [block, above=of s_blockn] (t_blockn) {Block N \\ $\theta_N^T$};
    \node [data, right=of t_blockn] (t_output) {Soft-Labels \\ $\hat{Y}_T$};

    \path (s_block1.west) -- (t_block1.west) coordinate[midway] (mid_point);
    \node [data, left=of mid_point] (in_x) {Input \\ $X$};

    \path (s_output) -- (t_output) node[midway, loss] (loss) {Loss \\ $\mathcal{L}$};

    \begin{scope}[on background layer]
      \node [surround, fit=(t_block1) (t_block2) (t_blockn), label=above:Teacher Model] {};
      \node [surround, fit=(s_block1) (s_block2) (s_blockn), label=below:Student Model] {};
    \end{scope}

    \draw [line] (in_x) -- (t_block1);
    \draw [line] (in_x) -- (s_block1);
    \draw [line] (t_block1) -- node[above,font=\scriptsize]{$h_1^T$} (t_block2);
    \draw [line, dotted] (t_block2) -- (t_blockn);
    \draw [line] (t_blockn) -- node[above,font=\scriptsize]{$h_N^T$} (t_output);
    \draw [line] (s_block1) -- node[below,font=\scriptsize]{$h_1^S$} (s_block2);
    \draw [line, dotted] (s_block2) -- (s_blockn);
    \draw [line] (s_blockn) -- node[below,font=\scriptsize]{$h_N^S$} (s_output);
    \draw [line] (s_output.north) -- (loss.south);
    \draw [line] (t_output.south) -- (loss.north);

    \draw [backprop] (loss.south) to[bend left=45]
        node[above,font=\tiny,pos=0.5]{$\frac{\partial \mathcal{L}}{\partial h_N^S}$} (s_blockn.north);
    \draw [backprop] (s_blockn.north) to[bend left=45]
        node[above,font=\tiny,pos=0.5]{$\frac{\partial \mathcal{L}}{\partial h_2^S}$} (s_block2.north);
    \draw [backprop] (s_block2.north) to[bend left=45]
        node[above,font=\tiny,pos=0.5]{$\frac{\partial \mathcal{L}}{\partial h_1^S}$} (s_block1.north);

    \end{tikzpicture}
    \caption{response-based KD: a teacher model produces soft-labels that the student learns to match. Only the student's parameters $\theta^S$ are updated; the teacher is frozen.}
    \label{fig:reskd_visualization}
\end{figure}

%% file: static/hkd_visualization.tex
\begin{figure}[htbp]
    \centering

    \begin{tikzpicture}[
        node distance=2cm and 1cm,
        >=Stealth,
        font=\sffamily
      ]

    \tikzset{
      block/.style={
        rectangle, draw, thick, fill=gray!10,
        minimum height=2cm, minimum width=1.5cm, align=center
      },
      data/.style={align=center, text width=2cm},
      loss/.style={
        circle, draw, thick, fill=red!10,
        minimum size=1cm, align=center
      },
      line/.style={draw, thick, -Stealth},
      backprop/.style={draw, thick, red, -Stealth},
      surround/.style={
        draw, thick, dashed, gray, rectangle,
        inner sep=0.25cm
      }
    }

    \node [block] (s_block1) {Block 1 \\ $\theta_1^S$ \\ \pgrad{\partial \mathcal{L}_1/\partial \theta_1^S} \\ \pgrad{+\,\partial \mathcal{L}_2/\partial \theta_1^S} \\ \pgrad{+\,\partial \mathcal{L}_N/\partial \theta_1^S}};
    \node [block, right=of s_block1] (s_block2) {Block 2 \\ $\theta_2^S$ \\ \pgrad{\partial \mathcal{L}_2/\partial \theta_2^S} \\ \pgrad{+\,\partial \mathcal{L}_N/\partial \theta_2^S}};
    \node [block, right=of s_block2] (s_blockn) {Block N \\ $\theta_N^S$ \\ \pgrad{\partial \mathcal{L}_N/\partial \theta_N^S}};
    \node [data, right=of s_blockn] (s_output) {Student \\ Prediction \\ $\hat{Y}_S$};

    \node [block, above=of s_block1] (t_block1) {Block 1 \\ $\theta_1^T$};
    \node [block, above=of s_block2] (t_block2) {Block 2 \\ $\theta_2^T$};
    \node [block, above=of s_blockn] (t_blockn) {Block N \\ $\theta_N^T$};
    \node [data, right=of t_blockn] (t_output) {Soft-Labels \\ $\hat{Y}_T$};

    \path (s_block1.west) -- (t_block1.west) coordinate[midway] (mid_point);
    \node [data, left=of mid_point] (in_x) {Input \\ $X$};

    \path (s_output) -- (t_output) node[midway, loss] (loss) {Loss \\ $\mathcal{L}_{N}$};

    \node [loss] (loss1) at ($(s_block1.north east)!.5!(t_block1.south east)$) [xshift=0.5cm] {$\mathcal{L}_1$};
    \node [loss] (loss2) at ($(s_block2.north east)!.5!(t_block2.south east)$) [xshift=0.5cm] {$\mathcal{L}_2$};

    \begin{scope}[on background layer]
      \node [surround, fit=(t_block1) (t_block2) (t_blockn), label=above:Teacher Model] {};
      \node [surround, fit=(s_block1) (s_block2) (s_blockn), label=below:Student Model] {};
    \end{scope}

    \draw [line] (in_x) -- (t_block1);
    \draw [line] (in_x) -- (s_block1);
    \draw [line] (t_block1) -- node[above,font=\scriptsize]{$h_1^T$} (t_block2);
    \draw [line, dotted] (t_block2) -- (t_blockn);
    \draw [line] (t_blockn.east) -- node[above,font=\scriptsize]{$h_N^T$} (t_output.west);
    \draw [line] (s_block1) -- node[below,font=\scriptsize]{$h_1^S$} (s_block2);
    \draw [line, dotted] (s_block2) -- (s_blockn);
    \draw [line] (s_blockn) -- node[below,font=\scriptsize]{$h_N^S$} (s_output);
    \draw [line] (s_output) -- (loss);
    \draw [line] (t_output) -- (loss);

    \draw [line] (s_block1.east) -- (loss1.south);
    \draw [line] (t_block1.east) -- (loss1.north);
    \draw [line] (s_block2.east) -- (loss2.south);
    \draw [line] (t_block2.east) -- (loss2.north);

    \draw [backprop] (loss.south) to[bend right=0]
        node[above,font=\tiny,pos=0.5]{$\frac{\partial \mathcal{L}_{N}}{\partial h_N^S}$} (s_blockn.south);
    \draw [backprop] (s_blockn.south) to[bend right=45]
        node[below,font=\tiny,pos=0.5]{$\frac{\partial \mathcal{L}_{N}}{\partial h_2^S}$} (s_block2.south);
    \draw [backprop] (s_block2.south) to[bend right=45]
        node[below,font=\tiny,pos=0.5]{$\frac{\partial \mathcal{L}_{N}}{\partial h_1^S}$} (s_block1.south);

    \draw [backprop] (loss2.south) to[bend left=0]
        node[above,font=\tiny,pos=0.6]{$\frac{\partial \mathcal{L}_{2}}{\partial h_2^S}$} (s_block2.north);
    \draw [backprop] (s_block2.north) to[bend left=20]
        node[below,font=\tiny,pos=0.5]{$\frac{\partial \mathcal{L}_{2}}{\partial h_1^S}$} (s_block1.north);

    \draw [backprop] (loss1.south) to[bend left=0]
        node[above,font=\tiny,pos=0.6]{$\frac{\partial \mathcal{L}_{1}}{\partial h_1^S}$} (s_block1.north);

    \end{tikzpicture}

    \caption{holistic KD, a \gls{featkd} strategy. The student performs a full end-to-end forward pass, and intermediate block outputs are compared with the corresponding teacher block outputs. Losses $\mathcal{L}_{1 \ldots N}$ all contribute gradients to $\theta_1^S$ (since all subsequent outputs depend on Block~1), losses $\mathcal{L}_{2 \ldots N}$ update $\theta_2^S$, while $\theta_N^S$ only receives gradients from $\mathcal{L}_{N}$.}
    \label{fig:hkd_visualization}
\end{figure}
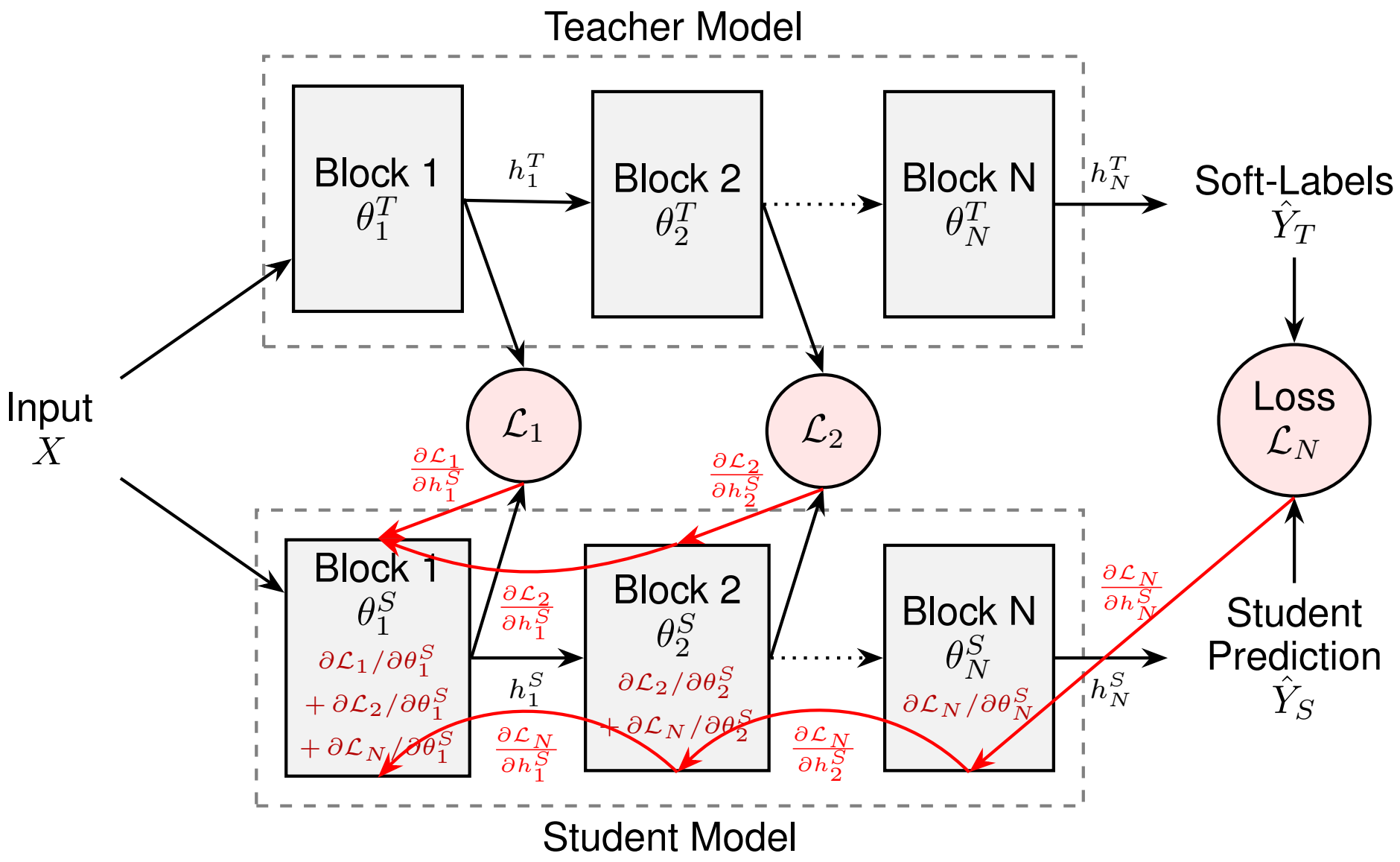

%% file: static/bkd_visualization.tex
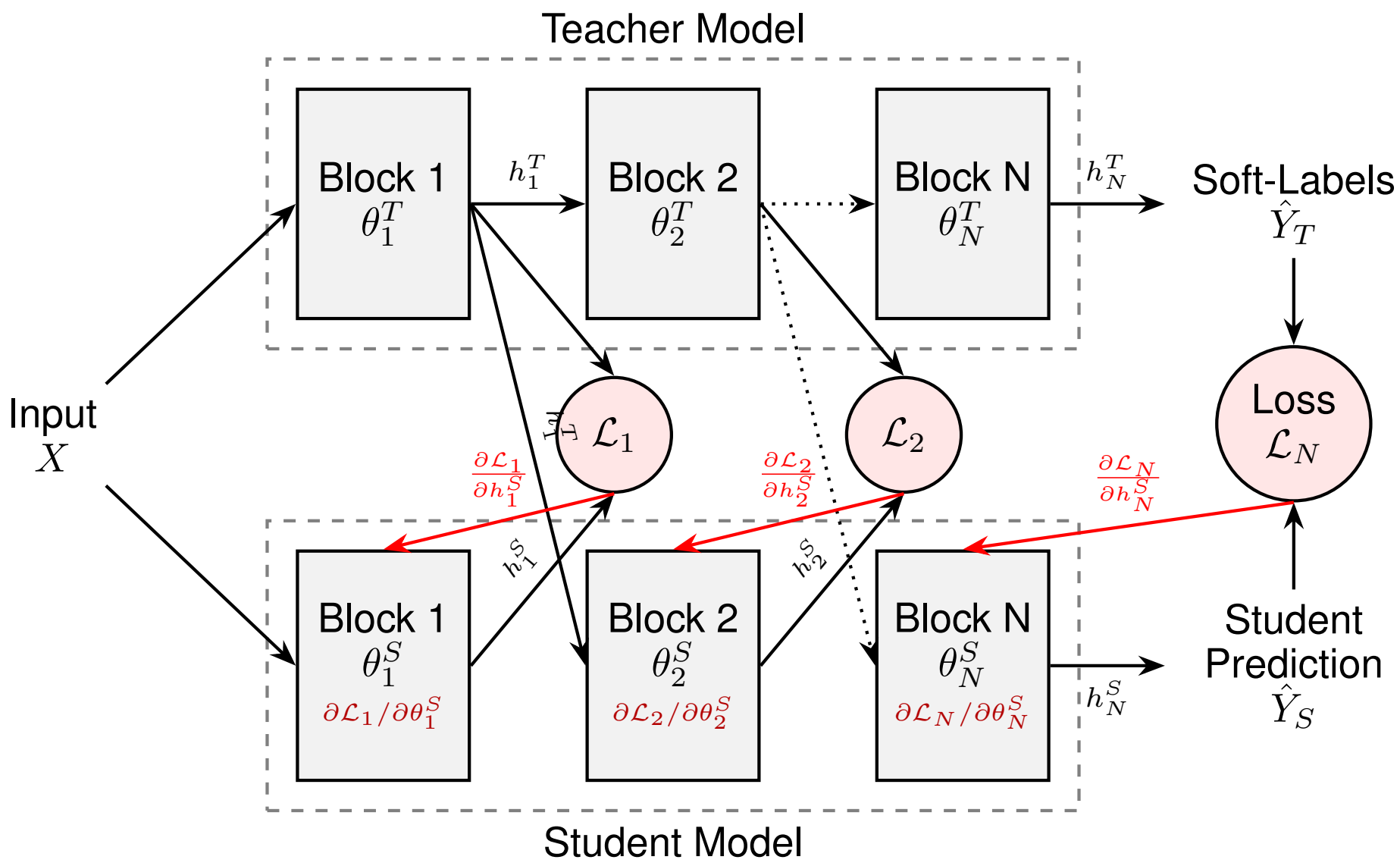
\begin{figure}[htbp]
    \centering

    \begin{tikzpicture}[
        node distance=2cm and 1cm,
        >=Stealth,
        font=\sffamily
      ]

    \tikzset{
      block/.style={
        rectangle, draw, thick, fill=gray!10,
        minimum height=2cm, minimum width=1.5cm, align=center
      },
      data/.style={align=center, text width=2cm},
      loss/.style={
        circle, draw, thick, fill=red!10,
        minimum size=1cm, align=center
      },
      line/.style={draw, thick, -Stealth},
      backprop/.style={draw, thick, red, -Stealth},
      surround/.style={
        draw, thick, dashed, gray, rectangle,
        inner sep=0.25cm
      },
      int_loss/.style={
        circle, draw, thick, fill=red!10,
        minimum size=1cm, align=center
      },
    }

    \node [block] (s_block1) {Block 1 \\ $\theta_1^S$ \\ \pgrad{\partial \mathcal{L}_1/\partial \theta_1^S}};
    \node [block, right=of s_block1] (s_block2) {Block 2 \\ $\theta_2^S$ \\ \pgrad{\partial \mathcal{L}_2/\partial \theta_2^S}};
    \node [block, right=of s_block2] (s_blockn) {Block N \\ $\theta_N^S$ \\ \pgrad{\partial \mathcal{L}_N/\partial \theta_N^S}};
    \node [data, right=of s_blockn] (s_output) {Student \\ Prediction \\ $\hat{Y}_S$};

    \node [block, above=of s_block1] (t_block1) {Block 1 \\ $\theta_1^T$};
    \node [block, above=of s_block2] (t_block2) {Block 2 \\ $\theta_2^T$};
    \node [block, above=of s_blockn] (t_blockn) {Block N \\ $\theta_N^T$};
    \node [data, right=of t_blockn] (t_output) {Soft-Labels \\ $\hat{Y}_T$};

    \path (s_block1.west) -- (t_block1.west) coordinate[midway] (mid_point);
    \node [data, left=of mid_point] (in_x) {Input \\ $X$};

    \path (s_output) -- (t_output) node[midway, loss] (loss) {Loss \\ $\mathcal{L}_{N}$};

    \node [int_loss] (loss1) at ($(s_block1.north east)!.5!(t_block1.south east)$) [xshift=1.25cm] {$\mathcal{L}_1$};
    \node [int_loss] (loss2) at ($(s_block2.north east)!.5!(t_block2.south east)$) [xshift=1.25cm] {$\mathcal{L}_2$};

    \begin{scope}[on background layer]
      \node [surround, fit=(t_block1) (t_block2) (t_blockn), label=above:Teacher Model] {};
      \node [surround, fit=(s_block1) (s_block2) (s_blockn), label=below:Student Model] {};
    \end{scope}

    \draw [line] (in_x) -- (t_block1.west);
    \draw [line] (in_x) -- (s_block1.west);
    \draw [line] (t_block1) -- node[above,font=\scriptsize]{$h_1^T$} (t_block2);
    \draw [line, dotted] (t_block2) -- (t_blockn);
    \draw [line] (t_blockn.east) -- node[above,font=\scriptsize]{$h_N^T$} (t_output.west);
    \draw [line] (t_block1.east) -- node[midway,sloped,above,font=\tiny]{$h_1^T$} (s_block2.west);
    \draw [line, dotted] (t_block2.east) -- (s_blockn.west);
    \draw [line] (s_blockn.east) -- node[below,font=\scriptsize]{$h_N^S$} (s_output);
    \draw [line] (s_output) -- (loss);
    \draw [line] (t_output.south) -- (loss.north);

    \draw [line] (s_block1.east) -- node[midway,sloped,above,font=\tiny]{$h_1^S$} (loss1.south);
    \draw [line] (t_block1.east) -- (loss1.north);
    \draw [line] (s_block2.east) -- node[midway,sloped,above,font=\tiny]{$h_2^S$} (loss2.south);
    \draw [line] (t_block2.east) -- (loss2.north);

    \draw [backprop] (loss.south) to[bend right=0]
        node[above,font=\tiny,pos=0.5]{$\frac{\partial \mathcal{L}_{N}}{\partial h_N^S}$} (s_blockn.north);
    \draw [backprop] (loss2.south) to[bend right=0]
        node[above,font=\tiny,pos=0.5]{$\frac{\partial \mathcal{L}_{2}}{\partial h_2^S}$} (s_block2.north);
    \draw [backprop] (loss1.south) to[bend right=0]
        node[above,font=\tiny,pos=0.5]{$\frac{\partial \mathcal{L}_{1}}{\partial h_1^S}$} (s_block1.north);

    \end{tikzpicture}

    \caption{blockwise KD, a \gls{featkd} strategy. Each student block receives the same input as the corresponding teacher block, and loss is calculated by comparing the resulting outputs. Since each student block is only used to calculate its own output and is not used for subsequent student blocks, only the configuration of student learnable parameters $\theta_i^S$ influences the loss $\mathcal{L}_i$, meaning backpropagation does not propagate to non-involved earlier student blocks.}
    \label{fig:bkd_visualization}
\end{figure}

%% file: static/multistage_visualization.tex
\begin{figure}[htbp]
    \centering

    \begin{tikzpicture}[
        node distance=0.6cm and 0.8cm,
        >=Stealth,
        font=\sffamily
      ]

    \tikzset{
      sblock/.style={
        rectangle, draw, thick, fill=gray!10,
        minimum height=0.7cm, minimum width=0.85cm, align=center,
        font=\scriptsize
      },
      stublock/.style={
        rectangle, draw, thick, fill=gray!10,
        minimum height=0.7cm, minimum width=0.85cm, align=center,
        font=\scriptsize
      },
      frozblock/.style={
        rectangle, draw, thick, fill=gray!22, draw=gray!60,
        minimum height=0.7cm, minimum width=0.85cm, align=center,
        font=\scriptsize, text=gray!50!black
      },
      adapter/.style={
        rectangle, draw, thick, fill=cMulti!22, draw=cMulti!75!black,
        minimum height=0.32cm, minimum width=0.85cm, align=center,
        font=\tiny
      },
      data/.style={align=center, text width=1.1cm, font=\scriptsize},
      mloss/.style={
        circle, draw, thick, fill=red!10,
        minimum size=0.4cm, align=center, font=\tiny,
        inner sep=0pt
      },
      line/.style={draw, thick, -Stealth},
      backprop/.style={draw, thick, red, -Stealth},
      surround/.style={
        draw, thick, dashed, gray, rectangle,
        inner sep=0.25cm
      },
      stagebox/.style={
        draw, thick, rounded corners=4pt, inner sep=0.2cm,
        line width=0.6pt
      },
      handoff/.style={
        draw, very thick, -Stealth, cMulti!75!black, line width=1.2pt
      },
      stagelabel/.style={
        font=\sffamily\bfseries\small, align=left
      },
      stagenote/.style={
        font=\sffamily\scriptsize, align=left, text=gray!30!black
      },
    }

    \begin{scope}[local bounding box=stage1]
      \node [sblock] (s1_t1) {$f_1^T$};
      \node [sblock, right=of s1_t1] (s1_t2) {$f_2^T$};
      \node [sblock, right=of s1_t2] (s1_t3) {$f_3^T$};
      \node [stublock, below=0.7cm of s1_t1] (s1_s1) {$f_1^S$};
      \node [stublock, right=of s1_s1] (s1_s2) {$f_2^S$};
      \node [stublock, right=of s1_s2] (s1_s3) {$f_3^S$};

      \node [mloss] (s1_l1) at ($(s1_t1)!.5!(s1_s1)$) [xshift=1.2cm] {$\mathcal{L}_1$};
      \node [mloss] (s1_l2) at ($(s1_t2)!.5!(s1_s2)$) [xshift=1.2cm] {$\mathcal{L}_2$};
      \node [mloss] (s1_l3) at ($(s1_t3)!.5!(s1_s3)$) [xshift=1.2cm] {$\mathcal{L}_3$};

      \path (s1_t1.west) -- (s1_s1.west) coordinate[midway] (s1_in);
      \node [data, left=0.35cm of s1_in] (s1_x) {Input $X$};

      \draw [line] (s1_x) -- (s1_t1);
      \draw [line] (s1_x) -- (s1_s1);
      \draw [line] (s1_t1) -- (s1_t2);
      \draw [line] (s1_t2) -- (s1_t3);
      \draw [line] (s1_t1.south east) -- (s1_s2.north west);
      \draw [line] (s1_t2.south east) -- (s1_s3.north west);
      \draw [line] (s1_s1.east) -- (s1_l1.south);
      \draw [line] (s1_t1.east) -- (s1_l1.north);
      \draw [line] (s1_s2.east) -- (s1_l2.south);
      \draw [line] (s1_t2.east) -- (s1_l2.north);
      \draw [line] (s1_s3.east) -- (s1_l3.south);
      \draw [line] (s1_t3.east) -- (s1_l3.north);

      \draw [backprop] (s1_l1.south) to[bend right=0] (s1_s1.north east);
      \draw [backprop] (s1_l2.south) to[bend right=0] (s1_s2.north east);
      \draw [backprop] (s1_l3.south) to[bend right=0] (s1_s3.north east);

      \begin{scope}[on background layer]
        \node [surround, fit=(s1_t1) (s1_t2) (s1_t3)] (s1_tbox) {};
        \node [surround, fit=(s1_s1) (s1_s2) (s1_s3)] (s1_sbox) {};
      \end{scope}

      \node [font=\scriptsize\sffamily\itshape, text=gray!40!black,
             anchor=east, align=right, text width=1.4cm]
          at (s1_tbox.west) [xshift=-0.1cm] {Teacher Model};
      \node [font=\scriptsize\sffamily\itshape, text=gray!40!black,
             anchor=east, align=right, text width=1.4cm]
          at (s1_sbox.west) [xshift=-0.1cm] {Student Model};
    \end{scope}

    \node [stagelabel, right=0.35cm of s1_t3, anchor=west, xshift=0.7cm]
        (s1_label) {Stage 1: blockwise KD};
    \node [stagenote, below=0.05cm of s1_label.south west, anchor=north west, text width=4.2cm]
        (s1_note) {init: scratch \\ trains: all $\theta_i^S$, projectors $\Phi$ \\ objective: per-block $\mathcal{L}_i$};

    \begin{scope}[local bounding box=stage2, shift={(0,-4.6cm)}]
      \node [sblock] (s2_t1) {$f_1^T$};
      \node [sblock, right=of s2_t1] (s2_t2) {$f_2^T$};
      \node [sblock, right=of s2_t2] (s2_t3) {$f_3^T$};
      \node [stublock, below=0.7cm of s2_t1] (s2_s1) {$f_1^S$};
      \node [stublock, right=of s2_s1] (s2_s2) {$f_2^S$};
      \node [stublock, right=of s2_s2] (s2_s3) {$f_3^S$};

      \node [mloss] (s2_l1) at ($(s2_t1)!.5!(s2_s1)$) [xshift=1.2cm] {$\mathcal{L}_1$};
      \node [mloss] (s2_l2) at ($(s2_t2)!.5!(s2_s2)$) [xshift=1.2cm] {$\mathcal{L}_2$};
      \node [mloss] (s2_l3) at ($(s2_t3)!.5!(s2_s3)$) [xshift=1.2cm] {$\mathcal{L}_3$};

      \path (s2_t1.west) -- (s2_s1.west) coordinate[midway] (s2_in);
      \node [data, left=0.35cm of s2_in] (s2_x) {Input $X$};

      \draw [line] (s2_x) -- (s2_t1);
      \draw [line] (s2_x) -- (s2_s1);
      \draw [line] (s2_t1) -- (s2_t2);
      \draw [line] (s2_t2) -- (s2_t3);
      \draw [line] (s2_s1) -- (s2_s2);
      \draw [line] (s2_s2) -- (s2_s3);
      \draw [line] (s2_s1.east) -- (s2_l1.south);
      \draw [line] (s2_t1.east) -- (s2_l1.north);
      \draw [line] (s2_s2.east) -- (s2_l2.south);
      \draw [line] (s2_t2.east) -- (s2_l2.north);
      \draw [line] (s2_s3.east) -- (s2_l3.south);
      \draw [line] (s2_t3.east) -- (s2_l3.north);

      \draw [backprop] (s2_l3.south) to[bend right=0] (s2_s3.north east);
      \draw [backprop] (s2_s3.south) to[bend right=45] (s2_s2.south);
      \draw [backprop] (s2_s2.south) to[bend right=45] (s2_s1.south);

      \begin{scope}[on background layer]
        \node [surround, fit=(s2_t1) (s2_t2) (s2_t3)] (s2_tbox) {};
        \node [surround, fit=(s2_s1) (s2_s2) (s2_s3)] (s2_sbox) {};
      \end{scope}
    \end{scope}

    \node [stagelabel, right=0.35cm of s2_t3, anchor=west, xshift=0.7cm]
        (s2_label) {Stage 2: holistic KD};
    \node [stagenote, below=0.05cm of s2_label.south west, anchor=north west, text width=4.2cm]
        (s2_note) {init: $\theta^{S}_{(1)}$ from Stage 1 \\ trains: all $\theta_i^S$ end-to-end \\ objective: $\sum_i w_i\,\mathcal{L}_i$};

    \begin{scope}[local bounding box=stage3, shift={(0,-8.8cm)}]
      \node [sblock] (s3_t1) {$f_1^T$};
      \node [sblock, right=of s3_t1] (s3_t2) {$f_2^T$};
      \node [sblock, right=of s3_t2] (s3_t3) {$f_3^T$};
      \node [frozblock, below=0.95cm of s3_t1] (s3_s1) {$f_1^S$};
      \node [frozblock, right=of s3_s1] (s3_s2) {$f_2^S$};
      \node [frozblock, right=of s3_s2] (s3_s3) {$f_3^S$};
      \node [adapter, above=0cm of s3_s1] (s3_a1) {$\Delta_1^S$};
      \node [adapter, above=0cm of s3_s2] (s3_a2) {$\Delta_2^S$};
      \node [adapter, above=0cm of s3_s3] (s3_a3) {$\Delta_3^S$};

      \node [mloss] (s3_loss) at ($(s3_t3)!.5!(s3_s3)$) [xshift=1.2cm] {$\mathcal{L}$};

      \path (s3_t1.west) -- (s3_s1.west) coordinate[midway] (s3_in);
      \node [data, left=0.35cm of s3_in] (s3_x) {Input $X$};

      \draw [line] (s3_x) -- (s3_t1);
      \draw [line] (s3_x) -- (s3_s1);
      \draw [line] (s3_t1) -- (s3_t2);
      \draw [line] (s3_t2) -- (s3_t3);
      \draw [line] (s3_s1) -- (s3_s2);
      \draw [line] (s3_s2) -- (s3_s3);
      \draw [line] (s3_t3.east) -- (s3_loss.north);
      \draw [line] (s3_s3.east) -- (s3_loss.south);

      \draw [backprop] (s3_loss.south) to[bend right=0] (s3_a3.north east);
      \draw [backprop] (s3_a3.west) to[bend left=15] (s3_a2.east);
      \draw [backprop] (s3_a2.west) to[bend left=15] (s3_a1.east);

      \begin{scope}[on background layer]
        \node [surround, fit=(s3_t1) (s3_t2) (s3_t3)] (s3_tbox) {};
        \node [surround, fit=(s3_s1) (s3_s2) (s3_s3) (s3_a1) (s3_a2) (s3_a3)] (s3_sbox) {};
      \end{scope}
    \end{scope}

    \node [stagelabel, right=0.35cm of s3_t3, anchor=west, xshift=0.7cm]
        (s3_label) {Stage 3: LoRA fine-tune (response-based)};
    \node [stagenote, below=0.05cm of s3_label.south west, anchor=north west, text width=4.6cm]
        (s3_note) {init: $\theta^{S}_{(2)}$ frozen, adapters $\Delta_i^S$ random \\ trains: $\Delta_i^S$ only \\ objective: response loss $\mathcal{L}$};

    \draw [handoff] (s1_sbox.south) -- (s2_tbox.north |- s2_tbox.north)
        node[midway, font=\scriptsize\sffamily, fill=white, inner sep=2pt,
             text=cMulti!50!black]
            {init Stage 2 from $\theta^{S}_{(1)}$};
    \draw [handoff] (s2_sbox.south) -- (s3_tbox.north |- s3_tbox.north)
        node[midway, font=\scriptsize\sffamily, fill=white, inner sep=2pt,
             text=cMulti!50!black]
            {init Stage 3 from $\theta^{S}_{(2)}$};

    \end{tikzpicture}

    \caption{Multi-stage distillation pipeline (BKD~$\rightarrow$~HKD~$\rightarrow$~LoRA, the headline GPT-2 Compression configuration of Appendix~\ref{exp:gpt2_dist}). Each stage reuses the same frozen teacher $\mathcal{T}$ and the trained student from the prior stage, switching distiller class to apply a different paradigm; the trained student weights $\theta^{S}_{(k)}$ from stage $k$ become the initialization for stage $k{+}1$. Stage 1 (BKD) trains all student blocks under per-block teacher-forced losses with no cross-block gradient flow; Stage 2 (HKD) keeps Stage 1's per-layer alignments but propagates gradients end-to-end through the full student chain; Stage 3 freezes the post-Stage-2 student weights, attaches LoRA adapters $\Delta_i^S$ on top, and trains only the adapters under a single response-based loss (no feature alignments). The optimizer state is reset at every stage boundary. Per-paradigm internals (forward chains, parameter-gradient annotations, alignment-loss formulae) are detailed in Figures~\ref{fig:reskd_visualization}--\ref{fig:bkd_visualization}.}
    \label{fig:multistage_visualization}
\end{figure}
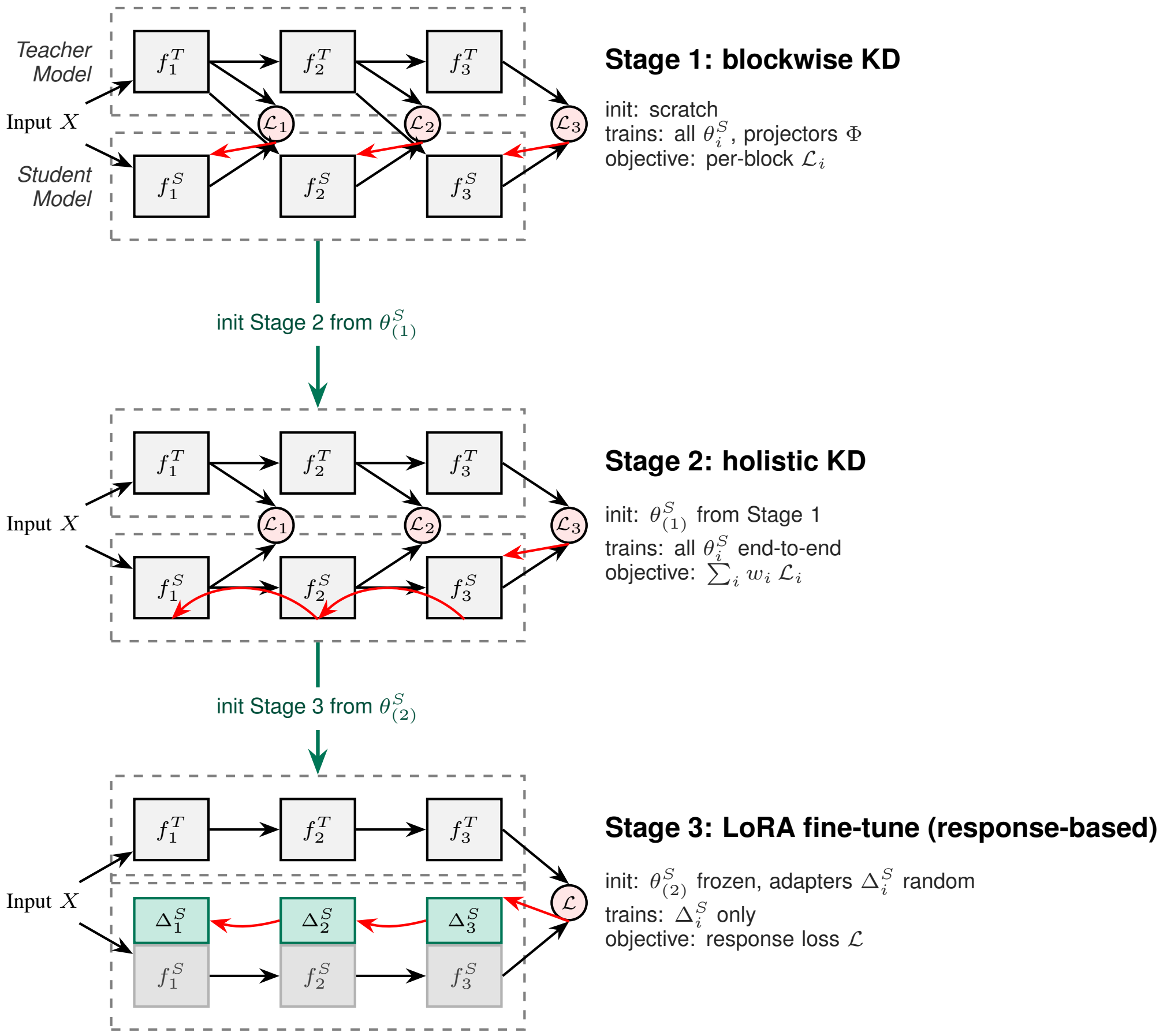

%% file: generated/vision_bkd.tex
\begin{table}[htbp]
\centering\footnotesize
\caption{\textbf{VGG Compression on CIFAR-10.} BKD~$\rightarrow$~ResKD wins accuracy at 90.42\% (+3.16pp over teacher). Bold = best per column among non-teacher rows.}
\label{tab:vision_bkd}
\begin{tabular}{@{}lcccc@{}}
\toprule
Method & Test Acc.\ (\%) & Peak Mem (MB) & Steps/s & Time (h) \\
\midrule
Baseline (no teacher)             & 82.69 & \textbf{692} & 29.09 & 0.37 \\
\gls{bkd}                         & 72.49 & 9\,991 & \textbf{32.60} & 0.33 \\
\gls{bkd} $\rightarrow$ Fine-tune & 88.92 & \textbf{692} & 28.45 & \textbf{0.23} \\
\gls{bkd} $\rightarrow$ \gls{hkd} & 87.97 & 1\,200 & 14.90 & 0.73 \\
\gls{reskd}                       & 84.35 & 720 & 20.16 & 0.54 \\
\gls{bkd} $\rightarrow$ \gls{reskd} & \textbf{90.42} & 1\,048 & 18.16 & 0.60 \\
\midrule
Teacher (VGG-16)                  & 87.26 & 734 & 81.54 & 0.13 \\
\bottomrule
\end{tabular}
\end{table}

%% file: generated/relkd.tex
\begin{table}[htbp]
\centering\footnotesize
\caption{\textbf{VGG Relational KD on CIFAR-100} (ResNet-50~$\rightarrow$~VGG-11-BN). ResKD leads our results at 73.65\%; relation-based variants span 70.40--71.61\%. All paradigms reachable through configuration-only changes. Bold = best per column among non-teacher rows (within Ours column).}
\label{tab:relkd}
\begin{tabular}{@{}lccccc@{}}
\toprule
Method & Ours (\%) & Park et al.\ (\%) & Peak Mem (MB) & Steps/s & Time (h) \\
\midrule
Teacher (ResNet-50)                                & 77.81 & --- & 4\,212 & 7.89 & 2.75 \\
Baseline (VGG-11-BN, no teacher)                   & 65.16 & 71.26 & \textbf{750} & \textbf{32.20} & \textbf{0.67} \\
\gls{reskd} ($T{=}16$)                             & \textbf{73.65} & 74.26 & 1\,190 & 13.20 & 1.64 \\
\gls{hkd} (MSE)                                    & 72.78 & --- & 1\,190 & 12.97 & 1.67 \\
\gls{hkd} (\acrshort{relkd} distance)              & 70.40 & 72.27 & 1\,192 & 12.75 & 1.70 \\
\gls{hkd} (\acrshort{relkd} angle)                 & 71.01 & --- & 1\,192 & 12.33 & 1.76 \\
\gls{hkd} (\acrshort{relkd} dist.\ + angle)        & 71.61 & 72.97 & 1\,192 & 11.91 & 1.82 \\
\bottomrule
\end{tabular}
\end{table}

%% file: generated/bert_pretrain.tex
\begin{table}[htbp]
\centering\footnotesize
\caption{\textbf{BERT distillation} (pre-training + downstream MNLI). HKD wins MNLI at both compression levels (T6: 80.33\%, +3.04pp; T4-tiny: 70.44\%, +3.52pp); the paradigm spread fans from $\sim$1.3pp at T6 to 5.2pp at T4-tiny, where standalone BKD turns harmful. Bold = best per column within each student-size group; Teacher row excluded from competition.}
\label{tab:bert_pretrain}
\begin{tabular}{@{}llccccccc@{}}
\toprule
Student & Method & Align Loss & MLM PPL & MNLI (\%) & Peak Mem & Steps/s & Time \\
        &        & (Stage 1)  & (Stage 1) & (Stage 2) & (MB)     &         & (h)  \\
\midrule
T6 & Baseline & --- & \textbf{14.08} & 77.29 & \textbf{6\,588} & \textbf{6.45} & \textbf{0.86} \\
T6 & \gls{bkd} & \textbf{0.077} & 297.02 & 79.13 & 12\,334 & 3.31 & 1.68 \\
T6 & \gls{hkd} & 0.408 & 40.57 & \textbf{80.33} & 13\,774 & 2.45 & 2.27 \\
T6 & \gls{reskd} & 1.661 & 36.51 & 79.01 & 9\,746 & 1.94 & 2.86 \\
\addlinespace
T4-tiny & Baseline & --- & \textbf{24.42} & 66.92 & 5\,458 & \textbf{12.97} & \textbf{0.43} \\
T4-tiny & \gls{bkd} & \textbf{0.359} & 2545.72 & 66.10 & \textbf{2\,962} & 5.18 & 1.07 \\
T4-tiny & \gls{hkd} & 0.548 & 2802.10 & \textbf{70.44} & 3\,904 & 3.97 & 1.40 \\
T4-tiny & \gls{reskd} & 2.187 & 6413.97 & 67.83 & 8\,160 & 2.72 & 2.04 \\
\midrule
\multicolumn{2}{@{}l}{Teacher (BERT-base)} & --- & --- & 83.59 & --- & --- & --- \\
\bottomrule
\end{tabular}
\end{table}

%% file: generated/bert_downstream.tex
\begin{table}[htbp]
\centering\footnotesize
\caption{\textbf{BERT Downstream MNLI: SilverSpoon-KD vs.\ TextBrewer.} HKD~$\rightarrow$~FT matches the BERT-base teacher within 0.12pp at T6 (83.47\% vs.\ 83.59\%) and beats TextBrewer by 3.38pp; at T4-tiny TextBrewer narrowly leads (81.20\% vs.\ 80.87\%). Reference-only rows (Teacher, baselines, TextBrewer) lack our system metrics. The T4-tiny HKD~$\rightarrow$~FT row's MNLI-mm cell is empty by design.\protect\footnote{The bert\_T4\_tiny student uses a pruned architecture (hidden=312 with the teacher's 768-dim Q/K/V projections retained); stock \texttt{BertModel.from\_pretrained} silently reinitializes the mismatched layers, so post-hoc evaluation cannot recover MNLI-mm without reconstructing the architecture via \texttt{silverspoon\_kd.utils.prune\_model}. The matched value comes from the in-training \texttt{trainer.evaluate()} call. MNLI-mm typically tracks MNLI-m within ${\pm}0.5$pp on the BERT family.} Bold = best per column within each student-size group; Teacher and TextBrewer rows excluded from competition.}
\label{tab:bert_downstream}
\begin{tabular}{@{}lccccc@{}}
\toprule
Run & MNLI-m (\%) & MNLI-mm (\%) & Peak Mem (MB) & Steps/s & Time (h) \\
\midrule
Teacher (BERT-base) & 83.59 & 83.76 & --- & --- & --- \\
\midrule
\multicolumn{6}{@{}l}{\emph{T6 (${\approx}59\%$ of teacher)}} \\
\quad Baseline (no distillation)            & 78.47 & 78.99 & --- & --- & --- \\
\quad TextBrewer                            & 80.09 & 80.48 & --- & --- & --- \\
\quad SilverSpoon \gls{hkd} $\rightarrow$ FT & \textbf{83.47} & \textbf{83.93} & 6\,154 & 3.57 & 7.16 \\
\midrule
\multicolumn{6}{@{}l}{\emph{T4-tiny (${\approx}13\%$ of teacher)}} \\
\quad Baseline (no distillation)             & 31.82 & 31.82 & --- & --- & --- \\
\quad TextBrewer                             & 81.20 & 81.42 & --- & --- & --- \\
\quad SilverSpoon \gls{hkd} $\rightarrow$ FT  & \textbf{80.87} & --- & \textbf{2\,960} & 4.61 & 5.54 \\
\quad SilverSpoon \gls{bkd} $\rightarrow$ FT  & 66.01 & \textbf{65.79} & 3\,786 & \textbf{23.84} & \textbf{1.07} \\
\bottomrule
\end{tabular}
\end{table}

%% file: generated/gpt2_multistage.tex
\begin{table}[htbp]
\centering\footnotesize
\caption{\textbf{GPT-2 Compression.} The three-stage BKD~$\rightarrow$~HKD~$\rightarrow$~LoRA pipeline rescues standalone-BKD divergence (PPL 245k$\to$93) and yields the best LAMBADA (19.56\%, +28\% over baseline). Benchmarks: LAMBADA~\cite{paperno2016lambada}, HellaSwag~\cite{zellers2019hellaswag}. Bold = best per column among non-teacher rows.}
\label{tab:gpt2_multistage}
\begin{tabular}{@{}lcccccc@{}}
\toprule
Method & Wiki PPL & LAMBADA (\%) & HellaSwag (\%) & Peak Mem (MB) & Steps/s & Time (h) \\
\midrule
Baseline (no teacher)                              & 107.15 & 15.25 & \textbf{27.24} & \textbf{20\,848} & 1.78 & 9.36 \\
\gls{bkd}                                          & 245{,}281.09 & 0.00 & 25.48 & 27\,304 & \textbf{4.53} & \textbf{3.68} \\
\gls{hkd}                                          & 97.69 & 17.74 & 26.59 & 23\,288 & 1.15 & 28.21 \\
\gls{reskd}                                        & 133.99 & 6.48 & 26.74 & 25\,598 & 1.52 & 21.32 \\
\gls{bkd} $\rightarrow$ \gls{hkd}                  & 109.56 & 16.90 & 27.12 & 23\,210 & 1.95 & 16.68 \\
\gls{bkd} $\rightarrow$ \gls{hkd} $\rightarrow$ LoRA & \textbf{93.39} & \textbf{19.56} & 27.21 & 21\,720 & 1.75 & 9.52 \\
\midrule
Teacher (GPT-2 Small)                              & 37.37 & 32.56 & 31.14 & --- & --- & --- \\
\bottomrule
\end{tabular}
\end{table}

%% file: generated/gpt2_deepseek.tex
\begin{table}[htbp]
\centering\footnotesize
\caption{\textbf{Cross-Architecture Distillation} (GPT-2~$\rightarrow$~DeepSeek-V3, no shared architectural lineage). Standalone KD degrades LAMBADA below baseline; FT-recovery (HKD~$\rightarrow$~FT, ResKD~$\rightarrow$~FT) brings results back near baseline. Bold = best per column among non-teacher rows.}
\label{tab:gpt2_deepseek}
\begin{tabular}{@{}lcccccc@{}}
\toprule
Method & Wiki PPL & LAMBADA (\%) & HellaSwag (\%) & Peak Mem (MB) & Steps/s & Time (h) \\
\midrule
Baseline (no teacher)                  & 117.78 & \textbf{16.34} & 27.48 & 22\,484 & 2.57 & 6.48 \\
\gls{hkd}                              & 246.44 & 0.87 & 26.29 & 33\,978 & \textbf{7.29} & \textbf{2.29} \\
\gls{reskd}                            & 127.14 & 6.02 & 26.47 & 39\,804 & 0.38 & 43.82 \\
\gls{hkd} $\rightarrow$ Fine-tune      & 118.49 & 15.06 & \textbf{27.67} & \textbf{17\,706} & 2.60 & 6.41 \\
\gls{reskd} $\rightarrow$ Fine-tune    & \textbf{117.28} & 14.96 & 27.65 & 22\,484 & 2.60 & 6.41 \\
\midrule
Teacher (GPT-2 Small)                  & 37.37 & 32.56 & 31.14 & --- & --- & --- \\
\bottomrule
\end{tabular}
\end{table}

%% file: generated/gpt2_linear.tex
\begin{table}[htbp]
\centering\footnotesize
\caption{\textbf{Attention Linearization} (softmax~$\rightarrow$~LoLCATs hybrid attention). Only BKD~$\rightarrow$~FT (PPL 52.01) beats the LoLCATs no-KD floor (53.49); standalone KDs push PPL above 230. Granular sub-module targeting makes this attention-only alignment accessible. Bold = best per column among non-teacher, non-LoLCATs-baseline rows.}
\label{tab:gpt2_linear}
\begin{tabular}{@{}lcccccc@{}}
\toprule
Method & Wiki PPL & LAMBADA (\%) & HellaSwag (\%) & Peak Mem (MB) & Steps/s & Time (h) \\
\midrule
LoLCATs baseline (no KD)             & 53.49 & 30.55 & 31.00 & 25\,408 & 2.31 & 7.22 \\
\gls{bkd}                            & 235.37 & 30.89 & 31.26 & 25\,428 & 2.19 & 7.61 \\
\gls{hkd}                            & 320.55 & 30.91 & 31.03 & 42\,620 & 1.01 & 16.44 \\
\gls{bkd} $\rightarrow$ \gls{hkd}    & 285.53 & 30.86 & \textbf{31.34} & 42\,622 & 0.64 & 12.94 \\
\gls{reskd}                          & 441.36 & 30.95 & 30.85 & 44\,130 & \textbf{3.62} & \textbf{4.61} \\
\gls{bkd} $\rightarrow$ \gls{reskd}  & 267.22 & 30.95 & 31.29 & 43\,066 & 0.56 & 37.56 \\
\gls{bkd} $\rightarrow$ Fine-tune    & \textbf{52.01} & \textbf{31.05} & 31.08 & \textbf{25\,408} & 2.29 & 9.22 \\
\midrule
Teacher (GPT-2, softmax)             & 37.37 & 32.56 & 31.14 & --- & --- & --- \\
\bottomrule
\end{tabular}
\end{table}

%% file: generated/qwen3_qat.tex
\begin{table}[htbp]
\centering\footnotesize
\caption{\textbf{Qwen3-1.7B INT4/INT5/INT6 QAT with distillation.} HKD dominates at INT4 (+6.76pp MMLU vs.\ QAT-alone, +9.42pp ARC-E); paradigms converge at INT5 and INT6 where PTQ already approaches the bf16 ceiling. Benchmarks: ARC-E/C~\cite{clark2018think}, HS = HellaSwag, (N) = normalized accuracy. Bold = best per column within each precision group; bf16 ceiling and PTQ floor rows excluded from competition.}
\label{tab:qwen3_qat}
\begin{tabular}{@{}lcccccccc@{}}
\toprule
Method & MMLU & ARC-E & ARC-C(N) & HS(N) & Wiki PPL & Peak Mem & Steps/s & Time(h) \\
\midrule
bf16 pretrained (ceiling)    & 55.55 & 72.26 & 43.09 & 60.35 & 21.05 & --- & --- & --- \\
\midrule
INT4 PTQ (floor)             & 40.71 & 58.46 & 32.42 & 54.26 & 37.20 & --- & --- & --- \\
INT4 QAT (no teacher)        & 41.50 & 58.96 & 32.85 & 54.53 & 36.15 & \textbf{29\,144} MB & \textbf{1.76} & \textbf{3.16} \\
INT4 QAT + \gls{bkd}         & 42.67 & 59.51 & 33.87 & 54.41 & 33.70 & \textbf{29\,144} MB & \textbf{1.76} & \textbf{3.16} \\
INT4 QAT + \gls{hkd}         & \textbf{48.26} & \textbf{65.57} & \textbf{36.69} & \textbf{54.66} & \textbf{29.98} & 35\,162 MB & 0.80 & 6.97 \\
INT4 QAT + \gls{reskd}       & 43.75 & 63.85 & 35.07 & 53.94 & 31.11 & 32\,436 MB & 0.92 & 6.02 \\
\midrule
INT5 PTQ (floor)             & 53.32 & 71.09 & 40.78 & 59.54 & 25.89 & --- & --- & --- \\
INT5 QAT (no teacher)        & 53.27 & 70.96 & 41.47 & 59.68 & 25.30 & 33\,624 MB & \textbf{1.13} & \textbf{4.90} \\
INT5 QAT + \gls{bkd}         & \textbf{54.96} & 72.77 & \textbf{42.49} & 59.23 & 22.39 & 40\,258 MB & 0.86 & 6.44 \\
INT5 QAT + \gls{hkd}         & 54.89 & \textbf{72.81} & \textbf{42.49} & 59.73 & 22.09 & 35\,162 MB & 0.79 & 6.99 \\
INT5 QAT + \gls{reskd}       & 51.99 & 70.83 & 41.47 & \textbf{59.75} & \textbf{21.96} & \textbf{31\,972} MB & 0.86 & 6.43 \\
\midrule
INT6 PTQ (floor)             & 54.66 & 72.77 & 43.26 & 60.18 & 21.26 & --- & --- & --- \\
INT6 QAT (no teacher)        & 55.33 & 72.69 & 43.60 & \textbf{60.29} & \textbf{20.77} & 33\,624 MB & \textbf{1.14} & \textbf{4.89} \\
INT6 QAT + \gls{bkd}         & 54.36 & 73.40 & 42.49 & 59.91 & 21.41 & 40\,260 MB & 0.78 & 7.15 \\
INT6 QAT + \gls{hkd}         & 55.58 & 73.02 & \textbf{43.86} & 59.89 & 21.66 & 35\,162 MB & 0.80 & 6.94 \\
INT6 QAT + \gls{reskd}       & \textbf{55.83} & \textbf{73.44} & 43.43 & 60.23 & 21.36 & \textbf{31\,972} MB & 0.86 & 6.47 \\
\bottomrule
\end{tabular}
\end{table}